\documentclass[]{research}
\usepackage{pgfplots}
\usepackage{graphicx}
\pgfplotsset{compat=1.18}
\usepackage{arydshln}

\usepackage{hyperref}
\usepackage{url}
\usepackage{enumitem}

\usepackage{amsmath}
\usepackage{amssymb}
\usepackage{mathtools}
\usepackage{amsthm}
\usepackage[table,dvipsnames]{xcolor}

\usepackage{tabularx}
\usepackage{booktabs} % for \toprule, \midrule, \bottomrule

\usepackage{listings}

\usepackage[most]{tcolorbox}
\tcbset{
    colback=gray!10,
    colframe=gray!50,
    arc=3mm,
    boxrule=0.5pt,
    left=6pt,
    right=6pt,
    top=6pt,
    bottom=6pt,
    center title,
    fonttitle=\bfseries,
}

\usepackage{microtype}
\usepackage{wrapfig}
\usepackage{float}
\usepackage{subcaption}

\usepackage{multirow}
\usepackage{threeparttable}
\usepackage{courier}
\usepackage{array,ragged2e}
\usepackage{tablefootnote}

\usepackage{fontawesome5}

\newcolumntype{P}[1]{>{\RaggedRight\arraybackslash}p{#1}}

\makeatletter
\def\adl@drawiv#1#2#3{%
        \hskip.5\tabcolsep
        \xleaders#3{#2.5\@tempdimb #1{1}#2.5\@tempdimb}%
                #2\z@ plus1fil minus1fil\relax
        \hskip.5\tabcolsep}
\newcommand{\cdashlinelr}[1]{%
  \noalign{\vskip\aboverulesep
           \global\let\@dashdrawstore\adl@draw
           \global\let\adl@draw\adl@drawiv}
  \cdashline{#1}
  \noalign{\global\let\adl@draw\@dashdrawstore
           \vskip\belowrulesep}}
\makeatother

\title{\textsc{PRISM}: Demystifying Retention and Interaction in Mid-Training}
\author[]{Bharat Runwal}
\author[]{Ashish Agrawal}
\author[]{Anurag Roy}
\author[]{Rameswar Panda}

\affiliation[]{IBM Research, MIT-IBM Watson AI Lab}

\abstract{
We present \textbf{\textsc{PRISM}}, a comprehensive empirical study of mid-training 
design choices for large language models (LLMs). Through controlled experiments across 
seven base models spanning four families (Granite, LLaMA, Mistral, Nemotron-H), two 
architecture types (dense Transformer and attention-Mamba hybrid), and scales from 3B 
to 24B parameters, we show that a mid-training phase of \textbf{$\sim$27B} high-quality 
tokens yields consistent gains of \textbf{+15 to +40 points} on math, \textbf{+5 to +12 
points} on code, and \textbf{+6 to +13 points} on science (GPQA-Diamond) benchmarks 
while preserving general performance. The full $\textsc{PRISM} \to \text{RL}$ pipeline 
improves the macro-average (domain-weighted) across six reasoning benchmarks from under 12 to \textbf{29--42} 
(a 3--4$\times$ improvement), whereas RL applied directly to most of the base models remains 
substantially less effective, with AIME scores near zero. Data composition choices matter 
most at mid-training, not at RL: including science data during mid-training unlocks 
\textbf{+17 to +28 point} GPQA-Diamond gains during RL, while changing the RL mix 
produces ${<}2$ point differences. Mechanistically, mid-training densely restructures 
${>}90\%$ of model weights, while RL makes sparse, front-loaded refinements to 
${\sim}5\%$ of parameters. Representation analysis (CKA) across three models and three 
input distributions confirms that RL consistently preserves mid-training's representational 
geometry ($>$0.998 CKA) across both dense Transformers and hybrid architectures. Crucially, 
RL applies identical weight changes regardless of starting point, yet only succeeds on 
mid-trained models, consistent with mid-training placing the model in a weight 
configuration from which RL can effectively improve performance. Our results demonstrate that retention-aware 
mid-training is a highly effective intermediate step for reliable reasoning enhancement 
and provide practical guidance for designing robust mid-training pipelines.
}

\correspondence{Bharat Runwal at \email{bharatrunwal@gmail.com}}

\metadata[\faGlobe\ Project Page]{\href{https://bharat-runwal.github.io/PRISM/}{Website}}
\metadata[\raisebox{-0.1em}{\includegraphics[height=1.1em]{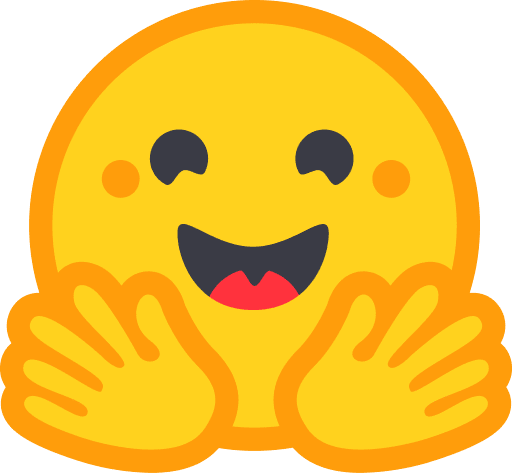}}\ Models \& Data]{\href{https://huggingface.co/PRISM-Midtraining}{HuggingFace}}

\begin{document}

\maketitle

% \section{Introduction}

\section{Introduction}

The training pipeline for Large Language Models (LLMs) has evolved beyond the traditional two-stage recipe of pre-training followed by alignment. State-of-the-art models now incorporate an additional intermediate stage, \emph{mid-training}, in which higher-quality, domain-focused data mixtures are used to imbue reasoning capabilities before downstream fine-tuning and reinforcement learning (RL)~\cite{5team2025glm45agenticreasoningcoding,olmo2025olmo3}. Yet despite its growing adoption, mid-training remains poorly understood: the field lacks systematic guidance on \emph{what data to use}, \emph{when to apply it}, \emph{how it interacts with RL}, and \emph{whether it generalizes across architectures}.

\begin{figure}[t]
    \centering
    \includegraphics[width=0.80\textwidth]{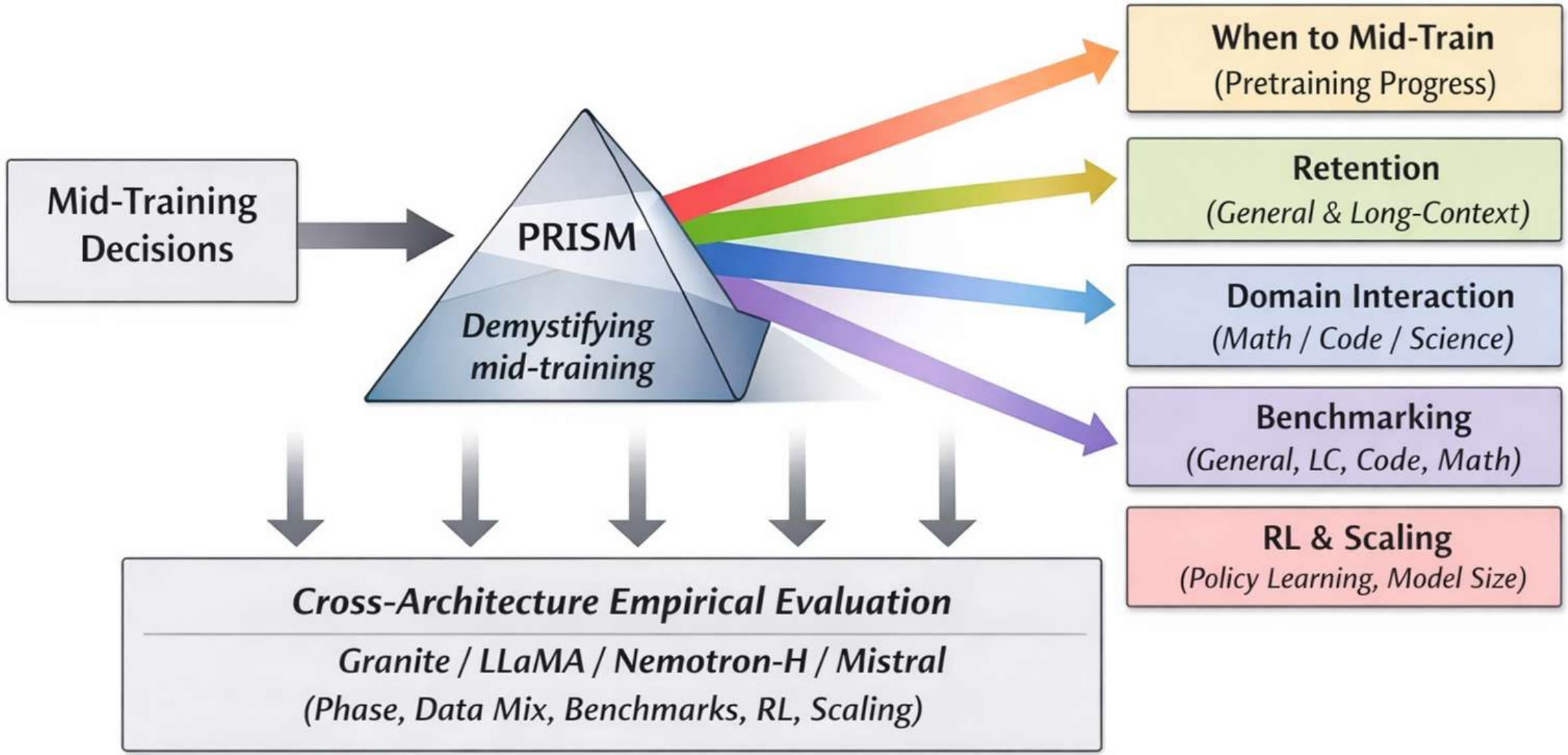}
    \caption{
    \textbf{\textsc{PRISM} overview.}
    Mid-training decisions are decomposed into their principal design axes, including retention of general and long-context abilities, domain interaction (math, code, science), benchmark selection, reinforcement learning compatibility, and scaling behavior. \textsc{PRISM} enables holistic evaluation of mid-training choices across model families at scale.
    }
    \label{fig:PRISM}
\end{figure}

We present \textbf{\textsc{PRISM}} (Demystifying Retention and Interaction in Mid-Training), shown in Fig.~\ref{fig:PRISM}, a comprehensive empirical study that addresses these questions through controlled experiments across seven open-source base models spanning four families (Granite, LLaMA, Mistral, Nemotron-H), two architecture types (dense Transformer and attention-Mamba hybrid), and scales from 3B to 24B parameters. Using targeted mid-training mixtures of only \textbf{$\sim$27B high-quality tokens}, we produce the following key findings:

\begin{itemize}
  \item[$\bullet$] \textbf{Mid-training substantially improves reasoning performance.} Across all tested models, \textsc{PRISM} yields \textbf{+15 to +40 point} gains on math benchmarks and \textbf{+5 to +12 points} on code, with science gains of \textbf{+6 to +13 points} on Granite and hybrid models, while preserving general-purpose performance.

  \item[$\bullet$] \textbf{Mid-training significantly enhances RL effectiveness.} The full $\textsc{PRISM} \to \text{RL}$ pipeline improves the macro-average (domain-weighted) across six reasoning benchmarks (AIME'24, AIME'25, MATH500, LiveCodeBench, Codeforces, GPQA-Diamond) from under 12 to \textbf{29--42}, a 3--4$\times$ improvement. RL applied directly to base models is substantially less effective, with AIME scores remaining near zero.

  \item[$\bullet$] \textbf{Data composition matters most at mid-training, not at RL.} Changing the mid-training mix from Math+Code to Math+Code+Science shifts AVG\footnote{AVG is computed as the mean of three domain scores: Code Avg (mean of LiveCodeBench and Codeforces), Math Avg (mean of AIME'24, AIME'25, and MATH500), and GPQA-Diamond.} by \textbf{+3 to +6 points}, while changing the RL mix produces $<$2 point differences. Science data at mid-training unlocks \textbf{+17 to +28 point} GPQA-Diamond gains during RL.

  \item[$\bullet$] \textbf{Benefits generalize across architectures and scales.} Both dense Transformers and attention-Mamba hybrids benefit consistently from \textsc{PRISM}, from 3B to 24B parameters.

  \item[$\bullet$] \textbf{RL expands the solvability frontier.} For Granite-3.3, RL on \textsc{PRISM}-mid-trained models progressively solves prompts that were initially unsolvable, with training curves that remain non-saturating across hundreds of steps.

  \item[$\bullet$] \textbf{Mid-training and RL operate through fundamentally different mechanisms.} Weight-level analysis reveals that mid-training densely restructures $>$90\% of parameters, while RL sparsely refines $\sim$5\%, with identical weight footprints regardless of whether mid-training preceded it. Representation analysis (CKA) across three models and three input distributions confirms that RL consistently preserves mid-training's representational geometry ($>$0.998 CKA) across both dense Transformers and hybrid architectures, while mid-training's representational impact is model-specific. RL optimization is front-loaded, with most weight changes in the first $\sim$200--400 steps. Behaviorally, mid-training produces extended reasoning chains in model outputs. On held-out MATH500 problems, the full pipeline improves pass rates from 2.6--66.6\% (base) to 64.6--83.0\% across three model families.
\end{itemize}

The term \emph{mid-training} has been used inconsistently in the literature. Some works treat it as a long-context extension phase~\cite{abdin2024phi3technicalreporthighly}, others as a higher-quality annealing stage for domain knowledge~\cite{olmo20252olmo2furious}, and recent work investigates mid-training choices that prepare models for RL by incorporating instruction-following data and chain-of-thought traces~\cite{wang2025octothinkermidtrainingincentivizesreinforcement}. These different usages have converged in practice, but the field lacks a holistic study that systematically quantifies the trade-offs induced by mid-training design choices across data mixtures, evaluation strategies, and downstream RL. \textsc{PRISM} fills this gap.

The rest of the paper is organized as follows. We first discuss limitations of prior mid-training approaches, then describe our data mixtures and benchmark selection. We study \emph{when} to mid-train, followed by domain-wise and cross-model-family analyses. We then present ablation studies on long-context restoration, context length, and token budget. We provide a detailed analysis of how reinforcement learning interacts with mid-trained models, including balanced vs.\ unbalanced RL mixes, base-model comparisons, solvability analysis, and a comprehensive pipeline-level evaluation. Finally, we present mechanistic analyses of the PRISM pipeline through weight divergence, representation similarity (CKA), prediction entropy, correctness studies, and RL weight trajectory dynamics across four model families and two architectures.

\section{Limitations of Prior Mid-Training Approaches}
\begin{center}
\begin{tcolorbox}[
  colback=cyan!10!teal!8,
  colframe=teal!50!black,
  boxrule=0.4pt,
  arc=2mm,
  left=10pt, right=10pt, top=7pt, bottom=7pt,
  width=0.82\linewidth,
  fontupper=\small,
  before skip=6pt,
  after skip=10pt
]
\textbf{Takeaway.}
Prior mid-training work often delivers domain-specific gains at the cost of generalization and holistic evaluation, and is rarely coupled with broad benchmark analysis or controlled studies of downstream RL behavior.
\end{tcolorbox}
\end{center}

Recent mid-training strategies for LLMs have demonstrated notable improvements in targeted capabilities such as coding and mathematical reasoning by introducing higher-quality or domain-focused data between pre-training and downstream fine-tuning or RL~\cite{olmo2025olmo3,wang2025octothinkermidtrainingincentivizesreinforcement}. However, the term \emph{mid-training} has been used inconsistently in the literature, referring to long-context extension, data annealing, and domain-specific capability refinement, without a unified framework or standardized evaluation.

\paragraph{\textbf{Narrow evaluation hides regressions.}}
Many studies report gains on a limited set of domain-specific benchmarks (e.g., math or code) without assessing whether these improvements preserve general-purpose capabilities or interact with other reasoning dimensions~\cite{wang2025octothinkermidtrainingincentivizesreinforcement}. Long-context extension work primarily evaluates context-window scaling and retrieval-style tasks, with limited analysis of its impact on general reasoning~\cite{abdin2024phi3technicalreporthighly}. Similarly, domain-focused mid-training recipes often emphasize improvements on math or code benchmarks while omitting broad generalization and cross-domain robustness evaluations~\cite{olmo20252olmo2furious,wang2025octothinkermidtrainingincentivizesreinforcement}.

\paragraph{\textbf{Interaction with RL remains underexplored.}}
A further shortcoming is the lack of controlled investigation into how mid-training interacts with downstream optimization, particularly reinforcement learning. While prior work suggests that certain mid-training strategies can facilitate RL by better aligning representations with downstream objectives, these claims are typically evaluated within narrow experimental settings and lack systematic comparison across model families, domains, and benchmark suites~\cite{wang2025octothinkermidtrainingincentivizesreinforcement,zhang2025interplaypretrainingmidtrainingrl}.

\paragraph{\textbf{Concurrent work.}}
Recent studies have begun to address parts of these gaps. \citet{liu2025midtrainingbridgespretrainingposttraining} show that mid-training can serve as a distributional bridge between pre-training and post-training, reducing distributional mismatch while preserving general capabilities. \citet{zhang2025interplaypretrainingmidtrainingrl} develop controlled experimental frameworks that isolate the contributions of pre-training, mid-training, and RL to reasoning generalization, highlighting mid-training as a critical yet underexplored stage. Small-scale controlled experiments provide valuable mechanistic insights with high ablation density. \textsc{PRISM} complements this line of work by examining mid-training design choices at 3B-24B scale across four model families, two architecture types, and multi-stage pipelines including RL, providing empirical coverage at a scale not addressed by prior work.

Taken together, these limitations motivate \textsc{PRISM}: a retention-aware empirical framework that evaluates mid-training choices across multiple domains, benchmark axes, and downstream RL behavior across model families to uncover trade-offs overlooked by prior work.

\section{Data Mixtures for Mid-Training}\label{sec:data-mix}

\begin{center}
\begin{tcolorbox}[
  colback=cyan!10!teal!8,
  colframe=teal!50!black,
  boxrule=0.4pt,
  arc=2mm,
  left=10pt, right=10pt, top=7pt, bottom=7pt,
  width=0.82\linewidth,
  fontupper=\small,
  before skip=6pt,
  after skip=10pt
]
\textbf{Takeaway.}
Mid-training performance is highly sensitive to data composition; carefully tuned mixtures that balance general web and instruction data with domain-specific reasoning sources yield robust retention and consistent gains, and we adopt these empirically validated splits across all experiments.
\end{tcolorbox}
\end{center}

\begin{table}[!tb]
\centering
\scriptsize
\setlength{\tabcolsep}{3pt}
\renewcommand{\arraystretch}{1.05}
\begin{tabular}{l l r}
\toprule
\textbf{Dataset} & \textbf{Type} & \textbf{Tokens (B)} \\
\midrule
DCLM-EDU~\cite{allal2025smollm2smolgoesbig} & General web data & 111.46 \\
\midrule
Open-R1 (MoT)~\cite{lozhkov2025openr1math220k}& Math reasoning & 0.60 \\
Nemotron Post-Training v1~\cite{NemotronPostTrainingDatasetV1}& Math & 35.93 \\
Megamath-Web-Pro~\cite{zhou2025megamathpushinglimitsopen} & Math web & 14.73 \\
\midrule
Open-R1 (MoT)~\cite{penedo2025codeforces} & Code reasoning & 1.18 \\
OpenCodeReasoning-2~\cite{ahmad2025opencodereasoningiisimpletesttime} & Code reasoning & 1.12 \\
RefinCode~\cite{huang2025opencoderopencookbooktoptier} & Code web & 186.44 \\
StarCoder2~\cite{lozhkov2024starcoder2stackv2}& Code web & 432.73 \\
\midrule
Open-R1 (MoT)~\cite{bercovich2025llamanemotronefficientreasoningmodels} & Science reasoning & 0.42 \\
OpenThoughts3~\cite{guha2025openthoughtsdatarecipesreasoning}& Science reasoning & 0.73 \\
\midrule
WildChat-1M~\cite{zhao2024wildchat}& Chat &  \\
Tulu-3 SFT Personas~\cite{lambert2025tulu3pushingfrontiers} & Chat & 0.91 \\
UltraChat-200k~\cite{ding2023enhancing}& Chat &  \\
\bottomrule
\end{tabular}
\caption{\textbf{Datasets used in mid-training mixtures.} Token counts are reported in billions (Granite 3.3, 8B).}
\label{tab:data-mixtures}
\end{table}

Table~\ref{tab:data-mixtures} summarizes the datasets used for mid-training. For the Math and Code domains, we use two data types: general web documents to retain knowledge from pretraining, and domain-specific reasoning datasets to imbue problem-solving ability. For Science, we include only reasoning-focused datasets. Prior work such as OctoThinker~\cite{wang2025octothinkermidtrainingincentivizesreinforcement} shows that incorporating a small amount of general instruction data can stabilize reinforcement learning; accordingly, we include chat and instruction-following datasets. However, unlike OctoThinker which focuses primarily on math, our goal is to support reasoning across diverse domains while retaining broad pretraining knowledge. To this end, we include general web data (DCLM-EDU) alongside domain-specific sources.

\subsection{Dataset Preprocessing}

We apply lightweight, deterministic preprocessing to all datasets to ensure data quality and evaluation integrity.

\paragraph{\textbf{Web data filtering.}}
For general web data, we use the DCLM-EDU corpus and retain documents with a quality score greater than or equal to 3, following the dataset's recommended filtering guidelines. This removes low-quality or noisy documents while preserving broad coverage of general knowledge.

\paragraph{\textbf{Reasoning datasets.}}
For OpenCodeReasoning-2, we retain only samples whose judgment is marked as \texttt{right} by the QwQ evaluator model and for which sufficient test coverage is available (i.e., \texttt{pass\_rate} $\neq -1$). From this filtered pool, we randomly sample 60k Python examples and 60k C++ examples. Other reasoning datasets are used as provided, without additional filtering beyond standard deduplication.

\paragraph{\textbf{Chat and instruction-following data.}}
For chat-style datasets, all conversations are normalized by explicitly prefixing utterances with speaker roles (``User:'' and ``Assistant:''). For WildChat-1M, we further restrict the data to high-quality conversations generated by GPT-4, following prior evidence that such filtering improves stability in downstream reinforcement learning. For all reasoning datasets and chat data, we concatenate the question and answer with a single line break between them, following~\cite{wang2025octothinkermidtrainingincentivizesreinforcement}.

\begin{figure*}[t]
\centering
\includegraphics[width=0.95\textwidth]{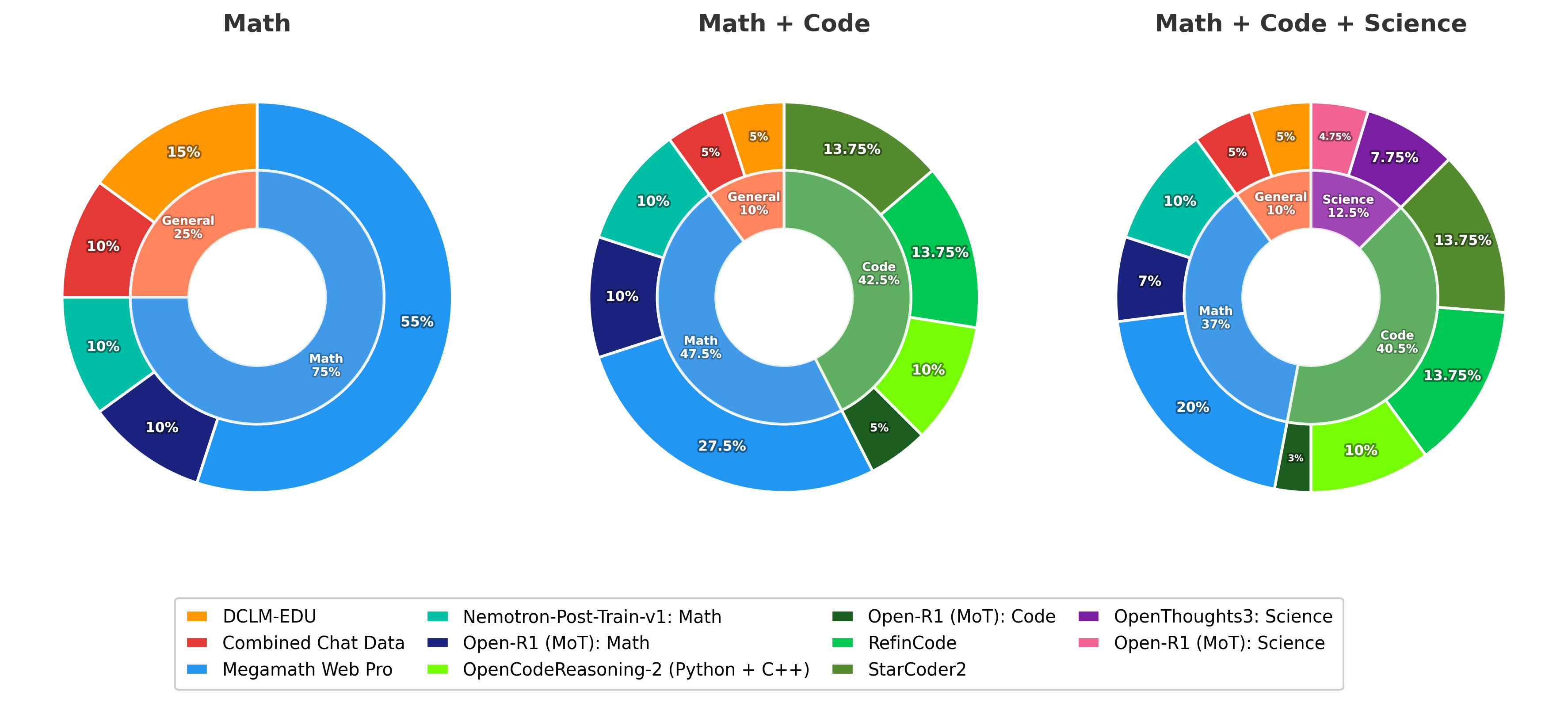}
\caption{\textbf{Mid-training data mixture configurations and per-source sampling percentages.} The outer ring shows individual data sources; the inner ring groups them by domain category.}
\label{fig:midtrain-mixtures}
\end{figure*}

Fig.~\ref{fig:midtrain-mixtures} reports the final per-source sampling weights for three progressively richer configurations: Math-only, Math+Code, and Math+Code+Science. After experimenting with various weightings across domains, we found these configurations to provide the best balance between retaining broad pretraining knowledge and inducing targeted domain improvements; consequently, we adopt these splits as the default sampling policy for all experiments reported in this paper.

\section{What to Evaluate: Benchmark Selection}

\begin{center}
\begin{tcolorbox}[
  colback=cyan!10!teal!8,
  colframe=teal!50!black,
  boxrule=0.4pt,
  arc=2mm,
  left=10pt, right=10pt, top=7pt, bottom=7pt,
  width=0.82\linewidth,
  fontupper=\small,
  before skip=6pt,
  after skip=10pt
]
\textbf{Takeaway.}
Evaluate mid-training with a \textit{balanced} suite that measures
(i) general LLM ability,
(ii) long-context behaviour, and
(iii) domain-specific reasoning;
otherwise, domain gains may mask regressions.
\end{tcolorbox}
\end{center}

In \textsc{PRISM} we adopt a deliberately broad evaluation setup to surface both gains and regressions introduced by mid-training. Concretely, we combine general leaderboards (LB-V1 and LB-V2) with focused long-context, code, math, and science evaluations so that improvements in a single domain cannot hide capability loss elsewhere. Table~\ref{tab:benchmarks} summarizes the benchmark categories and their roles.

\begin{table*}[t]
\centering
\scriptsize
\setlength{\tabcolsep}{3pt}
\renewcommand{\arraystretch}{0.9}
\begin{tabular}{l P{3.8cm} P{3.6cm} P{4.2cm}}
\toprule
\textbf{Category} & \textbf{Benchmarks} & \textbf{What it measures} & \textbf{Why it matters} \\
\midrule
General ability &
Leaderboard-V1 (LB-V1) (ARC, HellaSwag, MMLU, TruthfulQA, Winogrande, GSM8K), Leaderboard-V2 (LB-V2) (IFEval, BBH, MATH, GPQA, MUSR, MMLU-Pro) &
Broad multitask knowledge and robustness &
Detects generalization regressions hidden by domain-specific gains. \\

\midrule
Long-context &
RULER &
Long-context retrieval &
Ensures mid-training does not degrade long-context retrieval capabilities. \\

\midrule
Code &
LiveCodeBench ~\cite{jain2024livecodebenchholisticcontaminationfree}, Codeforces ~\cite{penedo2025codeforces} &
Executable program synthesis and reasoning &
Captures real-world coding ability. \\

\midrule
Math &
AIME~\cite{aime}, MATH500~\cite{lightman2023lets} & Mathematical reasoning &
Highly sensitive to data quality and mid-training composition. \\

\midrule
Science &
GPQA-Diamond ~\cite{rein2023gpqagraduatelevelgoogleproofqa} &
Expert-level scientific reasoning &
Probes scientific reasoning capabilities \\
\bottomrule
\end{tabular}
\caption{\textbf{Benchmark categories recommended for evaluating mid-training design choices.}}
\label{tab:benchmarks}
\end{table*}

\paragraph{\textbf{Practical guidance for benchmark selection.}}
As summarized in Table~\ref{tab:benchmarks}, effective evaluation of mid-training decisions requires both breadth and depth:

\begin{itemize}
    \item \textbf{Mix breadth and depth:} combine general-purpose leaderboards (LB-V1~\cite{open-llm-leaderboard-v1} and LB-V2~\cite{open-llm-leaderboard-v2}) with targeted domain benchmarks to expose global regressions while accurately measuring domain-specific gains.
    \item \textbf{Measure long-context retention explicitly:} evaluate long-context reasoning separately (e.g., RULER~\cite{hsieh2024rulerwhatsrealcontext}), as mid-training dominated by short-context data can degrade long-context capabilities, often necessitating an additional lightweight fine-tuning stage to recover performance (see Section~\ref{sec:long-context}).
\end{itemize}

\section{When to Mid-Train}
\begin{center}
\begin{tcolorbox}[
  colback=cyan!10!teal!8,
  colframe=teal!50!black,
  boxrule=0.4pt,
  arc=2mm,
  left=10pt, right=10pt, top=7pt, bottom=7pt,
  width=0.82\linewidth,
  fontupper=\small,
  before skip=6pt,
  after skip=10pt
]
\textbf{Takeaway.} On Granite-4 Micro (3B), mid-training is most effective when applied 
\textbf{after long-context pretraining}, yielding the largest gains in math, code, and 
science while preserving general reasoning. Whether this ordering generalizes across 
larger models or different architectures remains an open question. Conveniently, most 
open-source base models are released after long-context extension, making this the 
natural starting point in practice.
\end{tcolorbox}
\end{center}

\begin{wraptable}{r}{0.52\textwidth}
\vspace{-12pt}
\centering
\scriptsize
\setlength{\tabcolsep}{2.5pt}
\renewcommand{\arraystretch}{1.05}
\begin{tabular}{l cc cc c ccc}
\toprule
& \multicolumn{2}{c}{\textbf{Leaderbds.}}
& \multicolumn{2}{c}{\textbf{Code}}
& \textbf{Sci.}
& \multicolumn{3}{c}{\textbf{Math}} \\
\cmidrule(lr){2-3}
\cmidrule(lr){4-5}
\cmidrule(lr){6-6}
\cmidrule(lr){7-9}

\textbf{Stage}
& \textbf{V1} & \textbf{V2}
& \textbf{LCB} & \textbf{CF}
& \textbf{GPQA}
& \textbf{AI24} & \textbf{AI25} & \textbf{M500} \\
\midrule
Phase 3
& 63.30 & 19.44
& 7.05 & 8.61
& 19.53
& 9.38 & 16.09 & 65.88 \\

Phase 4
& 62.84 & 20.85
& 7.89 & 7.95
& 17.85
& 10.00 & 14.06 & 61.70 \\

After LC
& 62.91 & 20.53
& \textbf{10.39} & \textbf{6.18}
& \textbf{25.93}
& \textbf{23.59} & \textbf{20.94} & \textbf{77.44} \\
\bottomrule
\end{tabular}
\caption{\textbf{Effect of \emph{when} mid-training is applied} on Granite-4 Micro (3B).
Phase 3/4 = intermediate/late pretraining; After LC = after long-context extension.}
\label{tab:when-to-midtrain}
\vspace{-8pt}
\end{wraptable}
Mid-training is typically applied after pretraining, but the optimal timing within the pretraining pipeline remains unclear. Using Granite-4 Micro (3B), we apply the same mid-training recipe (Math+Code+Science, 8k context) at three different points: (i) after Phase~3 of pretraining, (ii) after Phase~4 (the final dense pretraining stage before long-context extension), and (iii) starting from the base model after long-context pretraining (Table~\ref{tab:when-to-midtrain}).

\paragraph{\textbf{Earlier phases yield gains, but later is better.}}
Mid-training at earlier phases already produces meaningful improvements, but later stages consistently translate the mid-training signal into stronger downstream performance. Compared to Phase 3, Phase 4 mid-training modestly improves Leaderboard V2 (from 19.44 to 20.85) while maintaining similar code performance. However, both Phase 3 and Phase 4 underperform the final base model on math and science benchmarks.

\paragraph{\textbf{After long-context extension produces the strongest results.}}
Applying mid-training after long-context extension yields the best overall performance. Math performance improves substantially, with AIME24 increasing from 9.38 (Phase 3) and 10.00 (Phase 4) to \textbf{23.59}, and MATH500 rising to \textbf{77.44}. Code performance also improves, with LiveCodeBench reaching \textbf{10.39}, while GPQA-Diamond reaches \textbf{25.93}, exceeding both earlier phases.

\paragraph{\textbf{General capabilities remain stable across timing choices.}}
General-purpose leaderboards remain relatively stable across stages, indicating that later mid-training does not introduce large regressions in broad capabilities. Overall, these results suggest that while mid-training can be effective at multiple stages, applying it after long-context capabilities are established yields the most consistent gains across math, code, and science. We note that this is a preliminary finding based on a single model (Granite-4 Micro, 3B), and whether the same ordering holds across larger models or different architectures remains an open question. Additionally, post-long-context base models may be stronger starting points in absolute terms, confounding the timing effect with base model quality. The practical implication is limited to: given a choice of when to apply mid-training, post-LC is a reasonable default, and it is also the natural starting point for our broader \textsc{PRISM} study since most publicly released base models (e.g., LLaMA, Mistral) have already undergone long-context extension.

\section{Domain-wise Effects of Mid-Training Data}
\begin{center}
\begin{tcolorbox}[
  colback=cyan!10!teal!8,
  colframe=teal!50!black,
  boxrule=0.4pt,
  arc=2mm,
  left=10pt, right=10pt, top=7pt, bottom=7pt,
  width=0.82\linewidth,
  fontupper=\small,
  before skip=6pt,
  after skip=10pt
]
\textbf{Takeaway.}
Mid-training performance is driven by data composition. Domain-specific data delivers large gains in its corresponding benchmarks, while balanced mixtures across math, code, and science achieve the best overall trade-off, improving domain reasoning while preserving general capabilities.
\end{tcolorbox}
\end{center}

Having established the data sources and empirically validated mixture configurations in Section~\ref{sec:data-mix}, we now examine how domain-specific data affects downstream performance. We mid-train the Granite-3.3 (8B) base model using three progressively richer data mixtures: Math-only, Math+Code, and Math+Code+Science, following the configurations in Fig.~\ref{fig:midtrain-mixtures}. All experiments use a fixed budget of \textbf{$\sim$27B} tokens at a context length of 8192; additional hyperparameters are in Appendix Section~\ref{sec:appendix-hparams}. We evaluate on both general-purpose leaderboards (LB-V1 and LB-V2) and domain-specific benchmarks, allowing us to isolate the effect of each domain and analyze the trade-offs between specialization and retention.

\begin{table*}[t]
\centering
\scriptsize
\setlength{\tabcolsep}{4pt}
\resizebox{\textwidth}{!}{
\begin{tabular}{l
  r r r r r r r
  r r r r r r r}
\toprule
 & \multicolumn{7}{c}{\textbf{Leaderboard V1}} & \multicolumn{7}{c}{\textbf{Leaderboard V2}} \\
\cmidrule(lr){2-8} \cmidrule(lr){9-15}
Mixture & ARC & HellaSwag & MMLU & TruthfulQA & Winogrande & GSM8K & OpenLLM V1 Avg
        & IFEval & BBH & MATH & GPQA & MUSR & MMLU-Pro & OpenLLM V2 Avg \\
\midrule
Base
& 61.95 & \textbf{83.46} & 62.56 & \textbf{52.24} & \textbf{80.35} & 56.33 & 66.15
& \textbf{46.62} & 24.68 & 10.20 & 6.38 & 8.88 & 23.82 & 20.10 \\

Math only
 & \textbf{62.54} & 78.72 & \textbf{64.29} & 46.04 & 75.30 & 71.95 & 66.47
 & 46.46 & 25.57 & 17.75 & 5.59 & 9.08 & \textbf{29.86} & 22.39 \\

Math + Code
 & 61.01 & 78.09 & 62.65 & 47.36 & 74.74 & 73.46 & 66.22
 & 45.56 & \textbf{26.87} & 18.43 & 5.93 & \textbf{10.60} & 28.40 & 22.63 \\

\rowcolor{teal!10}
Math + Code + Science
 & 61.69 & 78.12 & 62.98 & 46.96 & 74.90 & \textbf{74.22} & \textbf{66.48}
 & 46.44 & 26.32 & \textbf{20.02} & \textbf{7.27} & 8.60 & 29.55 & \textbf{23.03} \\
\bottomrule
\end{tabular}}
\caption{\textbf{Leaderboard V1 and V2 results for Granite-3.3-8B} mid-trained with the mixtures in Fig.~\ref{fig:midtrain-mixtures}.}
\label{tab:leaderboard-v1v2-results}
\end{table*}

\begin{wraptable}{R}{0.40\textwidth}
\vspace{-14pt}
\centering
\scriptsize
\setlength{\tabcolsep}{3.5pt}
\renewcommand{\arraystretch}{1.05}
\begin{tabular}{l r r r}
\toprule
\textbf{Mixture} & \textbf{Code} & \textbf{Math} & \textbf{GPQA} \\
\midrule
Base          & 2.07  & 8.95  & 22.56 \\
Math          & 2.81  & \textbf{36.43} & 17.34 \\
Math+Code     & \textbf{10.71} & 44.99 & 19.02 \\
\rowcolor{teal!10}
Math+Code+Sci & 10.58 & \textbf{48.75} & \textbf{29.12} \\
\bottomrule
\end{tabular}
\caption{\textbf{Domain-specific results} for Granite-3.3 (8B). Code/Math are averages; full results in Appendix Table~\ref{tab:app_domain-specific-results}.}
\label{tab:domain-specific-results}
\vspace{-6pt}
\end{wraptable}
\noindent\textbf{Math data drives the largest single-domain gains.}
Introducing math-specific data during mid-training leads to substantial improvements in mathematical reasoning. Compared to the baseline model, the Math-only mixture increases the Math average from 8.95 to \textbf{36.43}, a gain of \textbf{+27.48} points (Table~\ref{tab:domain-specific-results}). These gains demonstrate that high-quality math reasoning data is the primary driver of mathematical capability during mid-training.

\paragraph{\textbf{Code data is essential for programming benchmarks.}}
Adding code-specific data produces large improvements on programming benchmarks. While Math-only mid-training yields only marginal code gains over the baseline, increasing the Code average from 2.07 to 2.81 (\textbf{+0.74}), the Math+Code mixture raises the Code average to \textbf{10.71}, corresponding to a \textbf{+8.64} point improvement relative to the baseline (Table~\ref{tab:domain-specific-results}). Incorporating science data on top of code does not substantially alter code performance, with the Math+Code+Science mixture maintaining a similar Code average of \textbf{10.58}.

\paragraph{\textbf{Science data improves GPQA without sacrificing other domains.}}
Including science data during mid-training improves performance on GPQA-Diamond without deteriorating code or math performance. Compared to the Math+Code mixture, the Math+Code+Science mixture increases GPQA-Diamond from 19.02 to \textbf{29.12} (\textbf{+10.10} points). At the same time, the Code average remains stable (10.71 to \textbf{10.58}), and the Math average further improves from 44.99 to \textbf{48.75} (Table~\ref{tab:domain-specific-results}). These results show that science-focused data can be added without sacrificing gains in other reasoning domains.

\paragraph{\textbf{General performance is broadly maintained but with individual regressions.}}
Mid-training introduces measurable trade-offs on general-purpose benchmarks. On Leaderboard V1, the Math-only mixture improves the overall average from 66.15 to \textbf{66.47} (\textbf{+0.32}), driven primarily by gains on GSM8K, while exhibiting regressions on individual benchmarks such as HellaSwag ($\sim$5 points across all mixtures) and TruthfulQA (Table~\ref{tab:leaderboard-v1v2-results}). Leaderboard V2 averages increase monotonically with broader domain coverage, rising from 20.10 for the baseline to \textbf{22.39} for Math-only, \textbf{22.63} for Math+Code, and \textbf{23.03} for Math+Code+Science. Overall Leaderboard V1 averages remain near the baseline across mixtures, which we attribute in part to the consistent inclusion of general web data from DCLM-EDU; however, individual benchmarks such as HellaSwag show regressions of approximately 5 points, suggesting that domain-specific mid-training introduces some interference with general benchmarks.

\section{\textsc{PRISM} Effects Across Model Families}
\begin{center}
\begin{tcolorbox}[
  colback=cyan!10!teal!8,
  colframe=teal!50!black,
  boxrule=0.4pt,
  arc=2mm,
  left=10pt, right=10pt, top=7pt, bottom=7pt,
  width=0.82\linewidth,
  fontupper=\small,
  before skip=6pt,
  after skip=10pt
]
\textbf{Takeaway.}
Across model families, architectures, and scales, \textsc{PRISM} mid-training consistently improves reasoning performance. We observe gains of \textbf{+15 to +40 points} on math benchmarks and \textbf{+5 to +12 points} on coding benchmarks across all models. Science gains (GPQA-Diamond) are \textbf{+6 to +13 points} on Granite and hybrid models; for other families, science improvements primarily emerge after RL when science data is included at mid-training.
\end{tcolorbox}
\end{center}

We evaluate \textsc{PRISM} mid-training across a diverse set of model families, architectures, and scales. Our experiments include dense Transformer models: Granite-3.3 (8B)~\cite{granite2025granite33base}, LLaMA-3.1 (8B)~\cite{grattafiori2024llama3herdmodels}, Mistral-7B~\cite{jiang2023mistral7b}, Mistral-Small-24B~\cite{mistral2025mistralsmall3}, and Granite-4 Micro (3B). We additionally consider hybrid architectures within the Granite-4 family~\cite{granite2025granite40collection}: Granite-4-H Micro (3B) and Nemotron-H (8B)~\cite{nvidia2025nemotronhfamilyaccurateefficient}, which combine attention and Mamba layers. Additional architectural and training details are in Appendix Section~\ref{sec:appendix-hparams}. For most experiments, we perform \textsc{PRISM} mid-training at an 8k context length, which offers a favorable trade-off between computational cost and downstream performance (Section~\ref{sec:ablation-context}).

\begin{table*}[t]
\centering
\scriptsize
\setlength{\tabcolsep}{3pt}
\renewcommand{\arraystretch}{1.05}
\begin{tabular}{l l r r r r r r r r r r}
\toprule
 &  & \multicolumn{2}{c}{\textbf{Leaderboards}}
 & \multicolumn{3}{c}{\textbf{Code}}
 & \textbf{Science}
 & \multicolumn{4}{c}{\textbf{Math}} \\
\cmidrule(lr){3-4}
\cmidrule(lr){5-7}
\cmidrule(lr){8-8}
\cmidrule(lr){9-12}
Model & Variant
& LB V1 & LB V2
& LCB & CF & Code Avg
& GPQA-D
& AIME24 & AIME25 & MATH500 & Math Avg \\
\midrule

Granite-3.3 (8B) & Base
& 66.15 & 20.10
& 2.15 & 1.99 & 2.07
& 22.56
& 0.46 & 0.31 & 26.09 & 8.95 \\

\rowcolor{teal!10}
 & \textsc{PRISM}
& 66.48 & 23.03
& 10.63 & 10.52 & {10.58}
& 29.12
& 37.18 & 27.96 & 81.11 & {48.75} \\
\midrule

Granite-4 Micro (3B) & Base
& 66.01 & 21.82
& 0.24 & 2.28 & 1.26
& 21.55
& 16.09 & 12.34 & 50.42 & 26.28 \\

\rowcolor{teal!10}
 & \textsc{PRISM}
& 62.91 & 20.53
& 10.87 & 6.25 & 8.56
& 34.34
& 27.19 & 22.29 & 79.40 & 42.96 \\
\midrule

Granite-4-H Micro (3B) & Base
& 64.49 & 18.99
& 0.60 & 0.88 & 0.74
& 20.88
& 7.08 & 2.70 & 30.17 & 13.32 \\

\rowcolor{teal!10}
 & \textsc{PRISM}
& 64.21 & 18.75
& 15.53 & 8.02 & 11.78
& 32.66
& 33.69 & 23.49 & 82.73 & 46.64 \\
\midrule

Nemotron-H-8k (8B) & Base
& 71.35 & 23.84
& 1.19 & 3.60 & 2.39
& 4.21
& 2.13 & 2.29 & 49.46  & 17.96 \\

\rowcolor{teal!10}
 & \textsc{PRISM}
& 68.84  & 26.08
& 13.02 &10.52 &11.77
& 31.98
& 19.21 & 22.76 & 76.63 &  39.53 \\
\midrule

Mistral-7B & Base
& 60.88 & 14.89
& 0.00 & 0.15 & 0.07
& 26.94
& 0.00 & 0.10 & 1.68 & 0.59 \\

\rowcolor{teal!10}
 & \textsc{PRISM}
& 59.99 & 19.68
& 10.16 & 9.42 & 9.79
& 24.07
& 28.85 & 24.27 & 70.71 & 41.28 \\
\midrule

LLaMA-3.1 (8B) & Base
& 62.76 & 14.09
& 0.00 & 0.07 & 0.04
& 20.20
& 0.05 & 0.15 & 6.51 & 2.24 \\

\rowcolor{teal!10}
 & \textsc{PRISM}
& 65.21 & 21.46
& 6.09 & 5.45 & 5.77
& 21.04
& 16.45 & 19.32 & 73.47 & 36.41 \\
\midrule

Mistral-Small (24B) & Base
& 74.98 & 27.29
& 0.00 & 0.29 & 0.15
& 22.55
& 0.78 & 0.73 & 26.92 & 9.48 \\

\rowcolor{teal!10}
 & \textsc{PRISM}
& 69.52 & 27.42
& 10.03 & 10.08 & 10.06
& 22.05
& 32.91 & 27.34 & 80.80 & 47.02 \\

\bottomrule
\end{tabular}
\caption{\textbf{Base versus \textsc{PRISM} (Math+Code+Science) mid-training results across model families.} Code Avg is the mean of LiveCodeBench (LCB) and Codeforces (CF). Math Avg is the mean of AIME24, AIME25, and MATH500. All values are reported to two decimal places.}
\label{tab:model-family-results}
\end{table*}

Table~\ref{tab:model-family-results} summarizes the impact of \textsc{PRISM} mid-training across this diverse set of models. Across all families, \textsc{PRISM} consistently improves mathematical, coding, and scientific reasoning, while changes to general-purpose leaderboards are smaller and more model dependent.

\paragraph{\textbf{Mid-training benefits generalize across all model families.}}
\textsc{PRISM} yields strong improvements regardless of the underlying model family. Mistral-7B shows some of the largest gains, with MATH500 improving from 1.68 to \textbf{70.71} and Codeforces from 0.15 to \textbf{9.42}. Mistral-Small (24B) similarly improves MATH500 from 26.92 to \textbf{80.80}. LLaMA-3.1 (8B) benefits as well, improving AIME24 from 0.05 to \textbf{16.45} and LiveCodeBench from 0.00 to \textbf{6.09}. These trends demonstrate that \textsc{PRISM} is effective across distinct model families and training recipes.

\paragraph{\textbf{Hybrid architectures benefit as much as dense models.}}
Within the Granite-4 family, we observe that hybrid variants respond strongly to \textsc{PRISM} mid-training. The dense Granite-4 Micro (3B) shows substantial gains, improving MATH500 from 50.42 to \textbf{79.40} and LiveCodeBench from 0.24 to \textbf{10.87}. Hybrid models, including Granite-4-H Micro (3B) and Nemotron-H (8B), also exhibit large improvements. For example, Nemotron-H (8B) increases AIME24 from 2.13 to \textbf{19.21}, AIME25 from 2.29 to \textbf{22.76}, and MATH500 from 49.46 to \textbf{76.63}. While these results suggest that hybrid architectures can effectively leverage mid-training signal, differences in pretraining data and model scale prevent a direct attribution of these gains to architecture alone.

\paragraph{\textbf{Larger models achieve higher absolute scores, but gains are universal.}}
Although larger models achieve higher absolute scores, \textsc{PRISM} delivers meaningful gains at all scales. Smaller models often exhibit larger relative improvements, while larger models realize strong absolute gains without severe degradation on leaderboards. For instance, Mistral-Small (24B) improves MATH500 by more than \textbf{+50} points while maintaining Leaderboard V2 performance, whereas LLaMA-3.1 (8B) improves Leaderboard V2 from 14.09 to \textbf{21.46}. Overall, these results suggest that retention-aware, multi-domain mid-training provides consistent benefits across parameter scales.

\section{Ablation Studies}

Beyond data composition and model family, several practical design choices shape mid-training outcomes: how to restore long-context ability lost during short-context mid-training, how much context length to use during mid-training itself, and how many tokens are sufficient before gains saturate. We study each of these in controlled ablations on Granite models.

\subsection{Restoring Long-Context Ability After Mid-Training}\label{sec:long-context}

Mid-training is performed at an 8k context length, which naturally degrades long-context capabilities inherited from pretraining. In this section, we study practical strategies to restore long-context performance after mid-training using Granite-3.3 (8B). We evaluate two approaches: (i) directly performing a short long-context extension phase on the mid-trained checkpoint, and (ii) linearly \textit{merging} the mid-trained model with the base model prior to long-context extension. For both approaches, we further compare training all parameters versus training only attention modules during the long-context phase.

Details of the data construction and preprocessing used for long-context restoration are provided in Appendix Section~\ref{sec:appendix-lc}. In particular, we augment the training data with code examples containing longer chains of thought, apply filtering to remove short-context samples, and use best-fit packing to efficiently construct long-context training sequences.

\begin{table}[t]
\centering
\scriptsize
\setlength{\tabcolsep}{2pt}
\renewcommand{\arraystretch}{0.95}
\resizebox{\columnwidth}{!}{%
\begin{tabular}{
p{3.8cm} r r r r r r r r r r r r r
}
\toprule
 & \multicolumn{5}{c}{\textbf{RULER}}
 & \multicolumn{4}{c}{\textbf{Code / Science}}
 & \multicolumn{4}{c}{\textbf{Math}} \\
\cmidrule(lr){2-6}
\cmidrule(lr){7-10}
\cmidrule(lr){11-14}

\textbf{Model Variant}
& \textbf{8k} & \textbf{16k} & \textbf{32k} & \textbf{64k} & \textbf{128k}
& \textbf{LCB} & \textbf{CF} & \textbf{Code Avg} & \textbf{GPQA-D}
& \textbf{AIME24} & \textbf{AIME25} & \textbf{MATH500} & \textbf{Math Avg} \\
\midrule

Granite-3.3 Base
& 85.81 & 82.40 & 75.53 & 64.91 & 59.09
& 2.15 & 1.99 & 2.07 & \textbf{22.56}
& 0.46 & 0.31 & 26.09 & 8.95 \\
\midrule

Mid-Train (Math+Code)
& 89.02 & 60.44 & 21.52 & 11.71 & 6.46
& 11.11 & 10.30 & 10.71 & 19.02
& 32.44 & 28.33 & 74.22 & 44.99 \\
\rowcolor{teal!10}
Mid-Train + LC (Attention)
& 90.04 & 82.56 & 71.47 & 54.63 & 36.32
& 23.78 & 15.53 & 19.65 & 17.85
& 36.56 & 32.55 & 67.20 & \textbf{45.44} \\
\rowcolor{teal!10}
Mid-Train + LC (Full)
& 89.29 & 80.74 & 70.86 & 56.02 & 38.41
& 29.99 & 21.04 & 25.52 & 14.48
& 35.21 & 30.36 & 62.30 & 42.62 \\

\midrule

Merge (15\% Base + 85\% Mid-Train)
& 89.12 & 69.76 & 32.63 & 15.44 & 11.32
& 10.75 & 10.96 & 10.86 & 22.22
& 28.39 & 24.90 & 72.97 & 42.09 \\

Merge + LC (Attention)
& 90.00 & 84.27 & 73.31 & 57.27 & 37.75
& 26.16 & 17.29 & 21.73 & 17.51
& 33.85 & 28.75 & 71.28 & 44.63 \\

\rowcolor{teal!10}
Merge + LC (Full)
& 89.83 & 84.08 & 73.89 & 60.06 & \textbf{42.16}
& 29.51 & 21.56 & \textbf{25.54} & 15.82
& 33.75 & 30.78 & 68.91 & 44.48 \\

\bottomrule
\end{tabular}}
\caption{
\textbf{Restoring long-context capability after mid-training for Granite-3.3 (8B).}
RULER is evaluated from 8k to 128k input lengths.
Downstream performance includes Code (LiveCodeBench, Codeforces),
Science (GPQA-Diamond), and Math (AIME24, AIME25, MATH500).
}
\label{tab:long-context-merged}
\end{table}

\begin{figure}[!htbp]
    \centering
    \includegraphics[width=\linewidth]{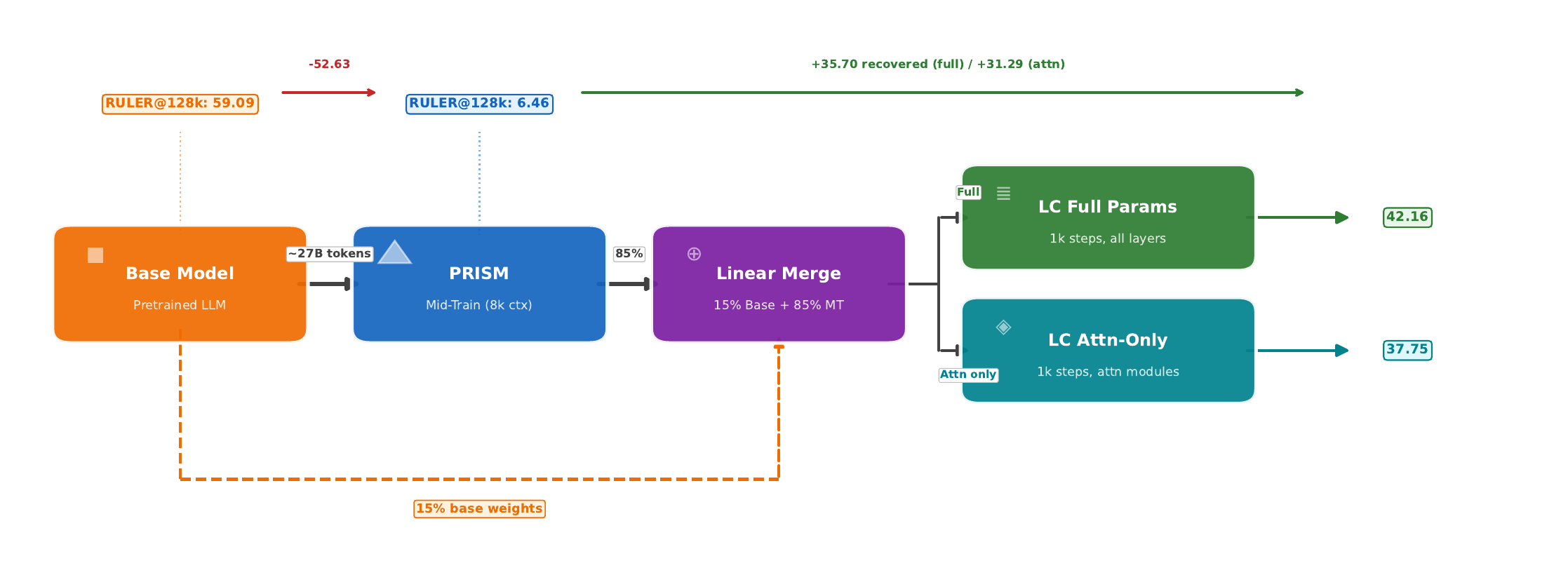}
    \caption{\textbf{Long-context restoration pipeline.} After PRISM mid-training degrades RULER@128k from 59.09 to 6.46, a linear merge (15\% base + 85\% mid-trained) followed by long-context extension recovers performance to 42.16 (full params) or 37.75 (attention-only).}
    \label{fig:merge_lc_flow}
\end{figure}

\paragraph{Mid-training severely degrades long-context ability.}
While the Granite-3.3 (8B) base model achieves a RULER score of 59.09 at 128k context, the Math+Code mid-trained model drops sharply to \textbf{6.46}, despite strong performance at short context lengths (89.02 at 8k). This confirms that mid-training with short-context data alone disrupts long-context behaviors learned during pretraining, motivating the need for explicit restoration strategies. Figure~\ref{fig:merge_lc_flow} illustrates the two restoration pipelines we evaluate.

\paragraph{A brief long-context extension phase largely restores performance.}
Applying 1k steps of long-context training directly on the mid-trained model raises RULER at 128k from 6.46 to \textbf{36.32} when training attention modules only, and to \textbf{38.41} when training all parameters. These improvements are consistent across intermediate context lengths, with RULER at 64k improving from 11.71 to over 54.63. At the same time, downstream reasoning performance is preserved or improved: Code Avg increases from 10.71 to \textbf{19.65} (attention-only) and \textbf{25.52} (full), while Math Avg remains above 42 across both variants (Table~\ref{tab:long-context-merged}).

\paragraph{Merging with the base model yields the strongest recovery.}
Merging the mid-trained model with the base model prior to long-context extension yields the strongest recovery at long context lengths. With a 15\% base and 85\% mid-trained linear merge followed by long-context training, RULER at 128k improves further to \textbf{42.16}, narrowing much of the gap to the base model. Importantly, this approach maintains strong downstream reasoning performance, achieving a Code Avg of \textbf{25.54} and a Math Avg of \textbf{44.48}. Across strategies, full-parameter long-context training yields the strongest recovery, while attention-only training still provides meaningful RULER improvements with competitive downstream performance, offering a practical efficiency/performance trade-off.

\subsection{Effect of Mid-Training Context Length}\label{sec:ablation-context}

\begin{wraptable}{R}{0.52\textwidth}
\vspace{-14pt}
\centering
\scriptsize
\setlength{\tabcolsep}{3pt}
\renewcommand{\arraystretch}{1.05}
\begin{tabular}{l r r r r r r r r}
\toprule
\textbf{Context} & \textbf{LB-V1} & \textbf{LB-V2} & \textbf{LCB} & \textbf{CF} & \textbf{GPQA} & \textbf{AIME24} & \textbf{AIME25} & \textbf{M500} \\
\midrule
Base
& 66.01 & 21.82
& 0.24 & 2.28
& 21.55
& 16.09 & 12.34 & 50.42 \\
\midrule
8k
& 62.91 & 20.53
& 10.87 & 6.25
& 34.34
& 27.19 & 22.29 & 79.40 \\
16k
& 64.23 & 20.37
& 12.19 & 8.90
& 38.89
& 31.82 & 25.26 & 82.47 \\
32k
& 64.48 & 21.05 
& 14.93 & 7.50
& 39.89
& 30.98 & 21.87 & 82.70 \\
\bottomrule
\end{tabular}
\caption{\textbf{Mid-training context length ablation} on Granite-4 Micro (3B) with Math+Code+Science mix. V1/V2 = Leaderboard V1/V2.}
\label{tab:context-ablation}
\vspace{-6pt}
\end{wraptable}
We study the effect of increasing the mid-training context length while keeping the data mixture fixed to Math+Code+Science and maintaining a comparable token budget (Table~\ref{tab:context-ablation}). All ablations use the Granite-4 Micro (3B) dense model.

Increasing context from 8k to 16k yields the largest gains: MATH500 improves from 79.40 to \textbf{82.47}, AIME24 from 27.19 to \textbf{31.82}, Codeforces from 6.25 to \textbf{8.90}, and GPQA-Diamond from 34.34 to \textbf{38.89}. These results indicate that moderate long-context mid-training strengthens the model's ability to leverage multi-step reasoning signals present in math, code, and science data.

However, gains largely saturate beyond 16k. Extending to 32k yields small additional improvements on LiveCodeBench (12.19 $\to$ \textbf{14.93}), but also observe slight regression in other benchmarks. General-purpose performance remains stable, with Leaderboard V1 partially recovering from 62.91 at 8k to \textbf{64.48} at 32k. Overall, 16k provides the most favorable balance between reasoning gains and training efficiency.

\subsection{Effect of Mid-Training Token Budget}

\begin{wraptable}{R}{0.44\textwidth}
\vspace{-14pt}
\centering
\scriptsize
\setlength{\tabcolsep}{3pt}
\renewcommand{\arraystretch}{1.05}
\begin{tabular}{l c c c c c}
\toprule
\textbf{Tok.~(B)} & \textbf{LB-V1} & \textbf{LB-V2} & \textbf{Code} & \textbf{GPQA} & \textbf{Math} \\
\midrule
Base  & 66.01 & 21.82 & 1.26 & 21.55 & 26.28 \\
\midrule
10.49 & 63.45 & 19.50 & \textbf{9.59} & 19.19 & 40.21 \\
15.73 & 63.24 & 19.79 & 9.02 & \textbf{23.06} & 42.07 \\
26.21 & 63.28 & 19.63 & 8.69 & 19.19 & 42.22 \\
31.46 & 63.16 & 20.05 & 7.62 & 21.38 & \textbf{42.42} \\
\bottomrule
\end{tabular}
\caption{\textbf{Token budget ablation} on Granite-4 Micro (3B), Math+Code mix. Full table in Appendix~\ref{appendix:token-budget-ablation}.}
\label{tab:token-budget-ablation}
\vspace{-6pt}
\end{wraptable}
We study the effect of increasing the mid-training token budget while keeping the context length fixed at 8k and using a Math+Code data mixture (Table~\ref{tab:token-budget-ablation}). All experiments use the Granite-4 Micro (3B) dense model.

Relative to the base model, mid-training yields large gains in both math and code with modest budgets. At 10.49B tokens, Math Avg increases from 26.28 to \textbf{40.21} (\textbf{+13.93}), while Code Avg improves from 1.26 to \textbf{9.59}. Increasing the budget to 15.73B further improves Math Avg to \textbf{42.07} while maintaining a strong Code Avg of \textbf{9.02}.

Beyond 26.21B tokens, gains largely saturate. Math Avg remains nearly constant (42.22 to \textbf{42.42}), while Code Avg declines from 8.69 to 7.62 as the budget increases to 31.46B. General-purpose leaderboard scores (LB V1 and V2) remain stable across budgets, and GPQA-Diamond shows no consistent trend. These results indicate that most benefits of Math+Code mid-training are realized within approximately 15B to 27B tokens for this model.

\section{Effects of Reinforcement Learning on Mid-Trained Models}
\begin{center}
\begin{tcolorbox}[
  colback=cyan!10!teal!8,
  colframe=teal!50!black,
  boxrule=0.4pt,
  arc=2mm,
  left=10pt, right=10pt, top=7pt, bottom=7pt,
  width=0.82\linewidth,
  fontupper=\small,
  before skip=6pt,
  after skip=10pt
]
\textbf{Takeaway.} The $\textsc{PRISM} \to \text{RL}$ pipeline improves the six-benchmark 
macro-average from under 12 to \textbf{29--42}, a 3--4$\times$ improvement. Mid-training 
contributes the dominant gains (\textbf{+14 to +18 points}), RL adds a consistent second 
stage (\textbf{+8 to +12 points}), and RL on base models without mid-training is 
substantially less effective, with AIME scores remaining near zero for most models 
(Nemotron-H being an exception, showing moderate AIME progress from base). Science 
data at mid-training unlocks large GPQA-Diamond gains during RL (\textbf{+17 to +28 
points} over MC-only), and RL progressively solves prompts that were initially unsolvable 
(shown for Granite-3.3).
\end{tcolorbox}
\end{center}

A central question for \textsc{PRISM} is whether mid-trained models provide a better foundation for reinforcement learning than base models, and if so, how the mid-training and RL data compositions interact. In this section we address both questions through controlled experiments across six model families, two RL data mixes (balanced and unbalanced), and direct comparisons with RL applied to base models.

%% ════════════════════════════════════════════════════════
\subsection{RL Setup: Data, Filtering, and Mixes}
\label{app:rl_data}

\begin{wraptable}{R}{0.45\textwidth}
\vspace{-14pt}
\centering
\scriptsize
\begin{tabularx}{0.44\textwidth}{
    >{\raggedright\arraybackslash}X
    >{\raggedright\arraybackslash}X
    >{\centering\arraybackslash}p{1.2cm}
}
\toprule
\textbf{Domain} & \textbf{Sources} & \textbf{Count} \\
\midrule
\multirow{3}{*}{Math}
& DeepScaleR-Preview
& \multirow{3}{*}{294K} \\
& INTELLECT-2-RL & \\
& Skywork-OR1-RL-Data & \\
\midrule
Science
& Nemotron-PT-v1-stem
& 100K \\
\midrule
\multirow{3}{*}{Code}
& DeepCoder-Preview
& \multirow{3}{*}{142K} \\
& Skywork-OR1-RL-Data & \\
& OpenCodeInstruct & \\
\bottomrule
\end{tabularx}
\caption{RL datasets and prompt counts.}
\label{tab:data-mixtures-rl}
\vspace{-6pt}
\end{wraptable}
Table~\ref{tab:data-mixtures-rl} summarizes the datasets used for RL across math, science, and code domains. We construct two RL data mixes, each subdivided into MC (math + code) and MCS (math + code + science) variants:

\paragraph{\textbf{Unbalanced mix.}}
\label{app:rl_data_filtering}
We use the Granite-3.3-8B mid-trained model to filter prompts by difficulty. For each prompt, we sample 16 responses (temperature 1.0, top\_p 1.0). For math, we select prompts with exactly one correct sample out of 16, yielding a hard subset of 19k prompts. For code and science, where most prompts are unsolvable, we retain all prompts with at least one correct sample, resulting in 7k code and 17k science prompts. Despite the domain imbalance, this mix produces strong improvements across all reasoning benchmarks.

\paragraph{\textbf{Balanced mix.}}
We equalize all domains to 19k prompts by augmenting code and science with a random subset of prompts having zero correct samples (out of 16) for the Granite-3.3-8B mid-trained model. We additionally apply randomized instruction-format templates to science prompts to increase format diversity. Note that some zero-score prompts may be solvable by other mid-trained models.

Training hyperparameters are consistent across model families. Algorithm details are provided in Appendix~\ref{app:rl_training_details}.

%% ════════════════════════════════════════════════════════
\subsection{RL on \textsc{PRISM}: Consistent Gains Across Models}

\begin{figure}[!htbp]
    \centering
    \begin{subfigure}[t]{\linewidth}
        \centering
        \includegraphics[width=\linewidth]{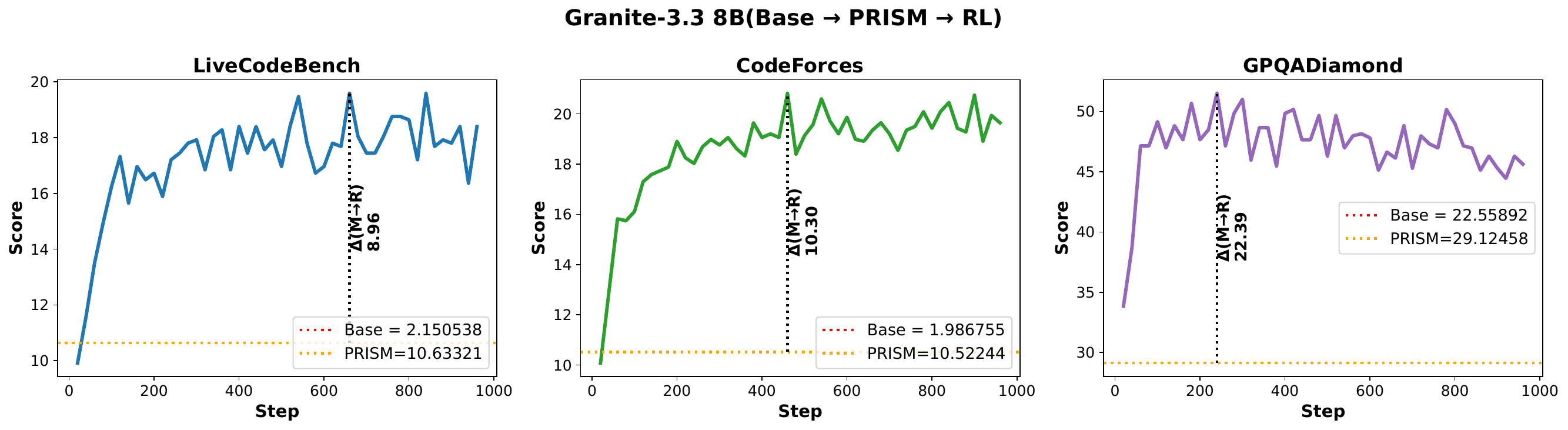}
        \caption{LiveCodeBench, Codeforces, and GPQA-Diamond over RL steps.}
        \label{fig:g33_code}
    \end{subfigure}
    \vspace{0.6em}
    \begin{subfigure}[t]{\linewidth}
        \centering
        \includegraphics[width=\linewidth]{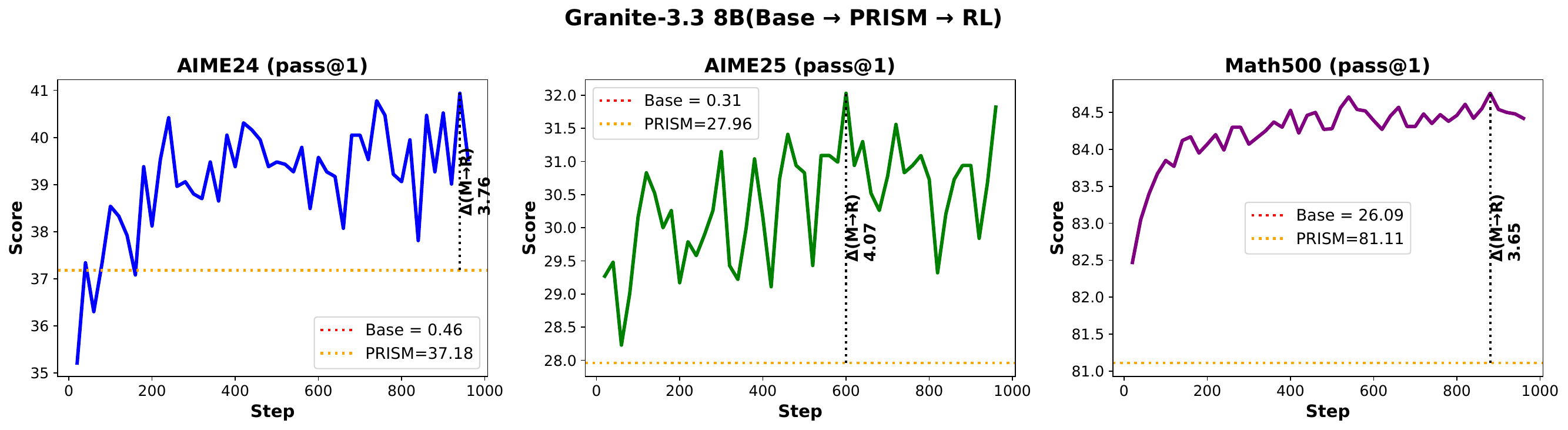}
        \caption{AIME24, AIME25, and MATH500 over RL steps.}
        \label{fig:g33_math}
    \end{subfigure}
    \caption{\textbf{$\textsc{PRISM} \to \text{RL}$: Granite-3.3-8B.} RL training curves on the \textsc{PRISM}-mid-trained checkpoint using the unbalanced MCS mix. All benchmarks show consistent, monotonic improvements.}
    \label{fig:g33_code_math}
\end{figure}

\begin{figure}[!htbp]
    \centering
    \begin{subfigure}[t]{\linewidth}
        \centering
        \includegraphics[width=\linewidth]{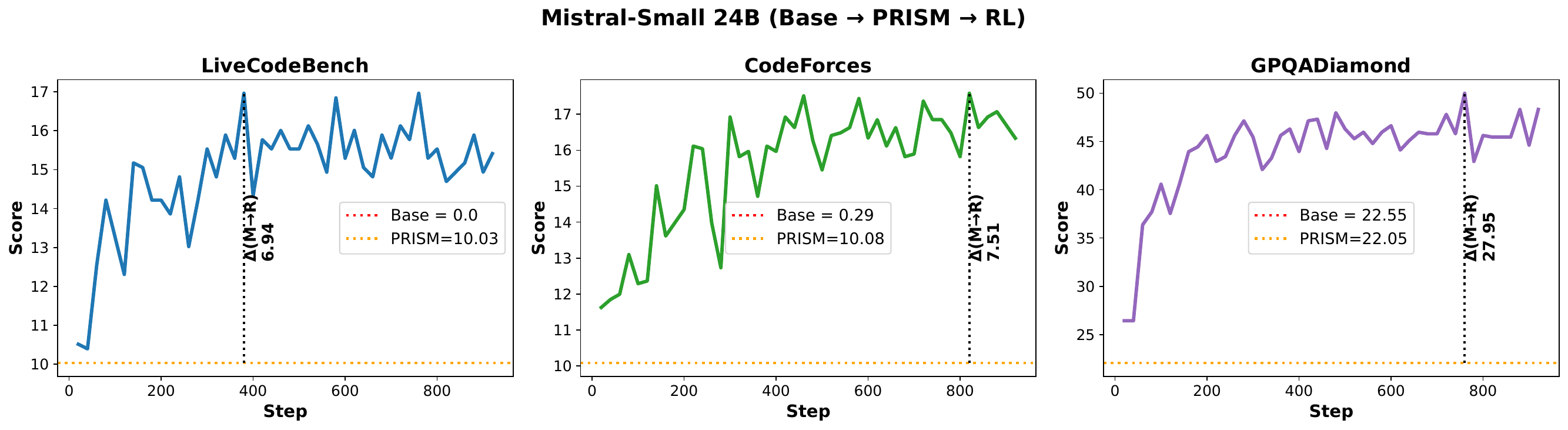}
        \caption{LiveCodeBench, Codeforces, and GPQA-Diamond over RL steps.}
        \label{fig:mistral24_code}
    \end{subfigure}
    \vspace{0.6em}
    \begin{subfigure}[t]{\linewidth}
        \centering
        \includegraphics[width=\linewidth]{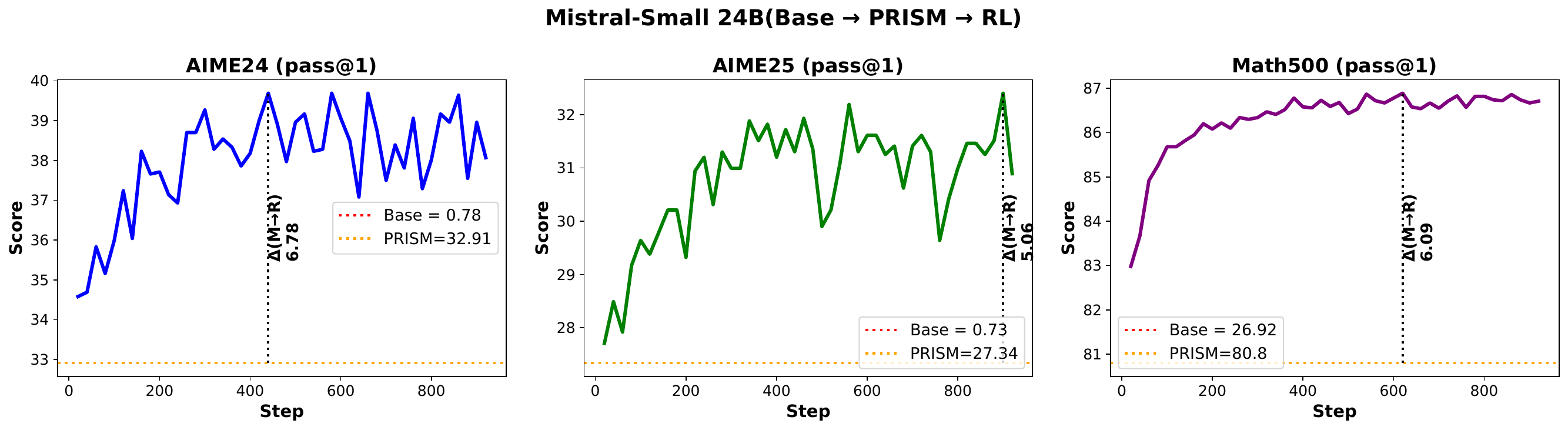}
        \caption{AIME24, AIME25, and MATH500 over RL steps.}
        \label{fig:mistral24_math}
    \end{subfigure}
    \caption{\textbf{$\textsc{PRISM} \to \text{RL}$: Mistral-Small 24B.} The largest model tested shows the strongest GPQA-Diamond gains (\textbf{+27.95}) and non-saturating code improvements.}
    \label{fig:mistral24_code_math}
\end{figure}

\begin{figure}[!htbp]
    \centering
    \begin{subfigure}[t]{\linewidth}
        \centering
        \includegraphics[width=\linewidth]{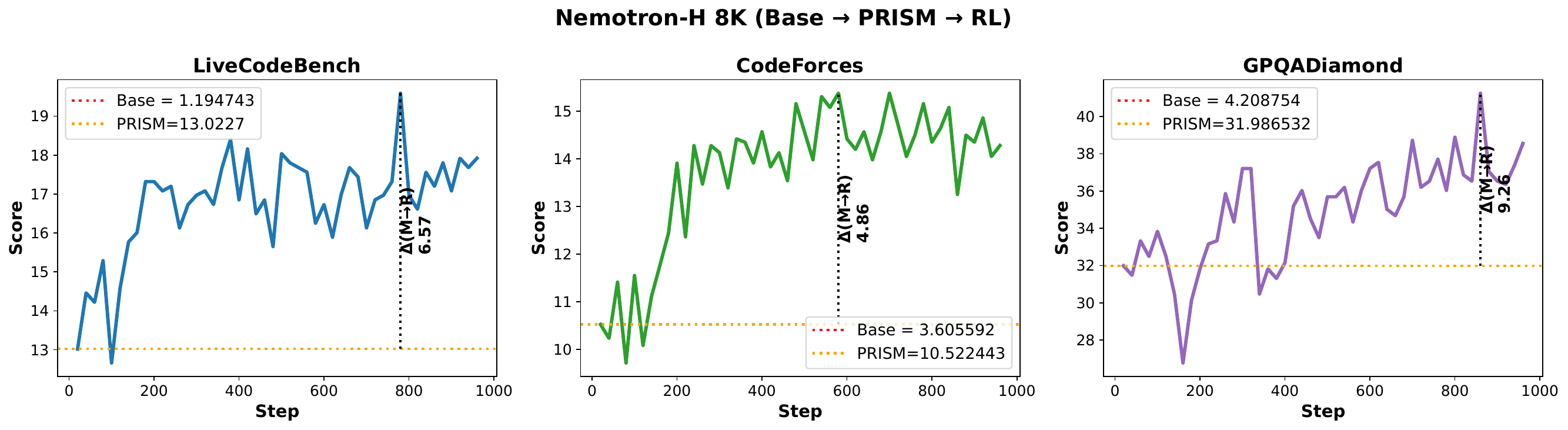}
        \caption{LiveCodeBench, Codeforces, and GPQA-Diamond over RL steps.}
        \label{fig:nemotronH_code}
    \end{subfigure}
    \vspace{0.6em}
    \begin{subfigure}[t]{\linewidth}
        \centering
        \includegraphics[width=\linewidth]{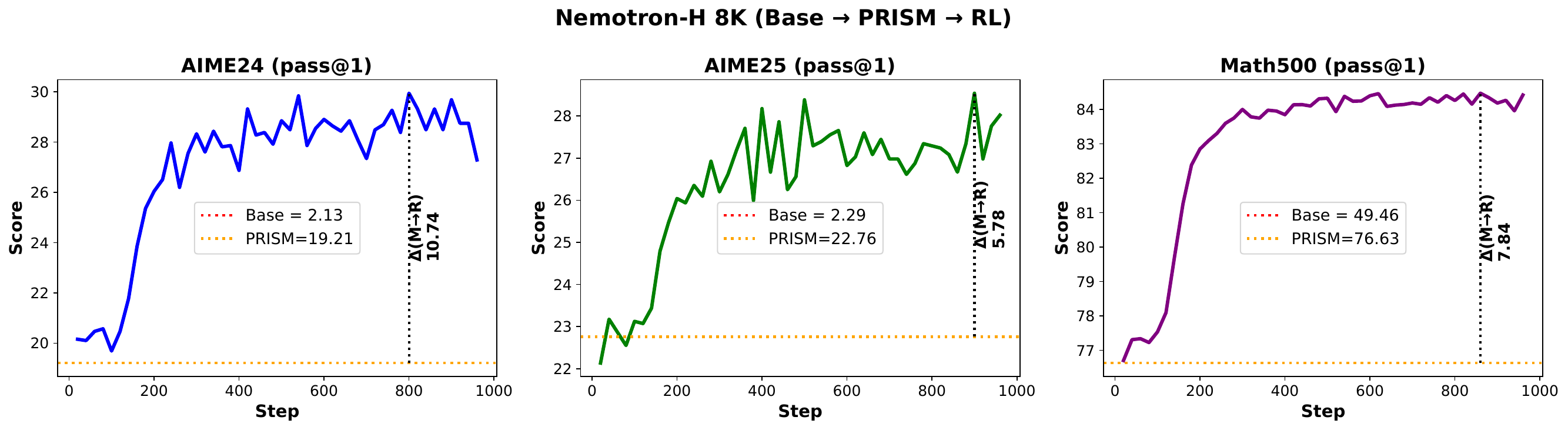}
        \caption{AIME24, AIME25, and MATH500 over RL steps.}
        \label{fig:nemotronH_math}
    \end{subfigure}
    \caption{\textbf{$\textsc{PRISM} \to \text{RL}$: Nemotron-H 8B (Hybrid).} RL yields stable gains on the hybrid attention-Mamba architecture, confirming that mid-training benefits extend beyond dense Transformers.}
    \label{fig:nemotron_H_math}
\end{figure}

We apply RL with the unbalanced MCS mix on top of \textsc{PRISM}-mid-trained models. Learning curves for Granite-3.3-8B, Mistral-Small 24B, and Nemotron-H (8B) are shown in Figs.~\ref{fig:g33_code_math}--\ref{fig:nemotron_H_math}; additional results for Mistral-7B, LLaMA-3.1-8B, and Granite-4 Micro (Dense, 3B) are provided in Appendix Figs.~\ref{fig:mistral_code_math},~\ref{fig:llama_code_math}, and~\ref{fig:g4micro_code_math}.

\subsubsection{Gains across benchmarks.}
RL on top of \textsc{PRISM} yields consistent, positive gains across nearly all benchmarks and model families. GPQA-Diamond shows the largest absolute improvements (e.g., Mistral-24B: \textbf{+27.95}, Granite-3.3: \textbf{+22.39}, Mistral-7B: \textbf{+19.19}, LLaMA: \textbf{+18.35}, Nemotron-H: \textbf{+9.26}). LiveCodeBench gains are substantial too (Granite-3.3: \textbf{+8.96}, Mistral-24B: \textbf{+6.94}, LLaMA: \textbf{+8.96}, Granite-4 Micro: \textbf{+5.62}, Mistral-7B: \textbf{+6.21}, Nemotron-H: \textbf{+6.57}), indicating improved code generation after $\textsc{PRISM} \to \text{RL}$ (see also Appendix~\ref{app:g33_rl_code}).
Codeforces improvements are more variable (\textbf{+2.65 to +10.30}), with Granite-3.3 showing the largest gain (\textbf{+10.30}). Math benchmark gains (AIME24/AIME25) are typically in the \textbf{3--10.74} point range across models. Granite-4 Micro (3B) shows consistent but smaller absolute gains compared with the larger 8B models.

\subsubsection{Non-saturating training curves.}
Across both code and math benchmarks, many RL curves continue to trend upward or exhibit oscillations around an improving mean rather than clean saturation. This is visible in LiveCodeBench, Codeforces, AIME24/25, and MATH500, where scores often keep improving late into training, suggesting that the $\textsc{PRISM} \to \text{RL}$ pipeline has not yet exhausted the available performance gains.
Several models show noticeable improvements well after hundreds of RL steps (e.g., Granite-3.3 on Codeforces and LiveCodeBench; Mistral-24B on Codeforces and MATH500). This strengthens the case for viewing \textsc{PRISM} not as a final training stage, but as a launch point for deeper RL or multi-stage RL pipelines.

\paragraph{Generalization to recently released held-out benchmark.}
To further validate generalization, we evaluate Granite-3.3~(8B) and 
Mistral-Small~(24B) on AIME~2026~\citep{maa2026aime}, which was published 
after the completion of all training runs. Both models show consistent 
improvement over RL training steps on this fully held-out benchmark 
(Appendix~\ref{app:aime26}), confirming that the gains from the 
\textsc{Prism} $\rightarrow$ RL pipeline transfer to unseen mathematical 
reasoning challenges.

%% ════════════════════════════════════════════════════════
\subsection{\textsc{PRISM} vs Base Models: Mid-Training is Essential for RL}

To quantify the value of mid-training as an initialization for RL, we apply RL directly to four base models: Granite-3.3 (8B), LLaMA-3.1 (8B), Mistral-7B, and Nemotron-H (8B), using the same unbalanced mix. Learning curves for Granite-3.3 and Nemotron-H are shown in Figs.~\ref{fig:granite3.3_code_math_base} and~\ref{fig:nemotronH_base}; LLaMA and Mistral-7B base RL curves are in Appendix Figs.~\ref{fig:llama_base} and~\ref{fig:mistral7_base}.

\begin{figure}[!htbp]
    \centering
    \begin{subfigure}[t]{\linewidth}
        \centering
        \includegraphics[width=\linewidth]{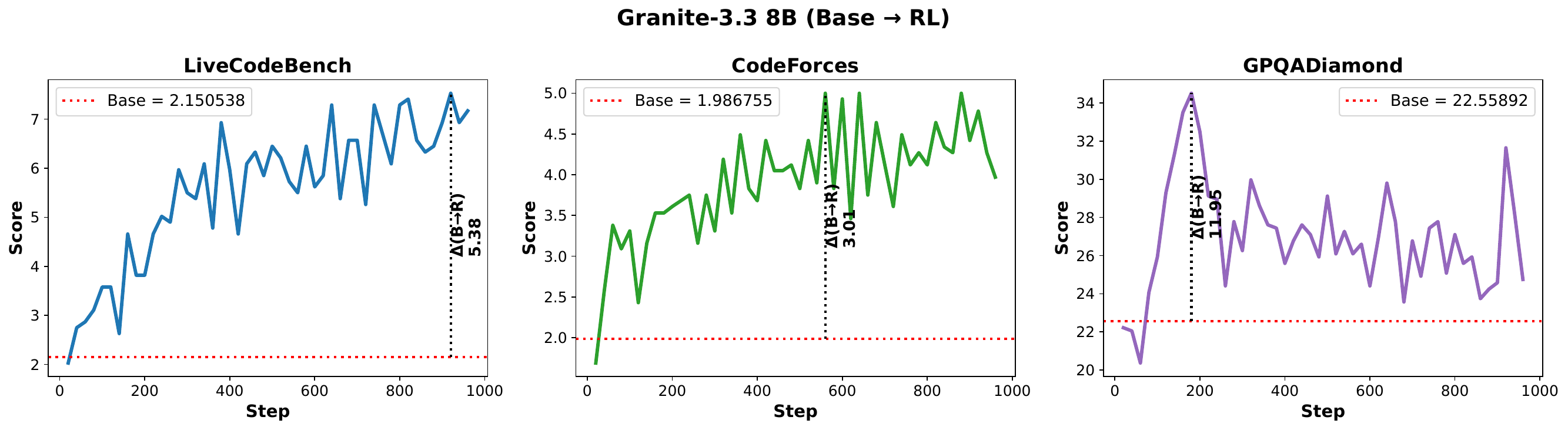}
        \caption{LiveCodeBench, Codeforces, and GPQA-Diamond over RL steps.}
        \label{fig:g33_code_base}
    \end{subfigure}
    \vspace{0.6em}
    \begin{subfigure}[t]{\linewidth}
        \centering
        \includegraphics[width=\linewidth]{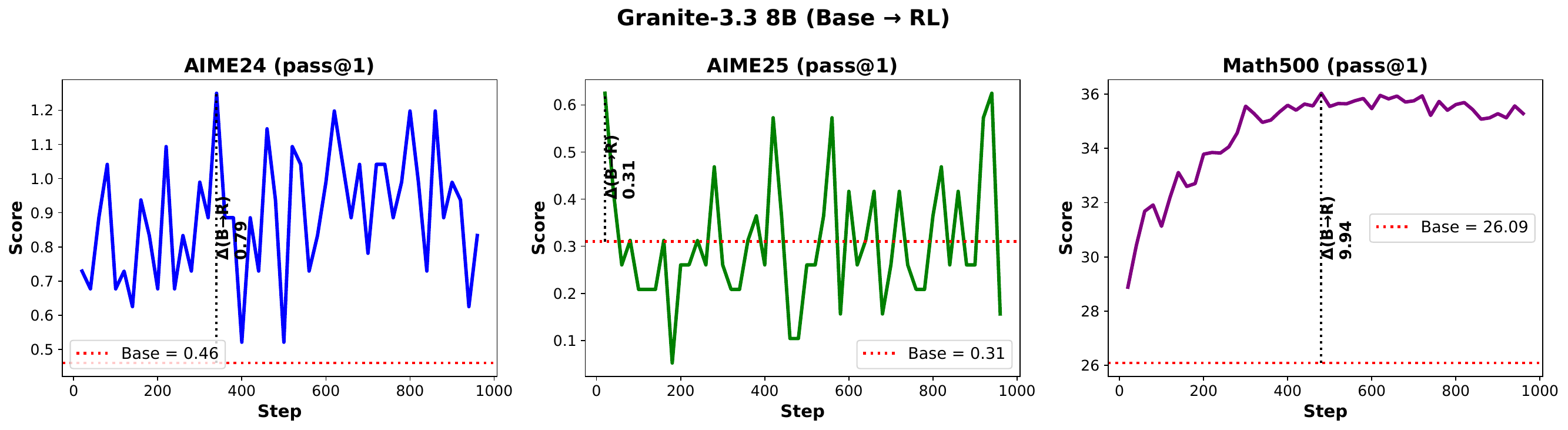}
        \caption{AIME24, AIME25, and MATH500 over RL steps.}
        \label{fig:g33_math_base}
    \end{subfigure}
    \caption{\textbf{RL on Granite-3.3-8B base (no mid-training).} AIME24/25 remain near zero throughout training, and overall gains are substantially smaller than the $\textsc{PRISM} \to \text{RL}$ pipeline (Fig.~\ref{fig:g33_code_math}).}
    \label{fig:granite3.3_code_math_base}
\end{figure}

\begin{figure}[!htbp]
    \centering
    \begin{subfigure}[t]{\linewidth}
        \centering
        \includegraphics[width=\linewidth]{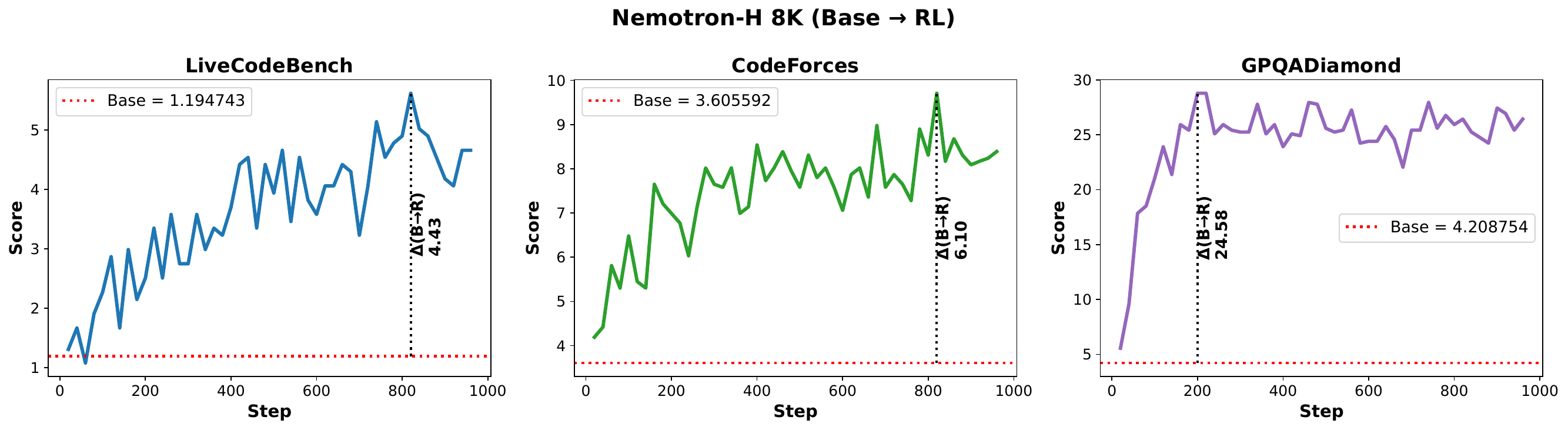}
        \caption{LiveCodeBench, Codeforces, and GPQA-Diamond over RL steps.}
    \end{subfigure}
    \vspace{0.6em}
    \begin{subfigure}[t]{\linewidth}
        \centering
        \includegraphics[width=\linewidth]{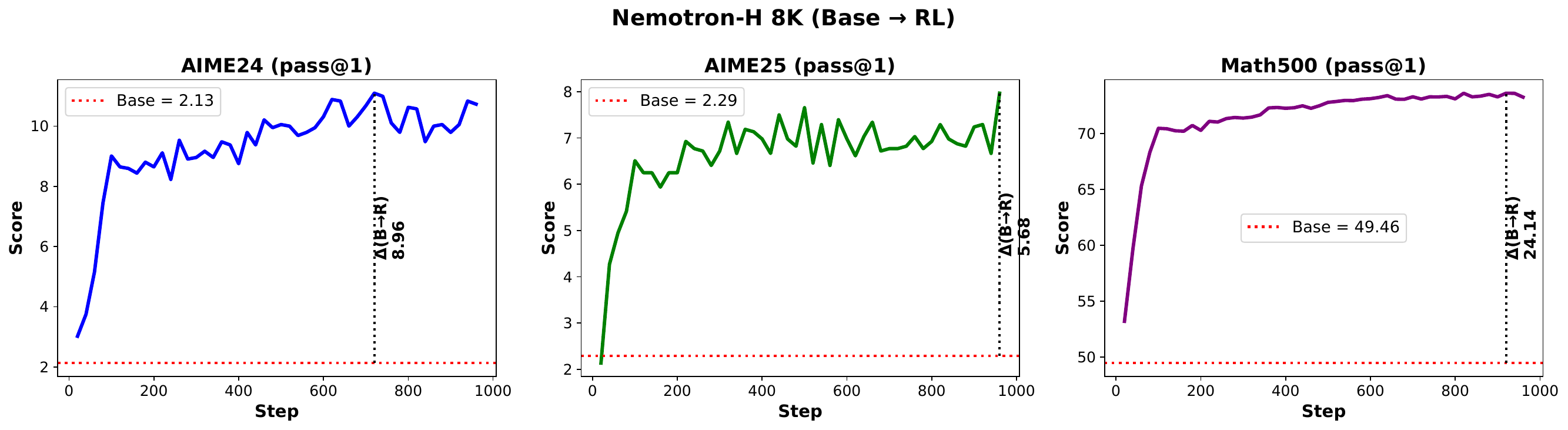}
        \caption{AIME24, AIME25, and MATH500 over RL steps.}
    \end{subfigure}
    \caption{\textbf{RL on Nemotron-H 8B base (no mid-training).} Even for hybrid architectures, RL on the base model shows limited progress on harder benchmarks compared to $\textsc{PRISM} \to \text{RL}$ (Fig.~\ref{fig:nemotron_H_math}).}
    \label{fig:nemotronH_base}
\end{figure}

\noindent\textbf{Granite-3.3 (8B).} Figure~\ref{fig:granite3.3_code_math_base} shows that RL on the base model produces noticeable gains on MATH500, coding, and science tasks, but fails to consistently improve on AIME24 and AIME25. Overall, RL on the base model underperforms RL on \textsc{PRISM} by a large margin, with final scores lower by \textbf{$\sim$37 points} in math, \textbf{$\sim$14 points} in code, and \textbf{$\sim$5 points} in science.

\noindent\textbf{LLaMA-3.1 (8B) and Mistral-7B.} Both models exhibit a similar pattern when RL is applied directly to their base checkpoints (Figs.~\ref{fig:llama_base} and~\ref{fig:mistral7_base} in Appendix): MATH500 and Coding benchmarks show modest gains, but AIME24 and AIME25 remain near zero throughout training, indicating that base models lack the foundational reasoning representations needed for RL to make progress on harder tasks. We see a regression in GPQA-Diamond performance, where RL on top of the base model leads to lower performance than the base model itself. In contrast, RL on the corresponding \textsc{PRISM}-mid-trained checkpoints achieves substantially higher scores across all benchmarks (Figs.~\ref{fig:llama_code_math} and~\ref{fig:mistral_code_math}).

\noindent\textbf{Nemotron-H (8B).} Nemotron-H base (Fig.~\ref{fig:nemotronH_base}) shows 
a slightly different pattern: RL produces some gains on MATH500 and moderate AIME24/25 
progress from base, unlike most other models where AIME scores remain near zero. This may 
be attributed to stronger mathematical knowledge in Nemotron-H's pretraining data, which 
provides a better initialization for RL even without mid-training. Nonetheless, the gap 
compared to the \textsc{PRISM} RL results (Fig.~\ref{fig:nemotron_H_math}) remains 
substantial, confirming that mid-training is critical even for hybrid architectures.

Across all four model families, a consistent conclusion emerges: RL on base models produces limited and often unstable improvements, particularly on harder benchmarks like AIME24/25, while RL on \textsc{PRISM}-mid-trained models yields large, stable, and monotonic gains. These results are consistent with prior findings \citep{wang2025octothinkermidtrainingincentivizesreinforcement, zhang2025interplaypretrainingmidtrainingrl} and highlight that \textsc{PRISM} provides a substantially stronger initialization for RL-driven reasoning expansion.

%% ════════════════════════════════════════════════════════
\subsection{Balanced vs Unbalanced RL Mix}

We next study whether equalizing prompt counts across domains affects RL outcomes. We apply RL with the balanced mix on top of \textsc{PRISM} for Mistral-Small 24B, Granite-4 Micro (Hybrid and Dense, 3B), and Granite-3.3 (8B). Learning curves for Granite-3.3 are shown in Fig.~\ref{fig:balanced_g33}; results for the remaining models are in Appendix Figs.~\ref{fig:balanced_mistral24}--\ref{fig:balanced_g4H}.

\begin{figure}[!htbp]
    \centering
    \begin{subfigure}[t]{\linewidth}
        \centering
        \includegraphics[width=\linewidth]{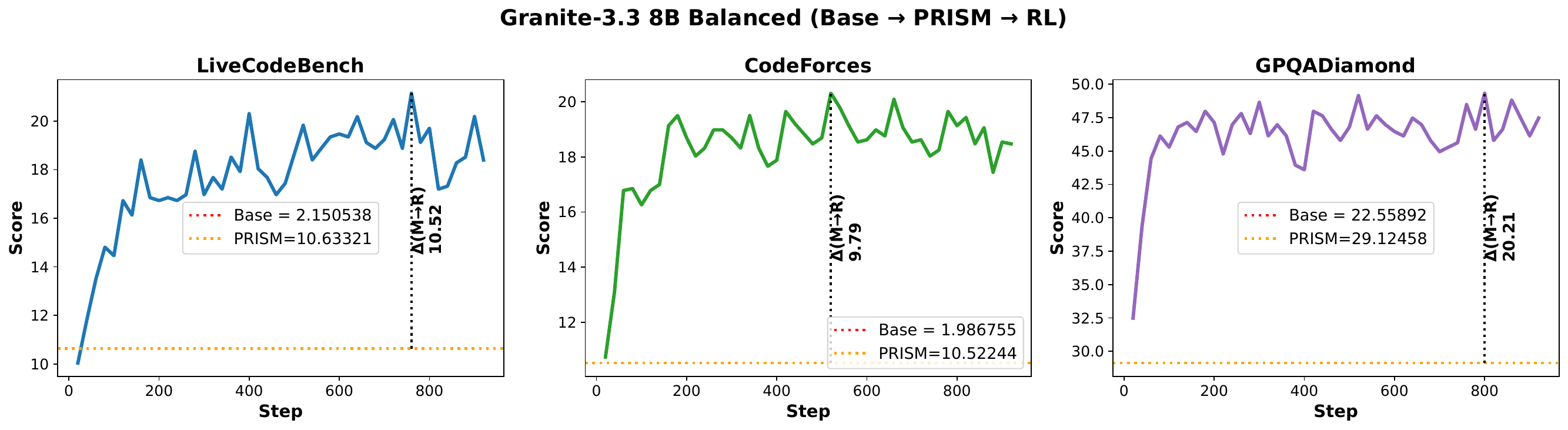}
        \caption{LiveCodeBench, Codeforces, and GPQA-Diamond over RL steps.}
    \end{subfigure}
    \vspace{0.6em}
    \begin{subfigure}[t]{\linewidth}
        \centering
        \includegraphics[width=\linewidth]{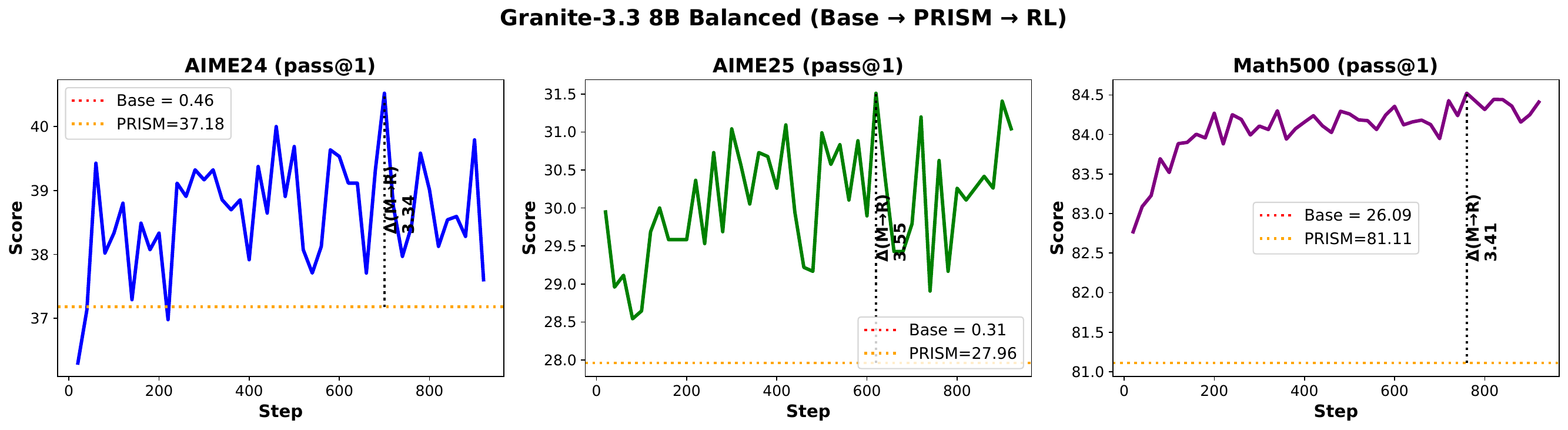}
        \caption{AIME24, AIME25, and MATH500 over RL steps.}
    \end{subfigure}
    \caption{\textbf{$\textsc{PRISM} \to \text{RL}$ with balanced mix: Granite-3.3-8B.} Domain-equalized RL produces comparable math and code gains to the unbalanced mix (Fig.~\ref{fig:g33_code_math}), with stable training throughout.}
    \label{fig:balanced_g33}
\end{figure}

Across all four models, RL with the balanced mix produces consistent improvements over \textsc{PRISM} on both math and code benchmarks. On the dense Granite-4 Micro (3B), the balanced mix yields gains of \textbf{+4.63} on AIME24, \textbf{+3.07} on AIME25, and \textbf{+3.38} on MATH500, with code improvements of \textbf{+4.30} on LiveCodeBench and \textbf{+6.06} on GPQA-Diamond (Fig.~\ref{fig:balanced_g4dense}). The hybrid Granite-4-H Micro (3B) shows even larger gains, particularly on Codeforces (\textbf{+8.09}) and GPQA-Diamond (\textbf{+11.95}), with math improvements of \textbf{+5.58} on AIME24 and \textbf{+6.41} on AIME25 (Fig.~\ref{fig:balanced_g4H}).

Mistral-Small 24B also shows steady improvements on math and code benchmarks under the balanced mix (Fig.~\ref{fig:balanced_mistral24}), though its GPQA-Diamond gain (\textbf{+25.93}) is slightly lower than that achieved by the unbalanced mix (\textbf{+27.95}, Fig.~\ref{fig:mistral24_code_math}). Granite-3.3 (8B) benefits consistently from the balanced mix (Fig.~\ref{fig:balanced_g33}), with improvements across all benchmarks.

Comparing with the unbalanced mix results (Figs.~\ref{fig:g33_code_math}--\ref{fig:g4micro_code_math}), we observe that math and code gains are broadly comparable across both mixes: for instance, the unbalanced mix on Granite-3.3 yields LiveCodeBench \textbf{+8.96} and GPQA-Diamond \textbf{+22.39} (Fig.~\ref{fig:g33_code_math}), while the balanced mix on the same model produces similar trajectories (Fig.~\ref{fig:balanced_g33}), showing that the balanced mix achieves comparable math and code gains to the unbalanced mix. For science, the effect of the balanced mix is model-dependent: the Granite-4 Micro variants show stronger GPQA-Diamond gains under the balanced mix, while Mistral-Small 24B performs slightly better with the unbalanced mix. We attribute the science improvements observed with the balanced mix primarily to the use of randomized instruction-format templates applied to science prompts, which expose the model to diverse question phrasings during RL and improve robustness to prompt formatting on GPQA-Diamond. Across all models, training curves under the balanced mix remain stable and monotonically improving, with no training instabilities observed.

%% ════════════════════════════════════════════════════════
\subsection{RL Expands the Solvability Frontier}

\begin{wrapfigure}{R}{0.48\textwidth}
\vspace{-14pt}
    \centering
    \includegraphics[width=0.46\textwidth]{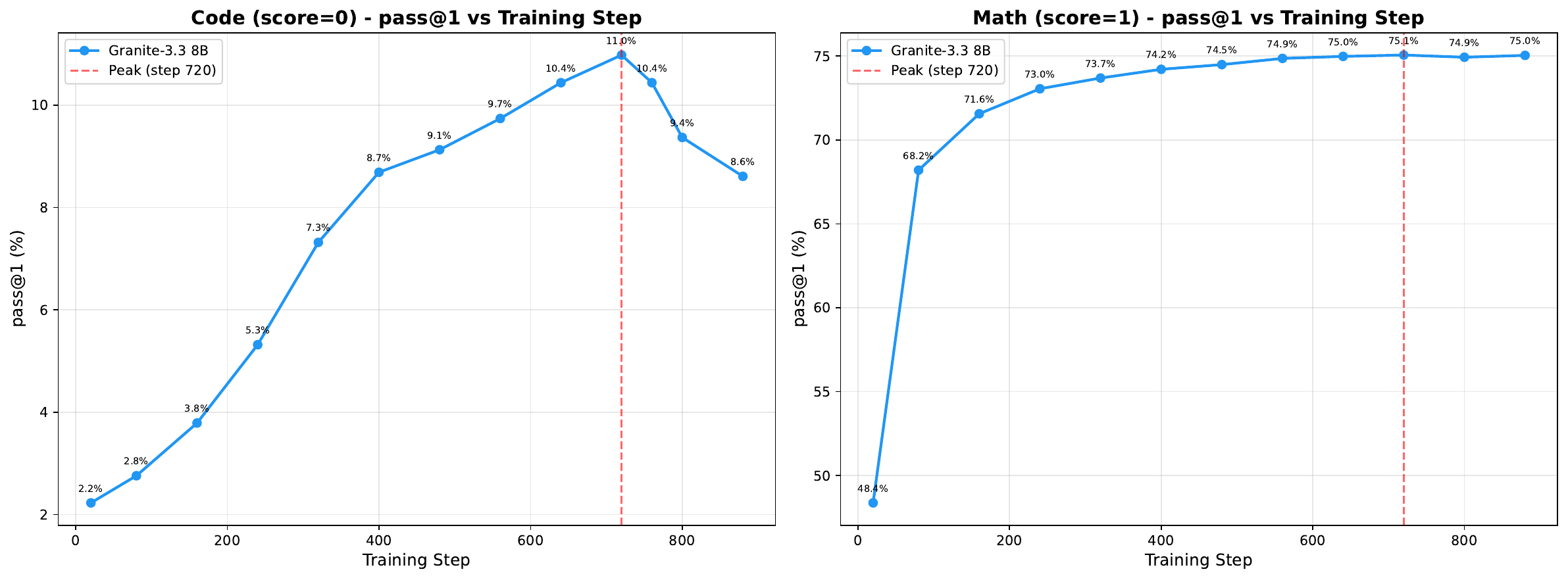}
    \caption{Pass rates on initially unsolved (code, score\,=\,0) and hardest (math, score\,=\,1) prompts during RL training of Granite-3.3 (8B) with the balanced mix.}
    \label{fig:solvability_g33}
\vspace{-6pt}
\end{wrapfigure}
A natural question is whether RL merely refines performance on already-solvable problems or actively expands the frontier of what the model can solve. Recall that the balanced mix includes prompts with zero correct samples out of 16 (score\,=\,0) for code, and prompts with exactly one correct sample (score\,=\,1) for math, representing the hardest tier of each domain. We track the pass rate of these prompts throughout RL training on Granite-3.3 (8B).

Figure~\ref{fig:solvability_g33} shows that the model progressively learns to solve prompts it could not handle at the start of RL. For code prompts that had a pass rate of zero under the mid-trained checkpoint, the pass rate steadily increases over training, indicating that RL enables the model to acquire new problem-solving strategies beyond what mid-training alone provides. Similarly, for the hardest math prompts (score\,=\,1), the pass rate improves consistently, showing that RL amplifies the model's ability to solve problems at the boundary of its initial competence.

These results, combined with the non-saturating training curves observed above, provide evidence that the $\textsc{PRISM} \to \text{RL}$ pipeline actively pushes the solvability boundary rather than merely polishing existing capabilities. This is consistent with recent findings by \citet{sun2025rlgrokkingrecipedoes}, who show that RL can unlock genuinely new algorithmic strategies in LLMs for previously unsolvable problem families. Mid-training produces a representation that is well-suited for RL-driven capability expansion.

%% ════════════════════════════════════════════════════════
\subsection{The Full Pipeline: Broader RL Analysis}

\begin{table*}[!htbp]
\centering
\scriptsize
\setlength{\tabcolsep}{4pt}
\resizebox{\textwidth}{!}{
\begin{tabular}{l l r r r r r r r r r r r}
\toprule
\textbf{Model}
& \textbf{MT}
& \textbf{RL}
& \textbf{LCB}
& \textbf{CF}
& \textbf{Code Avg}
& \textbf{AIME24}
& \textbf{AIME25}
& \textbf{MATH500}
& \textbf{Math Avg}
& \textbf{GPQA}
& \textbf{AVG} \\
\midrule
\multirow{7}{*}{LLaMA-3.1}
& -- & -- & 0.00 & 0.07 & 0.04 & 0.05 & 0.15 & 6.51 & 2.24 & 20.20 & 7.49 \\
& MC & -- & 6.93 & 6.03 & 6.48 & 20.67 & 19.58 & 73.70 & 37.98 & 19.53 & 21.33 \\
& MCS & -- & 6.09 & 5.45 & 5.77 & 16.45 & 19.32 & 73.47 & 36.41 & 21.04 & 21.07 \\
& MC & MC & 12.31 & 11.85 & 12.08 & 25.47 & 23.23 & 78.99 & 42.56 & 23.06 & 25.90 \\
& MC & MCS & 11.83 & 12.80 & 12.32 & 24.43 & 23.12 & 78.62 & 42.06 & 24.75 & 26.38 \\
& MCS & MC & 13.62 & 11.41 & 12.51 & 20.47 & 21.67 & 77.10 & 39.75 & 34.01 & 28.76 \\
\rowcolor{teal!10}
& MCS & MCS & 14.34 & 12.07 & 13.20 & 20.42 & 22.08 & 77.03 & 39.84 & 36.03 & 29.69 \\
\midrule
\multirow{7}{*}{Granite-3.3}
& -- & -- & 2.15 & 1.99 & 2.07 & 0.46 & 0.31 & 26.09 & 8.95 & 22.56 & 11.19 \\
& MC & -- & 11.11 & 10.30 & 10.71 & 32.44 & 28.33 & 74.22 & 44.99 & 19.02 & 24.91 \\
& MCS & -- & 10.63 & 10.52 & 10.58 & 37.18 & 27.96 & 81.11 & 48.75 & 29.12 & 29.48 \\
& MC & MC & 20.79 & 18.76 & 19.78 & 40.36 & 33.33 & 85.88 & 53.19 & 35.52 & 36.16 \\
& MC & MCS & 20.43 & 19.57 & 20.00 & 40.10 & 30.89 & 85.51 & 52.17 & 35.69 & 35.95 \\
\rowcolor{teal!10}
& MCS & MC & 20.31 & 20.46 & 20.38 & 40.62 & 30.89 & 84.62 & 52.04 & 52.86 & 41.76 \\
& MCS & MCS & 17.20 & 18.03 & 17.62 & 40.42 & 29.58 & 83.99 & 51.33 & 51.52 & 40.16 \\
\midrule
\multirow{7}{*}{Mistral-7B}
& -- & -- & 0.00 & 0.15 & 0.07 & 0.00 & 0.10 & 1.68 & 0.59 & 26.94 & 9.20\\
& MC & -- & 11.11 & 9.27 & 10.19 & 24.63 & 15.52 & 47.70 & 29.28 & 15.99 & 18.49\\
& MCS & -- & 10.16 & 9.42 & 9.79 & 28.85 & 24.27 & 70.71 & 41.28 & 24.07 & 25.05 \\
& MC & MC & 17.08 & 16.34 & 16.71 & 34.11 & 27.50 & 84.18 & 48.60 & 29.12 & 31.48\\
& MC & MCS & 16.61 & 15.60 & 16.10 & 33.02 & 26.93 & 83.80 & 47.92 & 28.28 & 30.77\\
& MCS & MC & 16.61 & 15.31 & 15.96 & 33.75 & 26.93 & 84.15 & 48.28 & 40.91 & 35.05\\
\rowcolor{teal!10}
& MCS & MCS & 16.01 & 15.16 & 15.58 & 32.86 & 27.03 & 84.37 & 48.09 & 41.75 & 35.14\\
\bottomrule
\end{tabular}}
\caption{Full $\text{Base} \to \text{Mid-training} \to \text{RL}$ pipeline results across LLaMA-3.1-8B, Granite-3.3-8B, and Mistral-7B. MC = math + code mix; MCS = math + code + science mix. MT = mid-training mix; RL = RL mix. Highlighted rows show the best configuration per model.}
\label{tab:midtrain-rl-math-code-science}
\end{table*}

Table~\ref{tab:midtrain-rl-math-code-science} presents a comprehensive view of the full $\text{Base} \to \text{Mid-training} \to \text{RL}$ pipeline across three model families, two mid-training mixes (MC and MCS), and two RL mixes (MC and MCS). Each row reports the best-step checkpoint for the corresponding configuration.

\subsubsection{A clear hierarchy: mid-training dominates, RL amplifies.}
The most striking pattern in Table~\ref{tab:midtrain-rl-math-code-science} is the consistent hierarchy of effect sizes across all three model families. Mid-training produces the largest single-stage jump: the six-benchmark macro-average (AVG) increases by \textbf{+13.84} for LLaMA (7.49 $\to$ 21.33), \textbf{+18.29} for Granite-3.3 (11.19 $\to$ 29.48), and \textbf{+15.85} for Mistral (9.20 $\to$ 25.05). RL then adds a consistent second-stage boost on top of these already-strong checkpoints: \textbf{+8.36} for LLaMA (21.33 $\to$ 29.69), \textbf{+12.28} for Granite-3.3 (29.48 $\to$ 41.76), and \textbf{+10.09} for Mistral (25.05 $\to$ 35.14). The combined $\textsc{PRISM} \to \text{RL}$ pipeline improves AVG from under 12 to \textbf{29--42}, a $3\text{--}4\times$ improvement.

\subsubsection{Science data at mid-training unlocks large RL gains on GPQA.}
One of the most impactful findings is that including science data during mid-training (MCS) dramatically amplifies GPQA-Diamond gains during RL. For Granite-3.3, MCS mid-training followed by MC RL achieves GPQA \textbf{52.86} (vs.\ 35.52 with MC mid-training + MC RL). The pattern is consistent: for LLaMA, MCS+MCS reaches GPQA \textbf{36.03} (vs.\ 23.06 for MC+MC), and for Mistral, MCS+MCS reaches \textbf{41.75} (vs.\ 29.12 for MC+MC). This suggests that science data during mid-training provides foundational representations that RL can leverage for scientific reasoning, even when the RL mix itself is not science-heavy.

\subsubsection{RL data mix matters less than mid-training mix.}
Changing the RL mix from MC to MCS produces comparatively small differences (typically $<$2 AVG points), whereas changing the mid-training mix from MC to MCS can shift AVG by \textbf{+3 to +6 points}. For example, for Granite-3.3 with MC mid-training, switching RL from MC to MCS changes AVG only from 36.16 to 35.95 ($-$0.21), while switching mid-training from MC to MCS (with MC RL) jumps AVG from 36.16 to 41.76 (\textbf{+5.60}). This confirms that data composition choices have their greatest impact during mid-training, and RL primarily serves to amplify whatever capabilities mid-training has established.

\subsubsection{Best configurations per model.}
The highlighted rows in Table~\ref{tab:midtrain-rl-math-code-science} show the best overall configuration for each family: \textbf{MCS mid-training + MCS RL} for LLaMA (AVG 29.69) and Mistral (AVG 35.14), and \textbf{MCS mid-training + MC RL} for Granite-3.3 (AVG 41.76). Granite-3.3 achieves the highest absolute scores across the board, with Code Avg of \textbf{20.38}, Math Avg of \textbf{52.04}, and GPQA of \textbf{52.86}, demonstrating that the $\textsc{PRISM} \to \text{RL}$ pipeline is most effective when built on a strong base model with broad mid-training coverage.

% Weight and entropy analyses moved to Section~\ref{sec:analysis} (text/analysis.tex)

\section{Understanding the PRISM Pipeline: Weight and Behavioral Analysis}
\label{sec:analysis}

\begin{center}
\begin{tcolorbox}[
  colback=cyan!10!teal!8,
  colframe=teal!50!black,
  boxrule=0.4pt,
  arc=2mm,
  left=10pt, right=10pt, top=7pt, bottom=7pt,
  width=0.82\linewidth,
  fontupper=\small,
  before skip=6pt,
  after skip=10pt
]
\textbf{Takeaway.} Mid-training makes broad weight changes and reshapes model behavior; RL makes targeted refinements while preserving representational structure.
\begin{itemize}[leftmargin=1.2em, itemsep=2pt, topsep=2pt, parsep=0pt]
\item \textbf{Weights:} Mid-training densely restructures $>$90\% of parameters; RL sparsely refines $\sim$5\%, with 370--580$\times$ smaller magnitude. This dense/sparse asymmetry holds at any threshold from 0.1\% to 10\%.
\item \textbf{Representations:} RL consistently preserves mid-training's representational geometry (CKA $>$ 0.998) across 3 models and 3 input distributions. Mid-training's representational impact is model-specific and cannot be universally characterized.
\item \textbf{Starting-point invariance:} RL targets the same sub-components in identical proportions whether or not mid-training preceded it, yet only succeeds on mid-trained models.
\item \textbf{Behavior:} Mid-training produces extended reasoning chains in model outputs. On held-out MATH500 problems, the full pipeline improves pass rates from 2.6--66.6\% (base) to 64.6--83.0\% (PRISM$\to$RL) across three model families.
\item \textbf{RL dynamics:} Optimization is front-loaded ($\sim$200--400 steps), with the active parameter set growing progressively from $\sim$1.5\% to $\sim$5\%.
\end{itemize}
\end{tcolorbox}
\end{center}

The preceding sections establish \emph{what} mid-training and RL achieve in terms of benchmark performance. In this section, we investigate \emph{how} these stages differ mechanistically, through four complementary lenses: (i) weight-level divergence and sparsity, (ii) representation similarity via CKA, (iii) prediction entropy and correctness, and (iv) RL weight trajectory dynamics. Weight and trajectory analyses use Granite-3.3 (dense) and Nemotron-H (attention-Mamba hybrid); CKA analysis additionally includes LLaMA-3.1 across three input distributions; and behavioral analyses include LLaMA-3.1.

%% ════════════════════════════════════════════════════════
\subsection{Weight-Level Analysis: Dense Restructuring vs.\ Sparse Refinement}
\label{sec:weight-analysis}

\begin{figure*}[t]
    \centering
    \includegraphics[width=\textwidth]{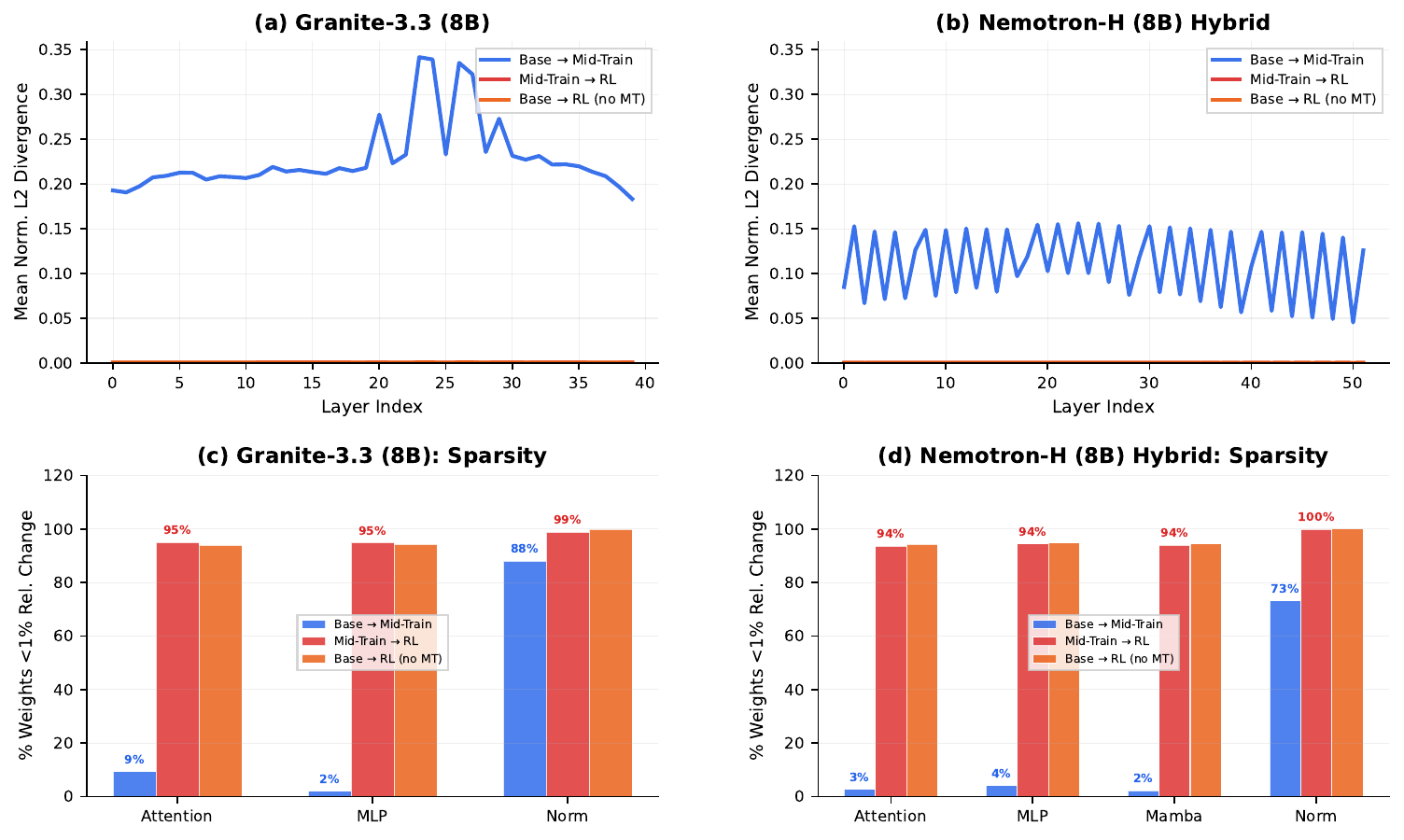}
    \caption{\textbf{Mid-training densely restructures the network; RL makes sparse, surgical refinements.} Top row: layer-wise normalized L2 divergence for Granite-3.3 (8B, left) and Nemotron-H (8B, right). Mid-training (blue) changes weights 370--580$\times$ more than RL (red, orange), broadly across all layers with some layer-wise variation. For Nemotron-H, the repeating pattern reflects its hybrid architecture where Mamba-2, self-attention and FFN are separate sequential layers with independent residual connections~\citep{nvidia2025nemotronhfamilyaccurateefficient}. Bottom row: update sparsity by component type. Mid-training modifies $>$90\% of all parameters (attention, MLP, and Mamba alike), while RL leaves $>$93\% unchanged.}
    \label{fig:weight-divergence}
\end{figure*}

We compute per-layer normalized L2 divergence and update sparsity across pipeline transitions. The normalized L2 divergence for a weight matrix $W$ is:
\begin{equation}
\delta(W) = \frac{\|W_{\text{new}} - W_{\text{old}}\|_2}{\|W_{\text{old}}\|_2}
\label{eq:norm-l2}
\end{equation}
Update sparsity is the fraction of parameters with $\delta < 1\%$ (Eq.~\ref{eq:norm-l2}); this threshold is illustrative and the dense/sparse asymmetry holds at any threshold from 0.1\% to 10\% (see Appendix~\ref{app:threshold-sensitivity}). For Granite-3.3, we additionally compare MC and MCS mid-training mixtures. Results are shown in Figure~\ref{fig:weight-divergence} and Table~\ref{tab:weight-divergence}.

\begin{table}[t]
\centering
\small
\setlength{\tabcolsep}{2.5pt}
\begin{tabular}{l l l c c c c c}
\toprule
\textbf{Model} & \textbf{MT} & \textbf{Transition} & \textbf{Attn} & \textbf{MLP} & \textbf{Mamba} & \textbf{Total} & \textbf{Sparsity} \\
\midrule
\multirow{5}{*}{\shortstack[l]{Granite-3.3\\(8B)}}
& \multirow{3}{*}{MCS}
& Base $\to$ MT          & 0.175 & 0.329 & -- & 0.175 & 9.3\% \\
& & MT $\to$ RL            & 0.0003 & 0.0006 & -- & 0.0003 & 95.9\% \\
& & Base $\to$ RL (no MT)  & 0.0004 & 0.0007 & -- & 0.0004 & 96.0\% \\
\cmidrule(l){2-8}
& \multirow{2}{*}{MC}
& Base $\to$ MT          & 0.177 & 0.333 & -- & 0.177 & 9.3\% \\
& & MT $\to$ RL            & 0.0003 & 0.0006 & -- & 0.0003 & 95.8\% \\
\midrule
\multirow{3}{*}{\shortstack[l]{Nemotron-H\\(8B, Hybrid)}}
& \multirow{3}{*}{MCS}
& Base $\to$ MT          & 0.230 & 0.289 & 0.138 & 0.112 & 2.7\% \\
& & MT $\to$ RL            & 0.0007 & 0.0007 & 0.0003 & 0.0003 & 93.5\% \\
& & Base $\to$ RL (no MT)  & 0.0006 & 0.0006 & 0.0003 & 0.0002 & 94.2\% \\
\bottomrule
\end{tabular}
\caption{\textbf{Weight divergence summary across models and architectures.} Normalized L2 = $\|w_{\text{new}} - w_{\text{old}}\|_2 / \|w_{\text{old}}\|_2$. Nemotron-H reports all three component types (Attention, MLP, Mamba). Sparsity = fraction of parameters with $<$1\% relative change. The dense/sparse asymmetry is consistent across all component types and architectures.}
\label{tab:weight-divergence}
\end{table}

\paragraph{\textbf{Mid-training is a dense, global restructuring.}}
Mid-training modifies the vast majority of parameters across all component types. For Granite-3.3, 90.7\% of attention and 98.1\% of MLP parameters change significantly during mid-training. For Nemotron-H, all three component types undergo dense updates: attention (97.3\%), MLP (95.9\%), and Mamba (97.8\%), with MLP showing the largest L2 divergence (0.289) followed by attention (0.230) and Mamba (0.138) (Table~\ref{tab:weight-divergence}). Changes are broadly distributed across all layers with some layer-wise variation (Figure~\ref{fig:weight-divergence}, top row), with the hybrid model showing a characteristic alternating pattern reflecting its architecture of separate Mamba-2, FFN, and attention layers (52 layers total: $\sim$24 Mamba, $\sim$24 FFN, 4 attention).

\paragraph{\textbf{RL is a sparse, surgical refinement.}}
In contrast, RL modifies only \textbf{$\sim$5\%} of parameters across all architectures. L2 divergence is 580$\times$ smaller for Granite-3.3 (0.0003 vs.\ 0.175) and 370$\times$ smaller for Nemotron-H (0.0003 vs.\ 0.112). Over 93\% of all weights remain within 1\% of their mid-trained values (Figure~\ref{fig:weight-divergence}, bottom row). Crucially, all three component types in the hybrid model show nearly identical sparsity during RL: attention (93.5\%), MLP (94.5\%), and Mamba (93.9\%), confirming that the sparse RL update pattern is consistent across component types within the hybrid architecture. This sparsity is consistent with concurrent findings by \citet{mukherjee2025reinforcementlearningfinetunessmall}, who identify in-distribution training as a key driver of update sparsity. We extend their analysis by demonstrating this asymmetry across two architectures and jointly with mid-training. We leave exploration of RL on domains not seen during mid-training to future work. At the sub-component level, value (V) and output (O) projections are consistently the most modified during RL (5.6--8.5\%), while SSM parameters (A, dt) remain completely frozen; see Appendix~\ref{app:subcomponent} for the full breakdown.

\begin{figure}[t]
    \centering
    \includegraphics[width=\linewidth]{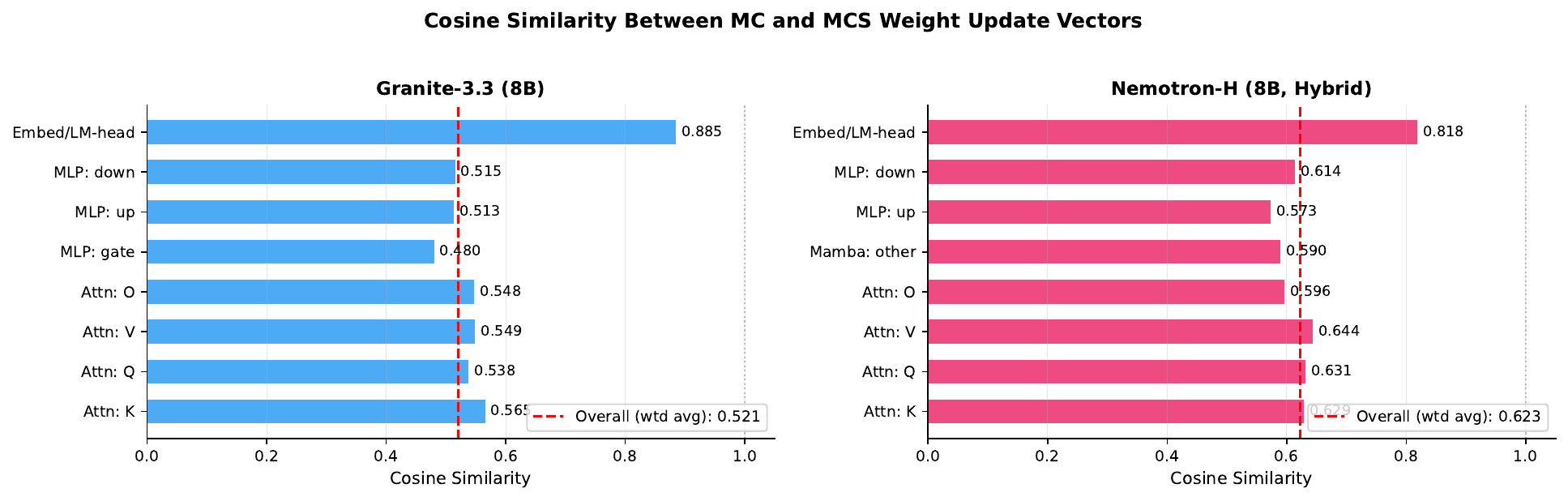}
    \caption{\textbf{Data composition redirects weight updates across all sub-components.} Cosine similarity between MC and MCS weight update vectors ($\Delta W = W_{MT} - W_{base}$) for Granite-3.3 (left) and Nemotron-H (right). Overall cosine similarity of 0.52 and 0.62 respectively confirms that different data compositions steer weights in substantially different directions despite nearly identical magnitudes. The embedding/LM-head layers are most aligned (0.82--0.88), while attention, MLP, and Mamba layers all show low directional similarity (0.48--0.64).}
    \label{fig:weight-direction}
\end{figure}

\paragraph{\textbf{Data composition determines the capabilities encoded, not the amount of change.}}
Table~\ref{tab:mc-mcs-comparison} shows that MC and MCS mid-training produce nearly identical weight divergence profiles for both models: total L2 of 0.177 vs.\ 0.175 for Granite-3.3, and 0.113 vs.\ 0.112 for Nemotron-H, with matching per-component breakdowns. Yet the downstream GPQA-Diamond capabilities differ dramatically: for Granite-3.3, MCS+RL achieves 52.86 vs.\ 35.52 for MC+RL (Table~\ref{tab:midtrain-rl-math-code-science}). To directly measure what differs, we compute the cosine similarity between the MC and MCS weight update vectors per component (Figure~\ref{fig:weight-direction}):
\begin{equation}
\cos(\Delta W_{MC}, \Delta W_{MCS}) = \frac{(W_{MC} - W_{\text{base}}) \cdot (W_{MCS} - W_{\text{base}})}{\|W_{MC} - W_{\text{base}}\|_2 \cdot \|W_{MCS} - W_{\text{base}}\|_2}
\label{eq:cos-sim}
\end{equation}
The overall cosine similarity (Eq.~\ref{eq:cos-sim}) is only 0.521 for Granite-3.3 and 0.623 for Nemotron-H, indicating that despite traveling nearly identical distances in weight space (L2: 0.177 vs.\ 0.175 for G33; 0.113 vs.\ 0.112 for Nemotron-H), the two data compositions reach substantially different weight configurations. All sub-components (attention, MLP, Mamba) show similarly low directional alignment (0.48--0.64), with only the embedding layers remaining closer (0.82--0.88). These results are consistent with the view that data composition primarily affects \emph{what configuration} the weights converge to, rather than the \emph{magnitude of the weight change} (as measured by normalized L2).

\begin{table}[t]
\centering
\small
\setlength{\tabcolsep}{3pt}
\begin{tabular}{l l c c c c}
\toprule
\textbf{Model} & \textbf{Mix} & \textbf{Attn} & \textbf{MLP} & \textbf{Mamba} & \textbf{Total} \\
\midrule
\multirow{2}{*}{Granite-3.3 (8B)}
& MC  & 0.177 & 0.333 & -- & 0.177 \\
& MCS & 0.175 & 0.329 & -- & 0.175 \\
\midrule
\multirow{2}{*}{Nemotron-H (8B)}
& MC  & 0.232 & 0.292 & 0.140 & 0.113 \\
& MCS & 0.230 & 0.289 & 0.138 & 0.112 \\
\bottomrule
\end{tabular}
\caption{\textbf{MC vs.\ MCS weight divergence (Base$\to$MT normalized L2).} Both models show nearly identical per-component L2 norms across data compositions, confirming that the training intensity is matched between MC and MCS despite their different downstream capabilities.}
\label{tab:mc-mcs-comparison}
\end{table}

\paragraph{\textbf{RL's weight footprint is independent of the starting point.}}
RL applied directly to base models (without mid-training) produces nearly identical weight changes to RL on mid-trained models, at both Granite-3.3 (0.0004 vs.\ 0.0003) and Nemotron-H (0.0002 vs.\ 0.0003). Yet the downstream outcomes differ drastically. A finer-grained sub-component analysis (Table~\ref{tab:subcomponent-rl}, Appendix~\ref{app:subcomponent}) confirms that this invariance extends to individual weight matrices: RL targets the same sub-components in nearly identical proportions regardless of whether mid-training preceded it. For Granite-3.3, value projections change 5.7\% (MT$\to$RL) vs.\ 7.5\% (Base$\to$RL), output projections 5.6\% vs.\ 6.7\%, and MLP gate projections 5.4\% vs.\ 6.1\%. Nemotron-H shows the same pattern, with Mamba parameters (A, dt) remaining completely frozen in both cases. This reveals that RL's sub-component targeting is an intrinsic property of the optimization process, not a consequence of mid-training. The large difference in outcomes despite similar weight change patterns suggests that mid-training appears to create model configurations from which RL can effectively improve performance, though the causal mechanism remains to be established, while base models do not benefit to the same degree despite receiving similar gradient-driven updates.

\begin{figure*}[t]
    \centering
    \includegraphics[width=\textwidth]{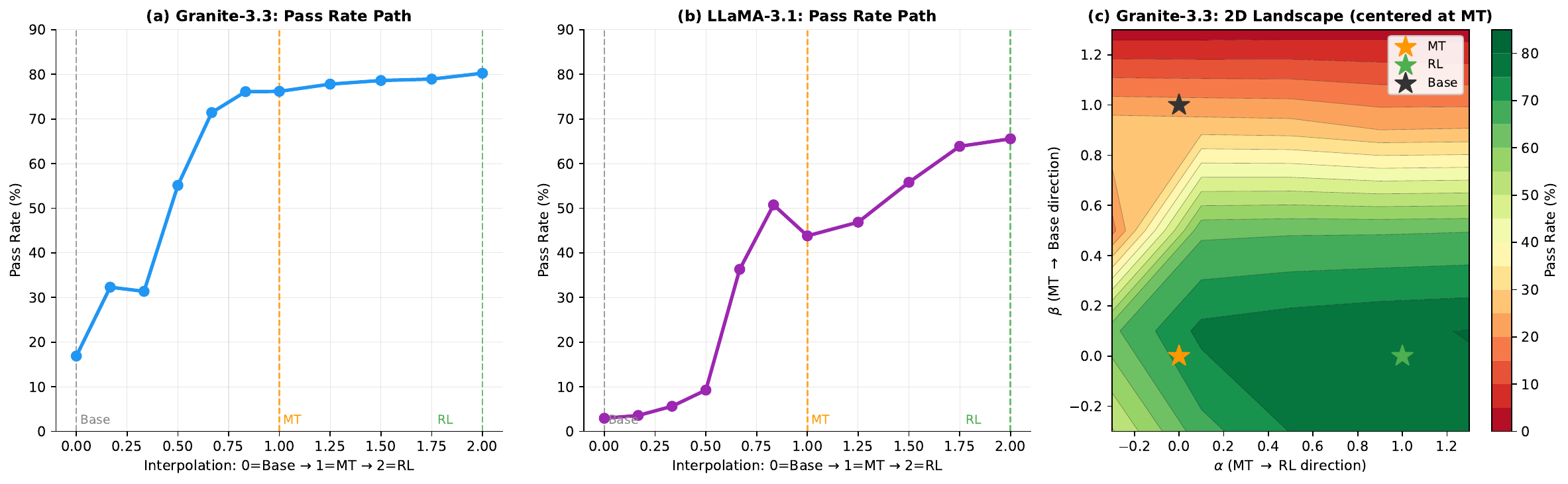}
    \caption{\textbf{Pass rate landscape on held-out MATH500 problems.} (a) Math pass rate at linearly interpolated weight checkpoints along the Base$\to$MT$\to$RL path for Granite-3.3 and LLaMA-3.1, evaluated on 200 held-out MATH500 problems (7680 generation tokens). Pass rate increases monotonically from Base to MT (16.9\%$\to$75.5\% for G33, 2.6\%$\to$43.1\% for LLaMA) and continues increasing through RL. (b) 2D pass rate landscape for Granite-3.3 centered at MT, with axes toward RL ($\alpha$) and toward Base ($\beta$). The RL direction consistently improves performance while moving toward Base degrades it.}
    \label{fig:reward-landscape}
\end{figure*}

\paragraph{\textbf{Pass rate landscape is consistent with mid-training creating a favorable configuration for RL.}}
To directly visualize this effect, we construct a \emph{pass rate landscape} by linearly interpolating model weights along the training path and evaluating math pass rate at each interpolated checkpoint. We use 200 held-out MATH500 problems (not included in the RL training pool) with temperature 0.6, top-$p$ 0.95, and 7680 max generation tokens, scored with the same verifier as RL training. We evaluate Granite-3.3 and LLaMA-3.1 (Figure~\ref{fig:reward-landscape}). The interpolated weights are:
\begin{equation}
W(\alpha, \beta) = W_{\text{base}} + \alpha(W_{MT} - W_{\text{base}}) + \beta(W_{RL} - W_{MT})
\label{eq:interp}
\end{equation}
where $\alpha=0, \beta=0$ recovers Base; $\alpha=1, \beta=0$ recovers MT; and $\alpha=1, \beta=1$ recovers RL (Eq.~\ref{eq:interp}). The 1D path sets $\beta=0$ and varies $\alpha$ from 0 to 1, then fixes $\alpha=1$ and varies $\beta$ from 0 to 1. The 2D landscape evaluates pass rate on a $5\times5$ grid over $(\alpha, \beta)$.

For Granite-3.3, pass rate increases from Base (17\%) to MT (76\%) as $\alpha$ increases from 0 to 1, then continues to RL (80\%) along the $\beta$ axis. LLaMA shows a similar trend: Base (3\%) to MT (44\%) to RL (66\%). The 2D landscape shows the RL direction consistently yields higher performance, while moving toward Base degrades it. No sharp barriers are apparent near the training path. 

The next section examines this further at the representation level: while RL's weight changes are consistent regardless of starting point, the resulting representations are dramatically more capable when built on top of mid-training.

%% ════════════════════════════════════════════════════════
\subsection{Representation Similarity Across Pipeline Stages}
\label{sec:cka-analysis}

\begin{figure*}[t]
    \centering
    \includegraphics[width=\textwidth]{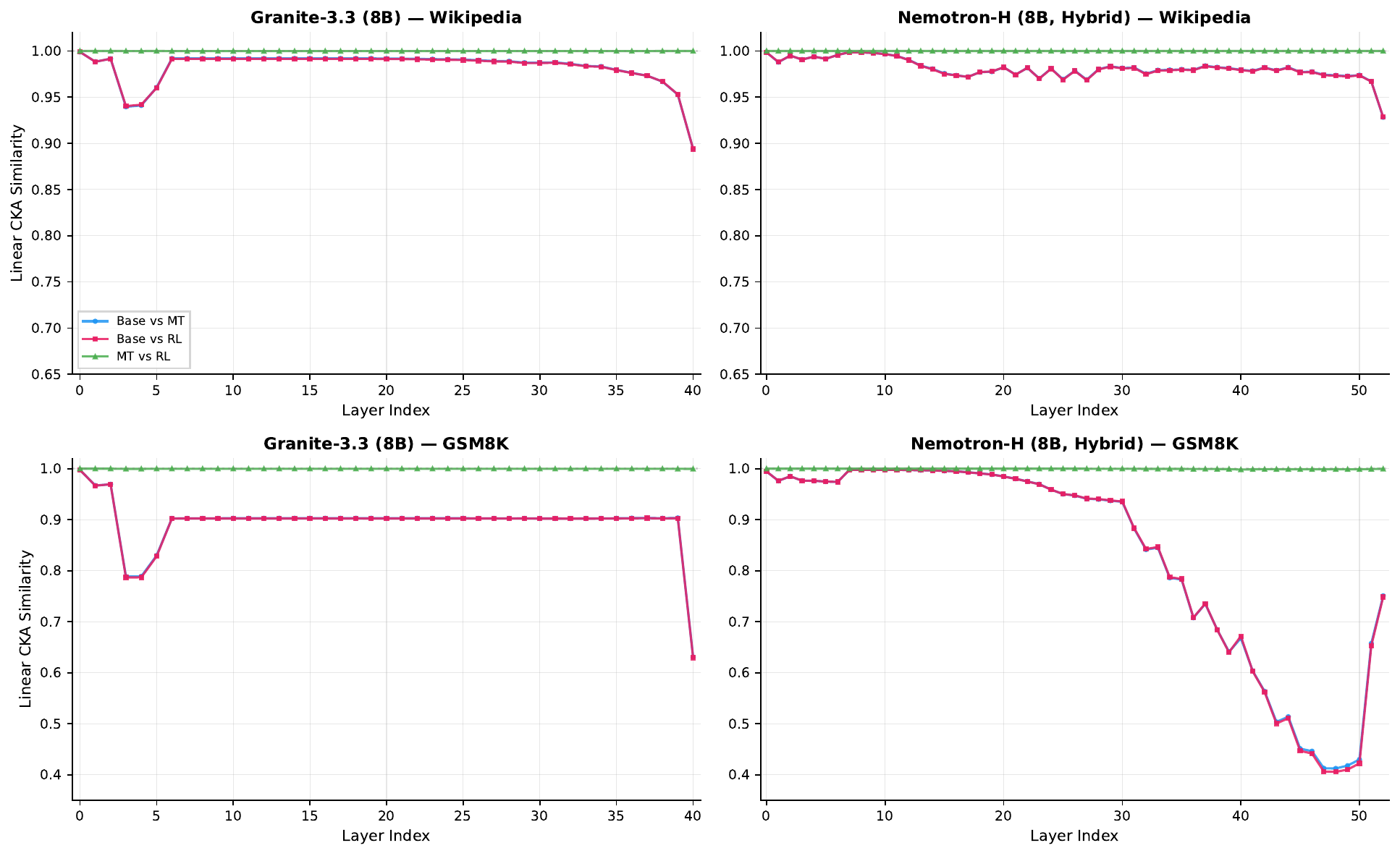}
    \caption{\textbf{RL preserves representational geometry; mid-training reshapes it in model-specific ways.} Layer-wise linear CKA~\cite{kornblith2019similarityneuralnetworkrepresentations} on Wikipedia (top) and GSM8K math prompts (bottom) for Granite-3.3 (left) and Nemotron-H (right), evaluated on 200 prompts per input type with batch-size-1 encoding. MT vs.\ RL (green) is $\approx$1.0 at every layer across both models and both input types, confirming RL preserves mid-training's representational geometry. Base vs.\ MT and Base vs.\ RL (blue, pink) are nearly identical, confirming all representational change comes from mid-training. The magnitude and layer pattern of mid-training's representational shift is model- and input-specific. See Table~\ref{tab:cka-summary} for the full summary.}
    \label{fig:cka-paper}
\end{figure*}

To complement the weight-level analysis, we measure how mid-training and RL reshape the model's internal \emph{representations} using linear Centered Kernel Alignment (CKA)~\citep{kornblith2019similarityneuralnetworkrepresentations}:
\begin{equation}
\text{CKA}(X, Y) = \frac{\|Y^{\top}X\|_F^2}{\|X^{\top}X\|_F \cdot \|Y^{\top}Y\|_F}
\label{eq:cka}
\end{equation}
where $X, Y \in \mathbb{R}^{n \times d}$ are mean-pooled hidden states from two checkpoints across $n$ inputs (Eq.~\ref{eq:cka}). CKA$=1$ indicates identical representational geometry; lower values indicate greater divergence. We feed identical text through the Base, MT, and RL checkpoints, extracting mean-pooled hidden states at each layer. To ensure robustness, we evaluate on three input distributions: Wikipedia (general text) ~\citep{merity2016pointersentinelmixturemodels}, C4 (web text) ~\citep{raffel2023exploringlimitstransferlearning}, and GSM8K (math prompts), across three models (Granite-3.3, LLaMA-3.1, Nemotron-H). To validate statistical stability, we perform bootstrap resampling (20 resamples of 100 from 200 inputs) and find that all MT vs.\ RL CKA estimates have standard deviations of at most 0.0001, confirming that the results are stable and not sensitive to the choice of input subset. Figure~\ref{fig:cka-paper} reports layer-wise linear CKA on Wikipedia and GSM8K for Granite-3.3 and Nemotron-H; additional models and input types are in Appendix~\ref{app:cka-extended}.

\paragraph{\textbf{RL preserves the representational geometry that mid-training creates.}}
Table~\ref{tab:cka-summary} shows MT vs.\ RL $>$0.998 for all three models across all three input types. This holds for dense Transformers (Granite-3.3, LLaMA-3.1) and the hybrid attention-Mamba architecture (Nemotron-H) alike. Furthermore, Base vs.\ MT and Base vs.\ RL curves are nearly identical at every layer, confirming that all representational geometry change is attributable to mid-training; RL achieves its gains through modifications within this established structure. RL achieves its benchmark gains through adjustments within the representational space that mid-training established, suggesting a division of roles between the two training stages.

\paragraph{\textbf{The output layer shows the largest mid-training shift.}}
For Granite-3.3, the sharpest Base vs.\ MT CKA divergence consistently occurs at the final transformer layer (layer 40) across all three inputs, but its depth is input-dependent: CKA $\approx$0.63 on GSM8K math prompts versus $\approx$0.89 on Wikipedia and C4. This input-specificity suggests the output layer restructuring is most pronounced for math reasoning content, consistent with the behavioral shift observed in Section~\ref{sec:entropy-analysis}: base models produce short, direct answers (median 124 tokens), while mid-trained models produce extended reasoning chains (2,196 tokens).

\paragraph{\textbf{Mid-training's representational impact is model- and input-specific.}}
Unlike the RL finding (which is consistent across all models), the Base vs.\ MT divergence pattern varies considerably across models and input types. For Granite-3.3, the largest divergence is at the final output layer across all inputs (CKA $\approx$0.63 on GSM8K, $\approx$0.89 on Wikipedia and C4). Nemotron-H shows the most pronounced divergence on GSM8K, with a deep dip in later layers (CKA $\approx$0.41 at layer 48) while recovering to $\approx$0.75 at the final layer; on Wikipedia the final layer CKA is $\approx$0.93, indicating the restructuring is heavily math-targeted. LLaMA-3.1 shows its deepest divergence on C4 web text (CKA $\approx$0.71 at layer 29) rather than GSM8K ($\approx$0.78), with the final layer recovering to $\approx$0.90. Each model was pretrained on a different data distribution, which is consistent with differences in how mid-training reshapes their representations, though we do not have access to the pretraining corpus compositions and cannot verify this hypothesis directly. Rather than making universal claims about where mid-training acts, we simply observe that its effect is model-dependent, whereas RL's preservation of representational geometry is consistent across all four models.

\begin{table}[t]
\centering
\small
\setlength{\tabcolsep}{3pt}
\begin{tabular}{l l c c c}
\toprule
\textbf{Model} & \textbf{Arch.} & \textbf{Wiki} & \textbf{C4} & \textbf{GSM8K} \\
\midrule
Granite-3.3 (8B) & Dense  & 0.9999\tiny{$\pm$0.0000} & 0.9999\tiny{$\pm$0.0000} & 0.9997\tiny{$\pm$0.0000} \\
LLaMA-3.1 (8B)   & Dense  & 0.9999\tiny{$\pm$0.0000} & 0.9999\tiny{$\pm$0.0000} & 0.9996\tiny{$\pm$0.0001} \\
Nemotron-H (8B)  & Hybrid & 0.9999\tiny{$\pm$0.0000} & 0.9998\tiny{$\pm$0.0000} & 0.9993\tiny{$\pm$0.0001} \\
\bottomrule
\end{tabular}
\caption{\textbf{MT vs.\ RL representational similarity (minimum linear CKA $\pm$ bootstrap std) across input distributions.} Values are the minimum layer-wise CKA across 20 bootstrap resamples of 100 from 200 inputs. RL consistently preserves mid-training's representational geometry ($>$0.998) across all three models and all three input types, spanning both dense Transformers and hybrid attention-Mamba architectures.}
\label{tab:cka-summary}
\end{table}

%% ════════════════════════════════════════════════════════
\subsection{Prediction Confidence and Correctness Across Pipeline Stages}
\label{sec:entropy-analysis}

\begin{figure*}[t]
    \centering
    \includegraphics[width=\textwidth]{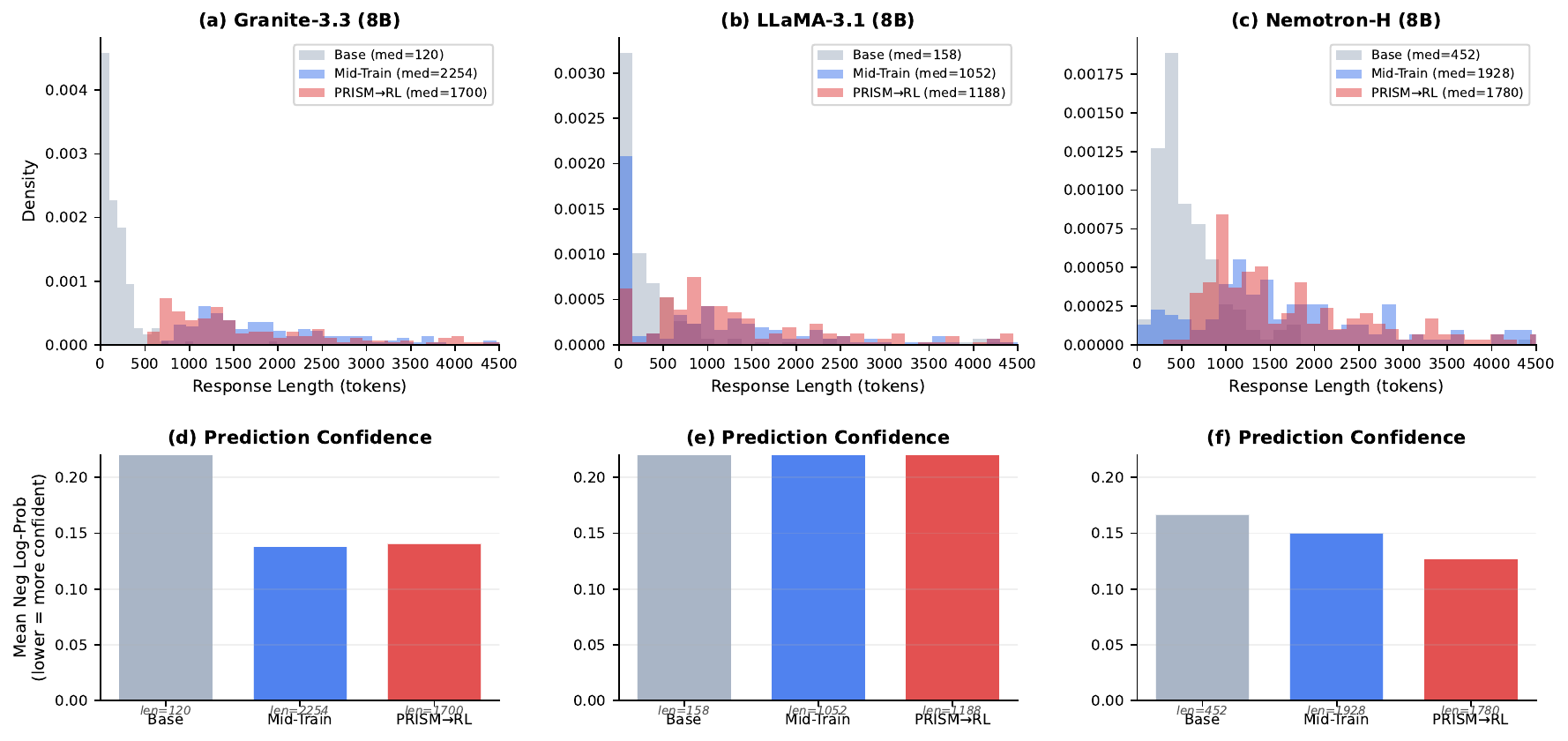}
    \caption{\textbf{Mid-training transforms prediction behavior: models learn to reason longer with calibrated confidence.} Evaluated on 200 held-out MATH500 problems. Top row: response length distributions shift from short outputs (Base, gray) to extended reasoning chains (MT, blue), with RL (red) adjusting length. Bottom row: mean negative log-probability at each stage.}
    \label{fig:token-entropy}
\end{figure*}

We sample 200 held-out MATH500 problems~\cite{lightman2023lets} and generate 8 responses per prompt at each pipeline stage using vLLM with temperature 0.6, top-$p$ 0.95, 7680 max generation tokens, and a step-by-step reasoning prompt suffix. Pass rate is averaged across all 8 samples per prompt and then across 200 prompts. We collect per-token log-probabilities during generation and score correctness using the same math verifier employed during RL training. We report mean \emph{negative log-probability} as a proxy for prediction confidence; note that this differs from predictive entropy, which would require marginalizing over the full output distribution. Results are in Table~\ref{tab:correctness-entropy} and Figure~\ref{fig:token-entropy}.

\begin{table}[t]
\centering
\small
\setlength{\tabcolsep}{3pt}
\begin{tabular}{l l c c c c c}
\toprule
\textbf{Model} & \textbf{Stage} & \textbf{Pass} & \textbf{Med. Len} & \textbf{Neg-LP} & \textbf{Corr.} & \textbf{Incorr.} \\
\midrule
\multirow{3}{*}{\shortstack[l]{Granite-3.3\\(8B)}}
& Base & 16.9\% & 120 & 0.382 & -- & 0.383 \\
& MT   & 75.5\% & 2,254 & 0.138 & 0.128 & 0.153 \\
& RL   & \textbf{79.5\%} & 1,700 & 0.141 & 0.135 & 0.160 \\
\midrule
\multirow{3}{*}{\shortstack[l]{LLaMA-3.1\\(8B)}}
& Base & 2.6\% & 158 & 0.758 & -- & 0.780 \\
& MT   & 43.1\% & 1,052 & 0.377 & 0.146 & 0.469 \\
& RL   & \textbf{64.6\%} & 1,188 & 0.267 & 0.149 & 0.320 \\
\midrule
\multirow{3}{*}{\shortstack[l]{Nemotron-H\\(8B, Hybrid)}}
& Base & 66.6\% & 452 & 0.167 & 0.040 & 0.258 \\
& MT   & 61.6\% & 1,928 & 0.150 & 0.116 & 0.156 \\
& RL   & \textbf{83.0\%} & 1,780 & 0.127 & 0.112 & 0.137 \\
\bottomrule
\end{tabular}
\caption{\textbf{Correctness, response length, and prediction confidence across pipeline stages} on 200 held-out MATH500 problems (8 samples/prompt, 7680 max generation tokens, step-by-step reasoning prompt). Pass = mean pass rate across 8 samples per prompt (\%). Med.\ Len = median response length (tokens). Neg-LP = mean negative log-probability. Corr./Incorr.\ = mean neg-LP for correct/incorrect responses; -- indicates too few correct samples. The $\textsc{PRISM} \to \text{RL}$ pipeline consistently achieves the highest pass rates across all three model families.}
\label{tab:correctness-entropy}
\end{table}

\paragraph{\textbf{Mid-training teaches models to reason, not just answer.}}
The most striking behavioral change is in response length. LLaMA base generates a median of just \textbf{158 tokens} on MATH500 problems, Granite-3.3 base produces 120, and Nemotron-H base 452. After mid-training, all three produce extended reasoning chains: LLaMA increases to 1,052 tokens, Granite-3.3 extends to 2,254, and Nemotron-H to 1,928 (Table~\ref{tab:correctness-entropy}). This is consistent with mid-training's primary behavioral effect being the acquisition of multi-step problem decomposition.

\paragraph{\textbf{The full pipeline dramatically improves correctness.}}
Granite-3.3 improves from 16.9\% to \textbf{79.5\%} pass rate, LLaMA from 2.6\% to \textbf{64.6\%}, and Nemotron-H from 66.6\% to \textbf{83.0\%}. Nemotron-H is a notable case: the base model already achieves 66.6\% on MATH500, generating 452-token responses that often reach direct correct answers. Mid-training introduces chain-of-thought reasoning patterns (extending to 1,928 tokens), but these extended generation strategies may conflict with the base model's existing direct-solution approaches, leading to a regression at the MT stage (61.6\%). This tension is resolved by RL, which optimizes for correctness and recovers well above the base level (83.0\%). This pattern of brief MT regression followed by strong RL recovery is consistent with the hypothesis that mid-training reshapes generation behavior in ways that require RL to fully unlock the capability gains. RL consistently improves over MT alone for all three models. Correct responses tend to have \emph{lower} negative log-probability than incorrect ones across all stages and models (Table~\ref{tab:correctness-entropy}, Corr.\ vs.\ Incorr.\ columns), suggesting that higher model confidence is on average associated with correctness. This effect is most pronounced for LLaMA-3.1 (e.g., 0.149 correct vs.\ 0.320 incorrect at RL) and smallest for Nemotron-H at the RL stage (0.112 vs.\ 0.137).

\paragraph{\textbf{Mid-training calibrates prediction confidence.}}
Mid-training substantially reduces mean negative log-probability across all models, indicating increased overall confidence: Granite-3.3 from 0.382 to 0.138, LLaMA from 0.758 to 0.377, and Nemotron-H from 0.167 to 0.150. For LLaMA, the gap between correct and incorrect response confidence \emph{widens} after mid-training (e.g., correct: 0.146 vs.\ incorrect: 0.469 at MT), indicating better calibration. Nemotron-H behaves differently: the base model is already highly confident on correct answers (neg-LP = 0.040) but very uncertain on incorrect ones (0.258); after mid-training and RL, confidence converges to a narrower range (correct: 0.112, incorrect: 0.137 at RL), making predictions more uniformly confident while still maintaining a separation between correct and incorrect responses.

\paragraph{\textbf{RL refines toward efficient, correct reasoning.}}
RL adjusts response length in a model-dependent direction: shortening for Granite-3.3 (2,254$\to$1,700), while Nemotron-H (1,928$\to$1,780) and LLaMA (1,052$\to$1,188) show modest changes. In all cases, RL maintains or improves the confidence gap between correct and incorrect answers while substantially increasing pass rates, demonstrating that it optimizes both the quality and efficiency of the reasoning process that mid-training established.

%% ════════════════════════════════════════════════════════
\subsection{RL Weight Trajectory: Front-Loaded Optimization}
\label{sec:rl-trajectory}

\begin{figure*}[t]
    \centering
    \includegraphics[width=\textwidth]{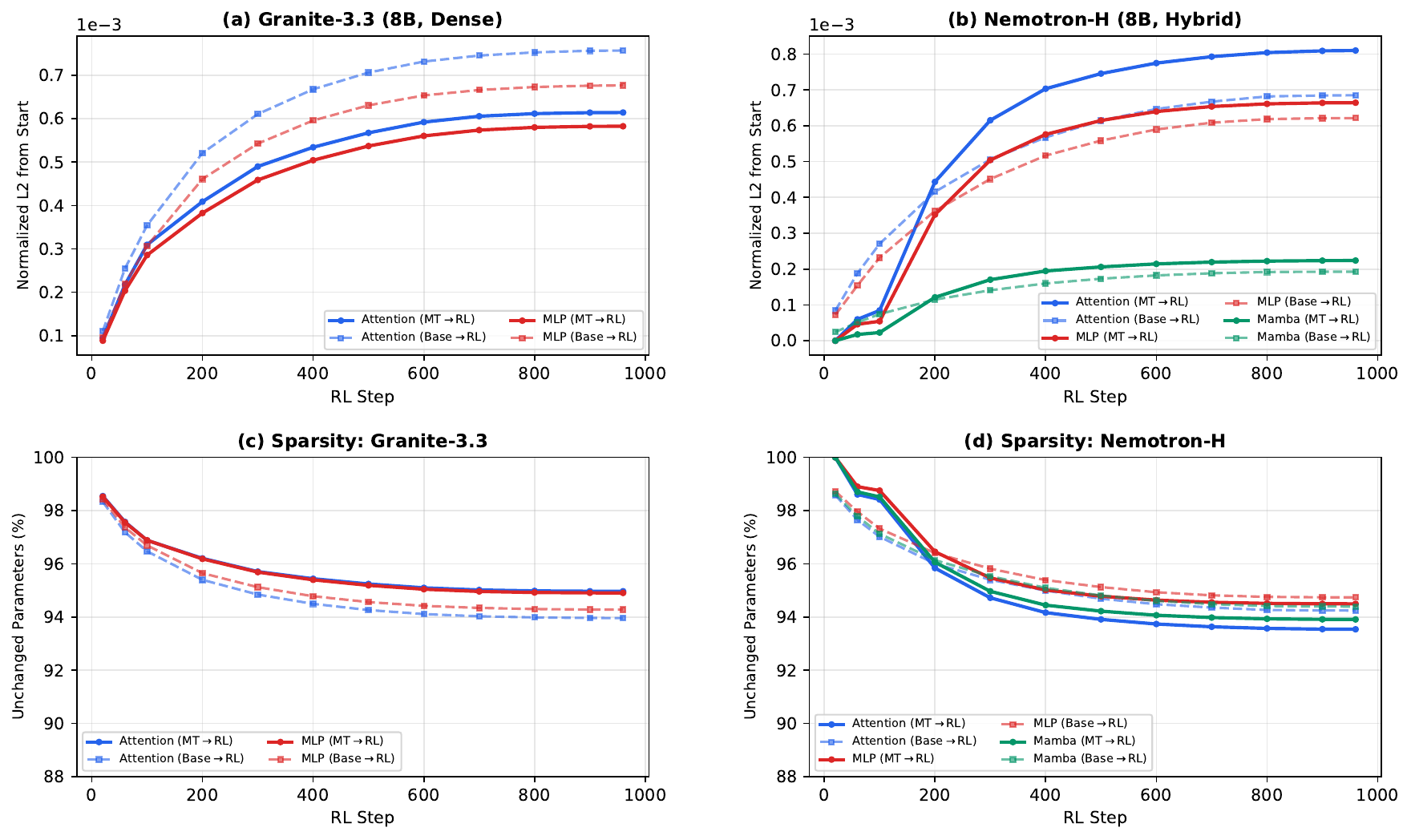}
    \caption{\textbf{RL optimization is front-loaded and starting-point-invariant.} Top row: cumulative L2 divergence from the initial checkpoint over RL steps for Granite-3.3 (left) and Nemotron-H (right). Solid lines: MT$\to$RL; dashed lines: Base$\to$RL. Most weight change occurs in the first $\sim$200--400 steps, then plateaus. MT$\to$RL and Base$\to$RL produce nearly identical divergence profiles, confirming that RL's weight footprint is independent of the starting point. Bottom row: sparsity evolution showing the fraction of parameters within 1\% of their initial values. The active parameter set grows progressively from $\sim$1.5\% at step 20 to $\sim$5--6\% by step 960, with all component types following the same trajectory.}
    \label{fig:rl-trajectory}
\end{figure*}

We track weight evolution across RL training steps (20 to 960) for both Granite-3.3 and Nemotron-H, comparing MT$\to$RL and Base$\to$RL trajectories. Results are shown in Figure~\ref{fig:rl-trajectory}.

\paragraph{\textbf{RL weight changes are front-loaded.}}
Across both architectures, the majority of RL's cumulative weight divergence accumulates in the first $\sim$200--400 steps, with the L2 curve plateauing thereafter (Figure~\ref{fig:rl-trajectory}, top row). For Nemotron-H, attention divergence reaches 80\% of its final value by step 400; for Granite-3.3, the pattern is similar. This front-loading is consistent with the benchmark learning curves, which show the steepest performance gains in early RL steps. The component hierarchy is also consistent across all runs: attention layers change most, followed by MLP, then Mamba (in hybrid models).

\paragraph{\textbf{The active parameter set is emergent, not predetermined.}}
RL does not modify a fixed subset of parameters from the outset. Instead, the fraction of changed parameters grows progressively: at step 20, only $\sim$1.5\% of parameters have moved beyond the 1\% relative threshold, expanding to $\sim$5--6\% by step 960 (Figure~\ref{fig:rl-trajectory}, bottom row). This gradual activation pattern, combined with the front-loaded divergence, shows that RL's sparse update set is not fixed from the outset but expands progressively over the course of training.

\paragraph{\textbf{Starting point does not affect RL's weight trajectory.}}
Comparing MT$\to$RL (solid) with Base$\to$RL (dashed) on the same axes reveals nearly identical L2 and sparsity trajectories for both Granite-3.3 and Nemotron-H. The final L2 divergence differs by less than 20\% between starting points, and sparsity converges to within 1 percentage point. This provides additional evidence, beyond the single-checkpoint analysis in Section~\ref{sec:weight-analysis}, that RL applies a similarly scaled and sparse update pattern regardless of the starting point. The difference in downstream performance is thus consistent with arising from \emph{where} in weight space the updates land, rather than from differences in the magnitude or sparsity of how RL modifies weights.

\section{Conclusion and Future Work}

We presented \textsc{PRISM}, a comprehensive empirical study of mid-training design choices for LLMs. Through controlled experiments across seven base models from four families (Granite, LLaMA, Mistral, Nemotron-H), two architecture types (dense Transformer and attention-Mamba hybrid), and scales from 3B to 24B parameters, we established several findings that we believe are valuable for practitioners designing mid-training pipelines:

\begin{itemize}
    \item[$\bullet$] A relatively small mid-training phase ($\sim$27B tokens) yields \textbf{+15 to +40 point} math gains and \textbf{+5 to +12 point} code gains across all tested models, with science gains of \textbf{+6 to +13 points} on Granite and hybrid models, while preserving general performance.
    \item[$\bullet$] Data composition choices matter most at mid-training, not at RL. Including science data during mid-training unlocks \textbf{+17 to +28 point} GPQA-Diamond gains during RL, while changing the RL mix produces $<$2 point differences.
    \item[$\bullet$] The full $\textsc{PRISM} \to \text{RL}$ pipeline improves the six-benchmark macro-average from under 12 to \textbf{29--42}, a 3--4$\times$ improvement. RL applied directly to base models is substantially less effective.
    \item[$\bullet$] For Granite-3.3, mid-training at 8k context degrades long-context ability, but this can be largely restored via a brief extension phase combined with model merging. Note that all models in our study were pretrained with long-context phases, so the interaction between long-context pretraining and mid-training effectiveness may vary in other settings.
    \item[$\bullet$] For Granite-3.3, RL on mid-trained models progressively solves initially unsolvable prompts, with non-saturating training curves suggesting further gains are achievable.
    \item[$\bullet$] At the weight level, mid-training densely restructures $>$90\% of parameters (370--580$\times$ larger than RL), while RL sparsely refines $\sim$5\%, with identical footprints regardless of whether mid-training preceded it. Representation analysis (CKA) across three models and three input distributions confirms that RL consistently preserves mid-training's representational geometry ($>$0.998) across both dense Transformers and hybrid architectures, while mid-training's representational impact is model-specific. RL optimization is front-loaded, with most weight changes in the first $\sim$200--400 steps. Behaviorally, mid-training produces extended reasoning chains in model outputs. 
\end{itemize}

\paragraph{\textbf{Limitations and future directions.}}
Our study has several limitations that point to productive future work.

\noindent\textbf{Model-specific RL data selection.} For consistency across model families, we filtered RL prompts using a single model (Granite-3.3-8B mid-trained) and applied the same mix to all models. In practice, different mid-trained models have different difficulty profiles, and model-specific prompt selection would likely yield stronger per-model results. Our goal was not to produce optimal per-model recipes but to enable controlled cross-model comparisons. Investigating adaptive, model-aware RL data curation is a natural next step.

\noindent\textbf{Broader domain coverage.} Our mid-training mixtures focus on math, code, and science. Extending \textsc{PRISM} to additional domains such as multilingual reasoning, agentic tasks, and tool use would test whether the patterns we observe (e.g., domain synergies, retention via general web data) hold more broadly.

\noindent\textbf{Scaling beyond 24B.} Our largest model is Mistral-Small (24B). Verifying that \textsc{PRISM}'s findings extend to models at the 70B+ scale, where mid-training compute budgets and data requirements may differ qualitatively, remains an open question.

\noindent\textbf{Long-context mid-training.} Our primary experiments use 8k context during mid-training. While our ablations show that 16k yields additional gains, we did not explore mid-training at 32k+ with proportionally larger token budgets. Jointly optimizing context length and token budget during mid-training could further improve the reasoning/retention trade-off.

Overall, \textsc{PRISM} demonstrates that retention-aware mid-training is a highly effective intermediate step for reliable reasoning enhancement and RL scaling. We hope that the practical guidelines and comprehensive analyses provided in this work will help the community design more effective mid-training pipelines for modern LLMs.

\clearpage
\newpage
\bibliographystyle{plainnat}
\bibliography{paper}

%\clearpage
\newpage
\beginappendix
This appendix provides supplementary details for the main paper. We begin with model specifications and training hyperparameters for \textsc{PRISM} mid-training, long-context restoration, and RL (Appendix~\ref{sec:appendix-hparams}). We then describe our evaluation benchmarks and settings (Appendix~\ref{app:eval_details}), followed by extended results tables referenced from the main text (Appendix~\ref{app:extended_results}). We present RL training details including the GRPO algorithm and hyperparameters (Appendix~\ref{app:rl_training_details}), RL training curves for Granite-4 Micro Dense (Appendix~\ref{app:hybrid_rl}), and additional RL learning curves (Appendix~\ref{app:additional_rl_curves}). We provide extended mechanistic analyses: RL sub-component weight analysis (Appendix~\ref{app:subcomponent}), extended CKA representation analysis across four models and three input distributions (Appendix~\ref{app:cka-extended}), and AIME 2026 evaluation (Appendix~\ref{app:aime26}). Finally, we present qualitative model generations (Appendix~\ref{app:model_generations}).

%% ════════════════════════════════════════════════════════
%% A. Model and Training Details (moved to first)
%% ════════════════════════════════════════════════════════

\section{Model and Training Details}\label{sec:appendix-hparams}

\subsection{Model Specifications}
\label{app:model_details}

To evaluate the cross-architecture robustness of \textsc{PRISM}, we select a diverse set of LLMs ranging from 3B to 24B parameters, including dense Transformers and attention-Mamba hybrids.

\begin{description}
    \item[\textbf{Dense Transformer Models:}] We utilize \textbf{LLaMA-3.1 8B}~\cite{grattafiori2024llama3herdmodels} and \textbf{Mistral-7B-v0.1}~\cite{jiang2023mistral7b} as primary baselines. For enterprise-focused evaluation, we include \textbf{Granite-3.3 8B}~\cite{granite2025granite33base} and the lightweight \textbf{Granite-4.0 Micro 3B}~\cite{granite2025granite40collection}, alongside the larger \textbf{Mistral-Small-24B}~\cite{mistral2025mistralsmall3}.

    \item[\textbf{Hybrid Attention-Mamba Architectures:}] We include \textbf{Granite-4.0-H Micro 3B} and \textbf{Nemotron-H 8B}, which alternate between standard attention layers and Mamba2 layers, representing the hybrid paradigm.
\end{description}

\subsection{PRISM Training Details}

Table~\ref{tab:prism-training-details} summarizes the training hyperparameters used for \textsc{PRISM} mid-training across all models unless otherwise specified.

\begin{table}[t]
\centering
\footnotesize

\begin{minipage}[t]{0.45\columnwidth}
\centering
\setlength{\tabcolsep}{4pt}
\begin{tabular}{ll}
\toprule
\textbf{Category} & \textbf{Setting} \\
\midrule
Training steps & 25,000 \\
Micro batch size & 1 \\
Gradient accumulation steps & 1 \\
Effective batch size & 1 \\
\midrule
Optimizer & AdamW \\
Learning rate & $5 \times 10^{-5}$ \\
Weight decay & 0.1 \\
Adam $\beta_1, \beta_2$ & (0.9, 0.95) \\
Adam $\epsilon$ & $1 \times 10^{-10}$ \\
\midrule
Learning rate schedule & Cosine decay \\
Warmup steps & 500 \\
Decay steps & 24,500 \\
Final LR factor & 0.1 \\
\midrule
Precision & bfloat16 (bf16) \\
\midrule
FSDP algorithm & 2 \\
Data parallel sharding & 8 \\
Data parallel replication & 16 \\
\bottomrule
\end{tabular}
\captionof{table}{\textbf{PRISM mid-training hyperparameters.}}
\label{tab:prism-training-details}
\end{minipage}
\hfill
\begin{minipage}[t]{0.45\columnwidth}
\centering
\setlength{\tabcolsep}{4pt}
\begin{tabular}{ll}
\toprule
\textbf{Category} & \textbf{Setting} \\
\midrule
Training steps & 1,000 \\
Micro batch size & 1 \\
Gradient accumulation steps & 1 \\
Effective batch size & 1 \\
Evaluation during training & Disabled \\
Evaluation interval & $10^9$ steps \\
\midrule
Optimizer & AdamW \\
Learning rate & $5 \times 10^{-5}$ \\
Weight decay & 0.1 \\
Adam $\beta_1, \beta_2$ & (0.9, 0.95) \\
Adam $\epsilon$ & $1 \times 10^{-10}$ \\
\midrule
Learning rate schedule & Exponential decay \\
Warmup steps & 100 \\
Constant steps & 0 \\
Final LR factor & 0 \\
\midrule
Precision & bfloat16 (bf16) \\
\midrule
FSDP algorithm & 2 \\
Context parallelism & 4 \\
Data parallel sharding & 4 \\
Data parallel replication & 9 \\
Gradient checkpointing & Enabled \\
\bottomrule
\end{tabular}
\captionof{table}{\textbf{Long-context restoration hyperparameters.}}
\label{tab:long-context-training-details}
\end{minipage}

\end{table}

\subsection{Long-Context Extension Phase}\label{sec:appendix-lc}

Table~\ref{tab:long-context-training-details} summarizes the hyperparameters used for the long-context extension phase applied after mid-training, and Table~\ref{tab:long-context-data} lists the datasets used.

\begin{table}[t]
\centering
\footnotesize
\setlength{\tabcolsep}{3pt}
\begin{tabular}{
>{\RaggedRight\arraybackslash}p{0.40\columnwidth}
>{\RaggedRight\arraybackslash}p{0.27\columnwidth}
>{\centering\arraybackslash}p{0.12\columnwidth}
}
\toprule
\textbf{Dataset} & \textbf{Type} & \textbf{Tokens (B)} \\
\midrule
DCLM-EDU & General Web Data & 2.51 \\
\midrule
Nemotron Post-Training v1 & Math (QA/Reasoning) & 5.08 \\
Megamath-Web-Pro & Math (web) & 4.33 \\
\midrule
StarCoder2 & Code (web) & 37.52 \\
xenArcAI-codex & Code (QA/Reasoning) & 3.860 \\
\bottomrule
\end{tabular}
\caption{\textbf{Datasets used in the long-context extension phase.} Token counts in billions (Granite 3.3, 8B).}
\label{tab:long-context-data}
\end{table}

\paragraph{\textbf{Long-context sequence packing via Best-Fit Decreasing (BFD).}}
To efficiently construct fixed-length long-context training sequences while minimizing truncation and wasted capacity, we employ a Best-Fit Decreasing (BFD) packing strategy. Documents are optionally split into overlapping chunks if they exceed the target context length $L$, then sorted in decreasing order of length. Each chunk is greedily assigned to an existing sequence buffer whose remaining capacity is sufficient and minimal among all feasible buffers; if no such buffer exists, a new buffer is created. Compared to naive concatenation or first-fit strategies, BFD packing significantly reduces unnecessary document truncation and improves token utilization while preserving document-level coherence.

%% ════════════════════════════════════════════════════════
%% B. Evaluation Details
%% ════════════════════════════════════════════════════════

\section{Evaluation Details}
\label{app:eval_details}

\paragraph{\textbf{Benchmark details.}}
General ability is assessed via LB-V1 (ARC, HellaSwag, MMLU, TruthfulQA, WinoGrande, GSM8K) and LB-V2 (IFEval, BBH, MATH, GPQA, MuSR, MMLU-Pro), which detect generalization regressions. Long-context capabilities are validated by RULER~\cite{hsieh2024rulerwhatsrealcontext}, which measures effective reasoning across massive token windows. LiveCodeBench~\cite{jain2024livecodebenchholisticcontaminationfree} and Codeforces~\cite{penedo2025codeforces} provide contamination-free code evaluation using time-stratified problems and elite algorithmic challenges. AIME~\cite{aime} and MATH500~\cite{lightman2023lets} track mathematical proficiency, highly sensitive to data mixture quality. GPQA-Diamond~\cite{rein2023gpqagraduatelevelgoogleproofqa} offers ``Google-proof'' PhD-level science challenges.

\paragraph{\textbf{Evaluation settings.}}
For math benchmarks (MATH500, AIME24/25), we use 32k max generation tokens, temperature 0.6, top-p 0.95, and 64 samples per prompt. For code benchmarks (Codeforces, LiveCodeBench), we use 32k max generation tokens, temperature 0.7, and 3 samples per prompt. Math benchmarks are evaluated using Qwen-eval~\cite{yang2024qwen2}, code benchmarks using Evalchemy~\cite{Evalchemy}, and RULER using HELMET~\cite{yen2025helmet}.

%% ════════════════════════════════════════════════════════
%% C. Extended Results Tables
%% ════════════════════════════════════════════════════════

\section{Extended Results Tables}
\label{app:extended_results}

This section presents the full benchmark breakdowns referenced in the main text. Table~\ref{tab:app_domain-specific-results} provides the per-benchmark results for the Granite-3.3-8B domain ablation (Math only, Math+Code, Math+Code+Science), complementing the summary in Section~5. Table~\ref{appendix:token-budget-ablation} reports the token-budget scaling experiment on Granite-4 Micro (3B), showing how performance evolves as the mid-training budget increases from 10.5B to 31.5B tokens.

\begin{table*}[t]
\centering
\scriptsize
\setlength{\tabcolsep}{6pt}
\resizebox{\textwidth}{!}{
\begin{tabular}{l r r r r r r r r}
\toprule
 & \multicolumn{3}{c}{\textbf{Code}} & \textbf{Science} & \multicolumn{4}{c}{\textbf{Math}} \\
\cmidrule(lr){2-4} \cmidrule(lr){5-5} \cmidrule(lr){6-9}
Model / Mixture
& LiveCodeBench & Codeforces & Code Avg
& GPQA-Diamond
& AIME 24 & AIME 25 & MATH 500 & Math Avg \\
\midrule
Base
& 2.15 & 1.99 & 2.07
& 22.56
& 0.46 & 0.31 & 26.09 & 8.95 \\

Math only
& 2.15 & 3.46 & 2.81
& 17.34
& 26.72 & 22.08 & 60.50 & 36.43 \\

Math + Code
& 11.11 & 10.30 &\textbf{ 10.71}
& 19.02
& 32.44 & 28.33 & 74.22 & 44.33 \\

\rowcolor{teal!10}
Math + Code + Science
& 10.63 & 10.52 & \textit{10.58}
& \textbf{29.12}
& 37.18 & 27.96 & 81.11 & \textbf{48.75} \\
\bottomrule
\end{tabular}}
\caption{\textbf{Domain-specific evaluation results for Granite-3.3-8B (full breakdown).} Code Avg is the mean of LiveCodeBench and Codeforces; Math Avg is the mean of AIME 24, AIME 25, and MATH 500.}
\label{tab:app_domain-specific-results}
\end{table*}

\begin{table*}[t]
\centering
\scriptsize
\setlength{\tabcolsep}{4pt}
\renewcommand{\arraystretch}{1.05}
\begin{tabular}{l r r r r r r r r r r r}
\toprule
Token Budget (B)
& LB V1 & LB V2
& LCB & CF & Code Avg
& GPQA-D
& AIME24 & AIME25 & MATH500 & Math Avg \\
\midrule
Base
& 66.01 & 21.82
& 0.24 & 2.28 & \textbf{1.26}
& 21.55
& 16.09 & 12.34 & 50.42 & \textbf{26.28} \\
\midrule
10.49
& 63.45 & 19.50
& 10.51 & 8.68 & \textbf{9.59}
& 19.19
& 23.95 & 19.17 & 77.52 & \textbf{40.21} \\

15.73
& 63.24 & 19.79
& 10.75 & 7.28 & \textbf{9.02}
& 23.06
& 26.14 & 21.30 & 78.76 & \textbf{42.07} \\

26.21
& 63.28 & 19.63
& 9.80 & 7.58 & \textbf{8.69}
& 19.19
& 28.49 & 20.10 & 78.08 & \textbf{42.22} \\

31.46
& 63.16 & 20.05
& 8.24 & 6.99 & \textbf{7.62}
& 21.38
& 28.02 & 22.08 & 77.15 & \textbf{42.42} \\
\bottomrule
\end{tabular}
\caption{\textbf{Effect of increasing mid-training token budget on Granite-4 Micro (3B)} using the Math+Code mixture with fixed 8k context length. LCB denotes LiveCodeBench, CF denotes Codeforces, and GPQA-D denotes GPQA-Diamond. \textbf{Code Avg} averages LCB and CF, while \textbf{Math Avg} averages AIME24, AIME25, and MATH500.}
\label{appendix:token-budget-ablation}
\end{table*}

%% ════════════════════════════════════════════════════════
%% D. RL Training Details
%% ════════════════════════════════════════════════════════

\section{RL Training Details}
\label{app:rl_training_details}

We use Group Relative Policy Optimization (GRPO)~\citep{shao2024deepseekmathpushinglimitsmathematical} as our RL algorithm for all models. The objective and advantage computation are:

\begin{equation}
\label{eqn:grpo_dapo}
\begin{split}
\mathcal{J}_{GRPO}(\theta) = \mathbb{E} & _{q \sim P(Q), \{o_i\}_{i=1}^G \sim \pi_{\theta_{gen}}} \left[ \frac{1}{\sum_{j=1}^G |o_j|} \sum_{i=1}^G \sum_{t=1}^{|o_i|} \right. \\
& \min \left( \frac{\pi_{\theta_{old}}(o_{i,t}|q)}{\pi_{\theta_{gen}}(o_{i,t}|q)}, C \right) \times \\
& \min \left( \frac{\pi_\theta(o_{i,t}|q)}{\pi_{\theta_{old}}(o_{i,t}|q)} \hat{A}_i, \right. \\
& \text{clip} \left( \frac{\pi_\theta(o_{i,t}|q)}{\pi_{\theta_{old}}(o_{i,t}|q)}, 1-\epsilon_{\text{low}}, 1+\epsilon_{\text{high}} \right) \hat{A}_i \bigg) \\
& \left. \vphantom{\frac{1}{\sum_{j=1}^G}} - \beta \mathbb{D}_{KL}[\pi_\theta || \pi_{ref}] \right]
\end{split}
\end{equation}

\begin{equation}
\label{eqn:adv}
\hat{A}_i = \frac{r_i - \operatorname{mean}(r_1, \dots, r_G)}{\operatorname{std}(r_1, \dots, r_G) + \eta}
\end{equation}

\begin{table}[ht]
\centering
\caption{\textbf{RL training hyperparameters and configuration.}}
\label{tab:rl_hyperparams}
\begin{tabular}{ll}
\toprule
\textbf{Hyperparameter} & \textbf{Value} \\ \midrule
\rowcolor{gray!10} \textbf{Sampling Parameters} & \\
Responses per Prompt & 16 \\
Temperature & 1.0 \\
Top-$p$ & 1.0 \\
\midrule
\rowcolor{gray!10} \textbf{Training Dynamics} & \\
Total Batch Size & 1024 \\
Unique Prompts per Batch & 64 \\
Training Steps & 1,000 \\
Context Length & 16,384 \\
Learning Rate & $5 \times 10^{-7}$ \\
LR Schedule & Linear Decay \\
KL-loss Beta ($\beta$) & 0.05 \\
\midrule
\rowcolor{gray!10} \textbf{RL Environment \& Rewards} & \\
Importance Sampling Ratio & 1.0 (On-policy) \\
Truncated IS Constant ($C$) & 2.0 \\
Format Reward & \texttt{<think> ... </think>} \\
Penalties & Stop token, Repetition \\
Reference Policy Update & Soft-update \\
\midrule
\rowcolor{gray!10} \textbf{Infrastructure} & \\
Optimizer & AdamW \\
Parallelism Strategy & DeepSpeed Stage 3 \\
Packing Length & 16,384 \\
 \bottomrule
\end{tabular}
\end{table}

Table~\ref{tab:rl_hyperparams} provides the full RL training configuration. We use on-policy model updates ($\text{train\_batch\_size} = \text{inference\_batch\_size}$). Following~\citep{gorbatovski2025learnreferencemodelreal}, we soft-update the reference policy for better convergence. Since our mid-training data teaches the model to think, we apply a \textit{thinking format reward}: the model's response is evaluated for correctness only if it adheres to the format \texttt{<think> thoughts </think> response}, otherwise it receives a reward of 0.0. We also apply stop-token and repetition penalties. We apply the Truncated Importance Sampling ratio~\citep{yao2025offpolicy} to control for training-inference mismatch. All experiments use the open-instruct codebase.\footnote{https://github.com/allenai/open-instruct/tree/main}

%% ════════════════════════════════════════════════════════
%% E. RL Training Curves for Granite-4 Micro Dense
%% ════════════════════════════════════════════════════════

\section{RL Training Curves for Granite-4 Micro Dense}
\label{app:hybrid_rl}

\begin{figure}[t]
    \centering
    \begin{subfigure}{0.24\linewidth}
        \centering
        \includegraphics[width=\linewidth]{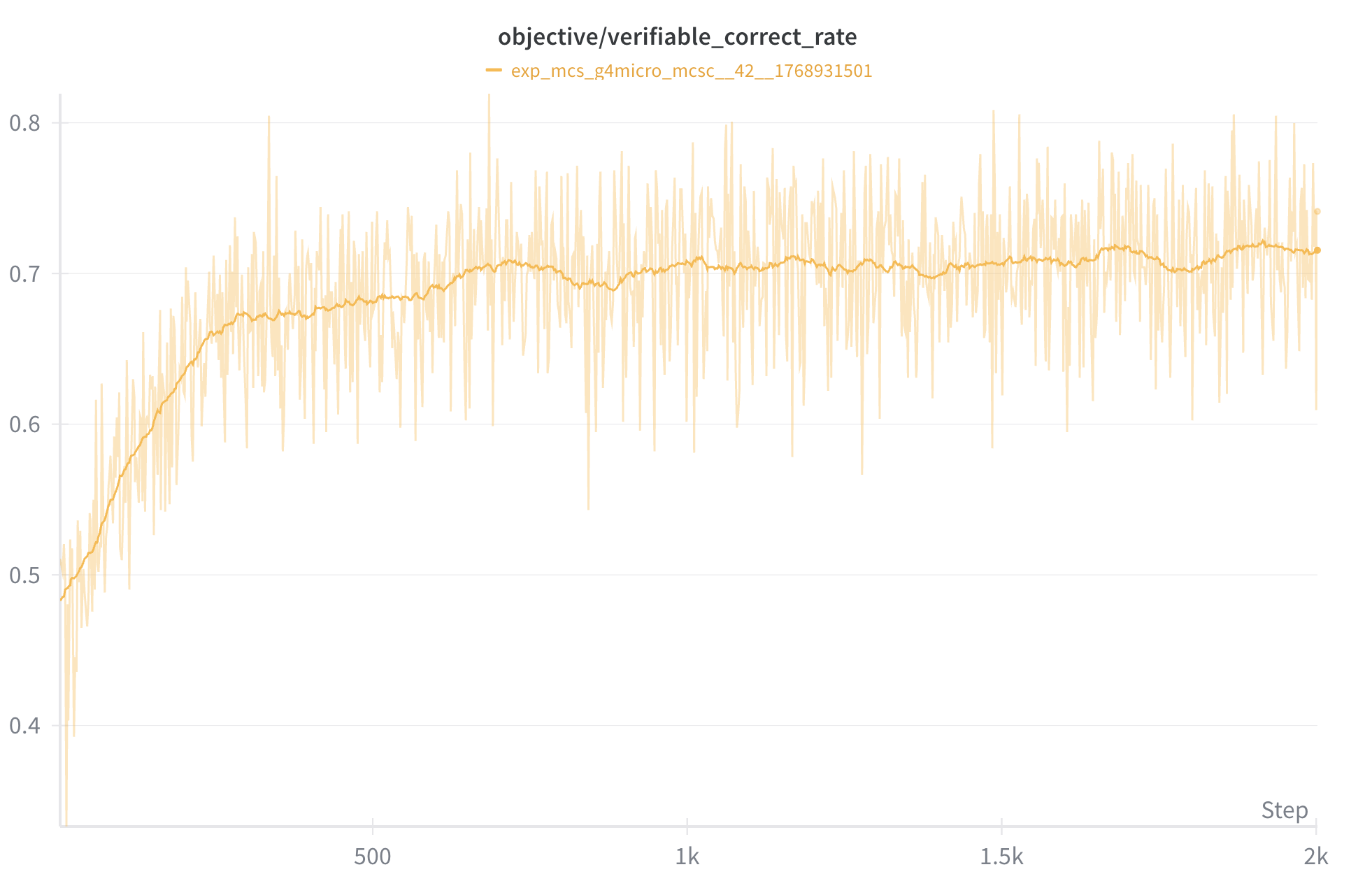}
        \caption{Overall}
        \label{fig:g4micro_cr}
    \end{subfigure}\hfill
    \begin{subfigure}{0.24\linewidth}
        \centering
        \includegraphics[width=\linewidth]{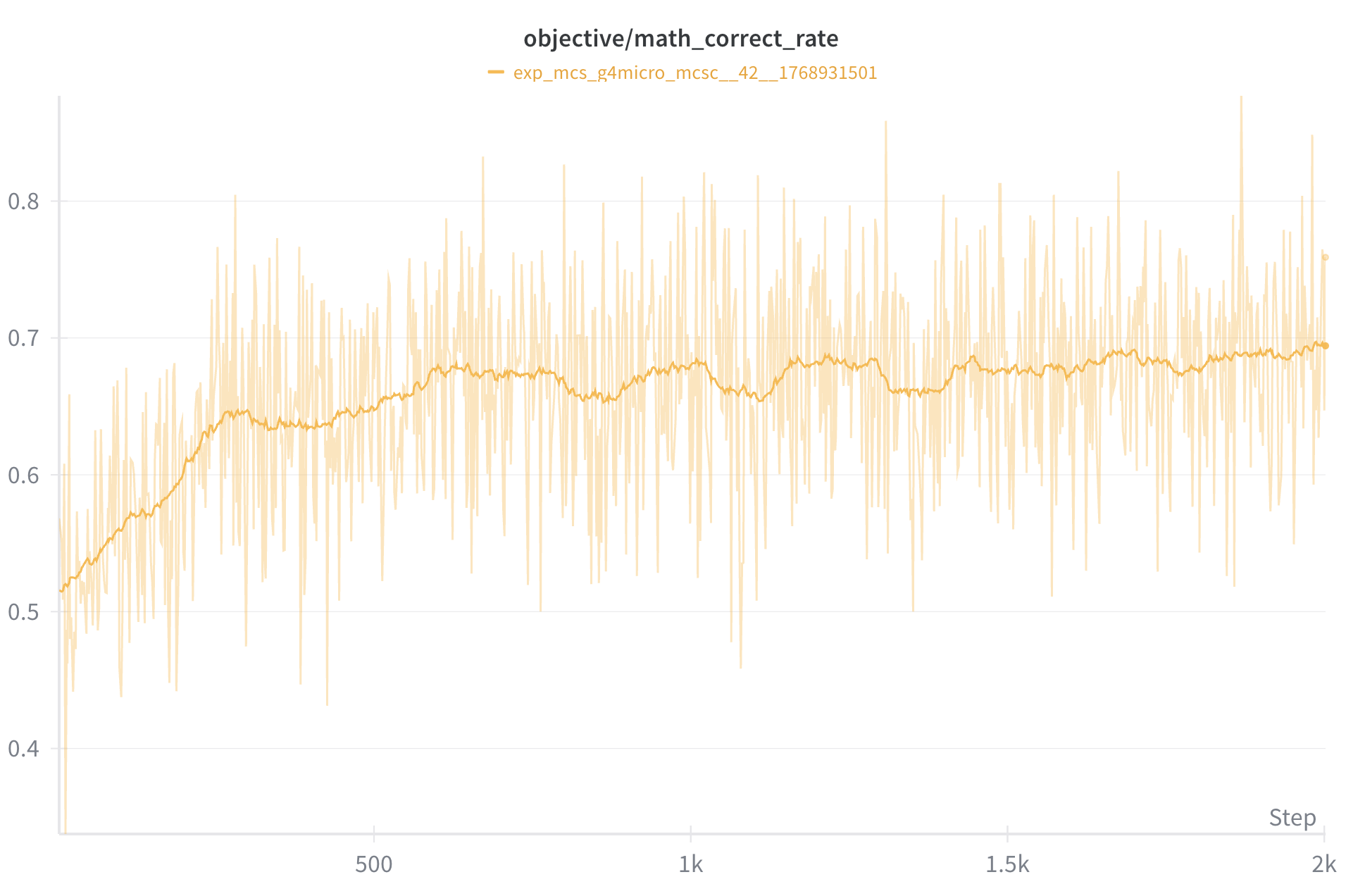}
        \caption{Math}
        \label{fig:g4micro_math_cr}
    \end{subfigure}\hfill
    \begin{subfigure}{0.24\linewidth}
        \centering
        \includegraphics[width=\linewidth]{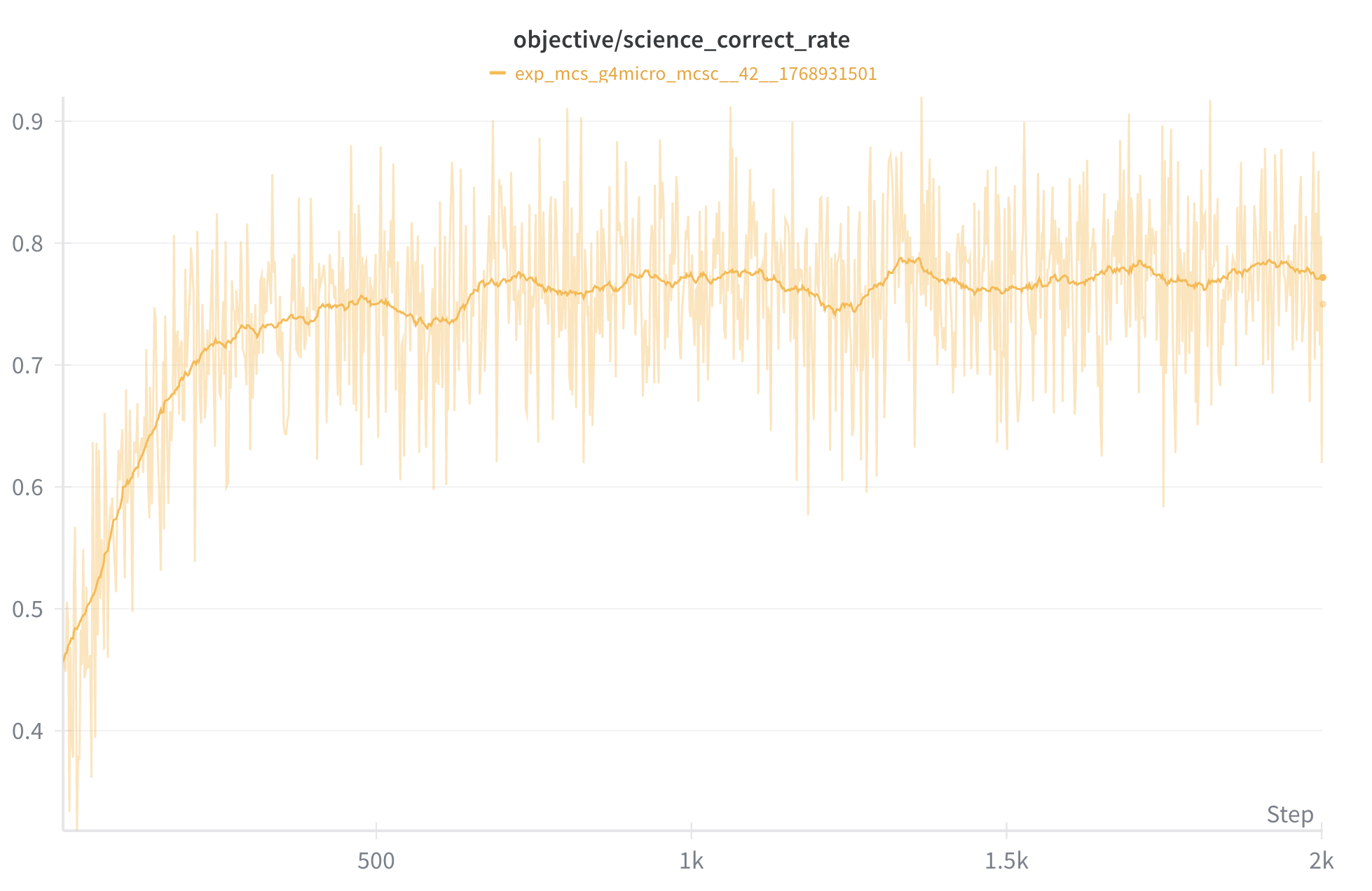}
        \caption{Science}
        \label{fig:g4micro_science_cr}
    \end{subfigure}\hfill
    \begin{subfigure}{0.24\linewidth}
        \centering
        \includegraphics[width=\linewidth]{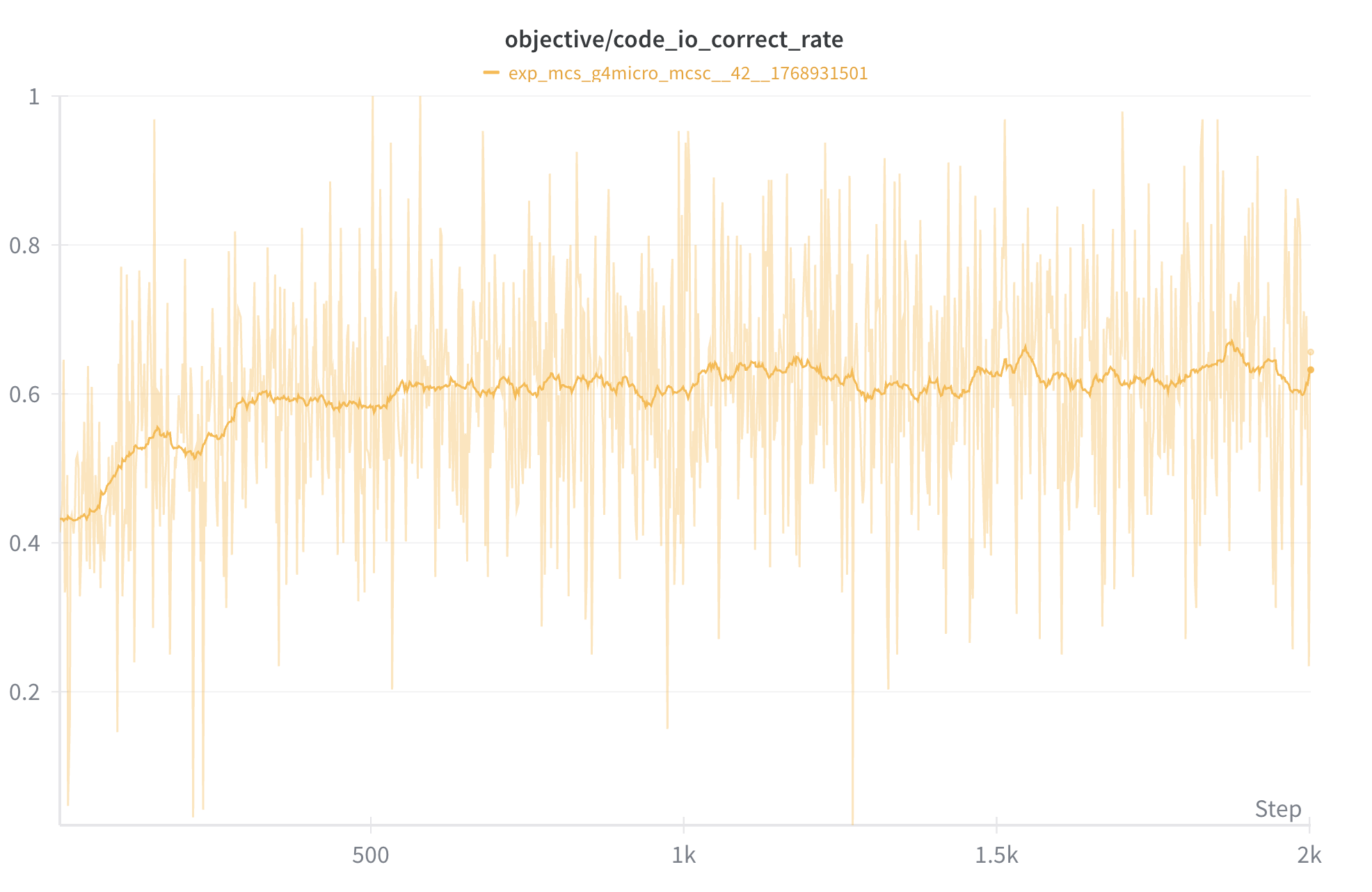}
        \caption{Code}
        \label{fig:g4micro_code_cr}
    \end{subfigure}
    \caption{\textbf{RL training correctness curves for Granite-4 Micro Dense (3B) after PRISM mid-training.} All four domains show stable, monotonically improving training dynamics over 2k steps.}
    \label{fig:g4micro_curves}
\end{figure}

Figure~\ref{fig:g4micro_curves} shows the RL training dynamics for Granite-4 Micro Dense (3B) after PRISM mid-training with the Math+Code+Science mixture. Despite being the smallest model in our study, RL training is stable across all domains. The overall verifiable correctness rate rises steadily from $\sim$0.48 to $\sim$0.72 over 2k steps (Figure~\ref{fig:g4micro_cr}). Domain-specific curves show consistent trends: math correctness climbs from $\sim$0.52 to $\sim$0.70 (Figure~\ref{fig:g4micro_math_cr}), science correctness from $\sim$0.45 to $\sim$0.78 (Figure~\ref{fig:g4micro_science_cr}), and code correctness from $\sim$0.35 to $\sim$0.65 (Figure~\ref{fig:g4micro_code_cr}). These results confirm that the PRISM mid-training recipe produces a stable foundation for RL even at the 3B scale.

%% ════════════════════════════════════════════════════════
%% F. Additional RL Learning Curves
%% ════════════════════════════════════════════════════════

\section{Additional RL Learning Curves}
\label{app:additional_rl_curves}

This section provides RL learning curves for models not shown in the main text, including PRISM RL results (Figs.~\ref{fig:mistral_code_math}--\ref{fig:g4micro_code_math}), base model RL results (Figs.~\ref{fig:llama_base}--\ref{fig:mistral7_base}), and balanced mix RL results (Figs.~\ref{fig:balanced_mistral24}--\ref{fig:balanced_g4H}).

\subsection{PRISM RL: Additional Models}

\begin{figure}[!htbp]
    \centering
    \begin{subfigure}[t]{\linewidth}
        \centering
        \includegraphics[width=\linewidth]{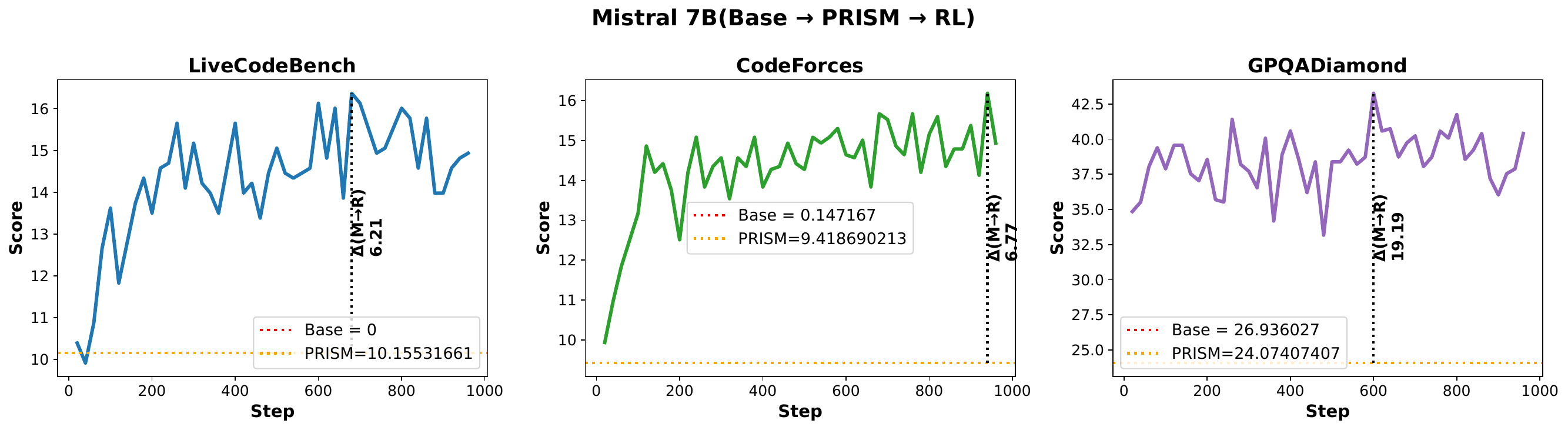}
        \caption{LiveCodeBench, Codeforces, and GPQA-Diamond over RL steps.}
        \label{fig:mistral_code}
    \end{subfigure}
    \vspace{0.6em}
    \begin{subfigure}[t]{\linewidth}
        \centering
        \includegraphics[width=\linewidth]{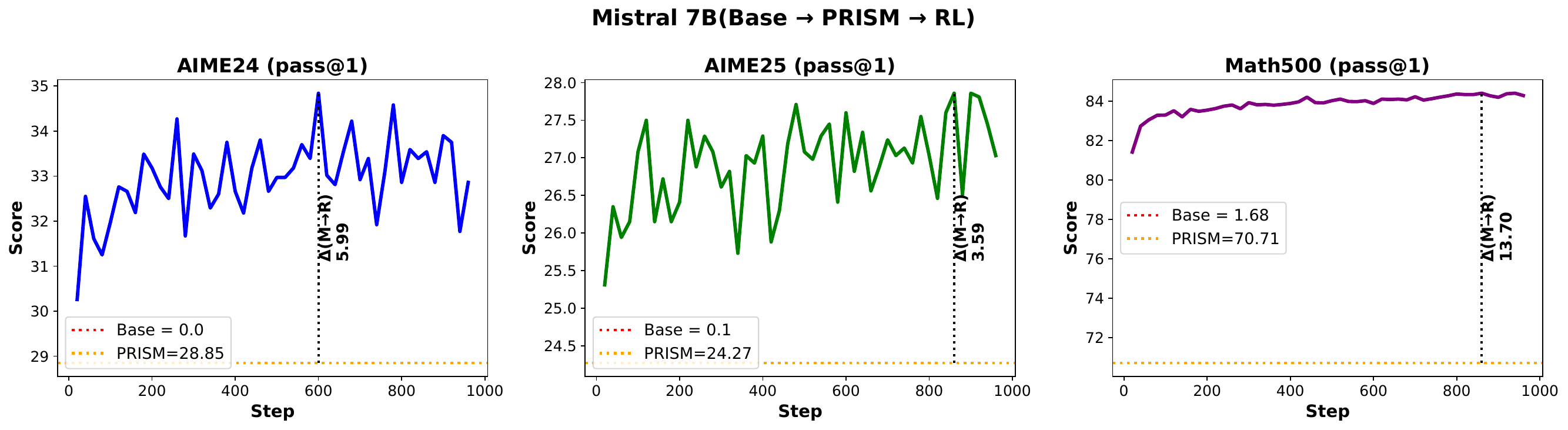}
        \caption{AIME24, AIME25, and MATH500 over RL steps.}
        \label{fig:mistral_math}
    \end{subfigure}
    \caption{\textbf{$\textsc{PRISM} \to \text{RL}$: Mistral-7B.} Consistent improvements across code and math benchmarks using the unbalanced MCS mix.}
    \label{fig:mistral_code_math}
\end{figure}

\begin{figure}[!htbp]
    \centering
    \begin{subfigure}[t]{\linewidth}
        \centering
        \includegraphics[width=\linewidth]{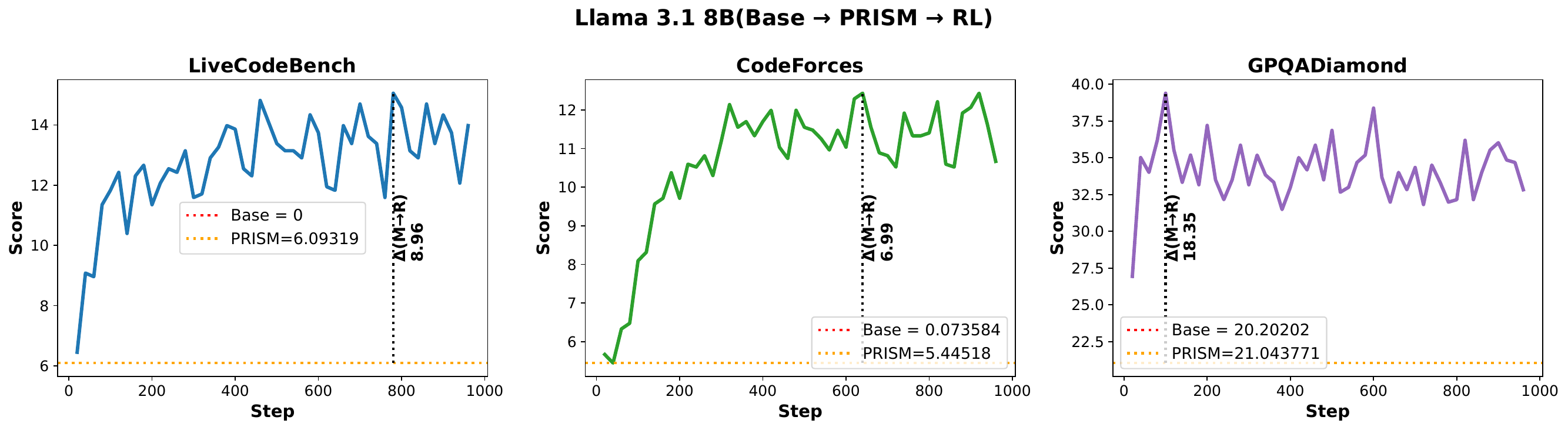}
        \caption{LiveCodeBench, Codeforces, and GPQA-Diamond over RL steps.}
        \label{fig:llama_code}
    \end{subfigure}
    \vspace{0.6em}
    \begin{subfigure}[t]{\linewidth}
        \centering
        \includegraphics[width=\linewidth]{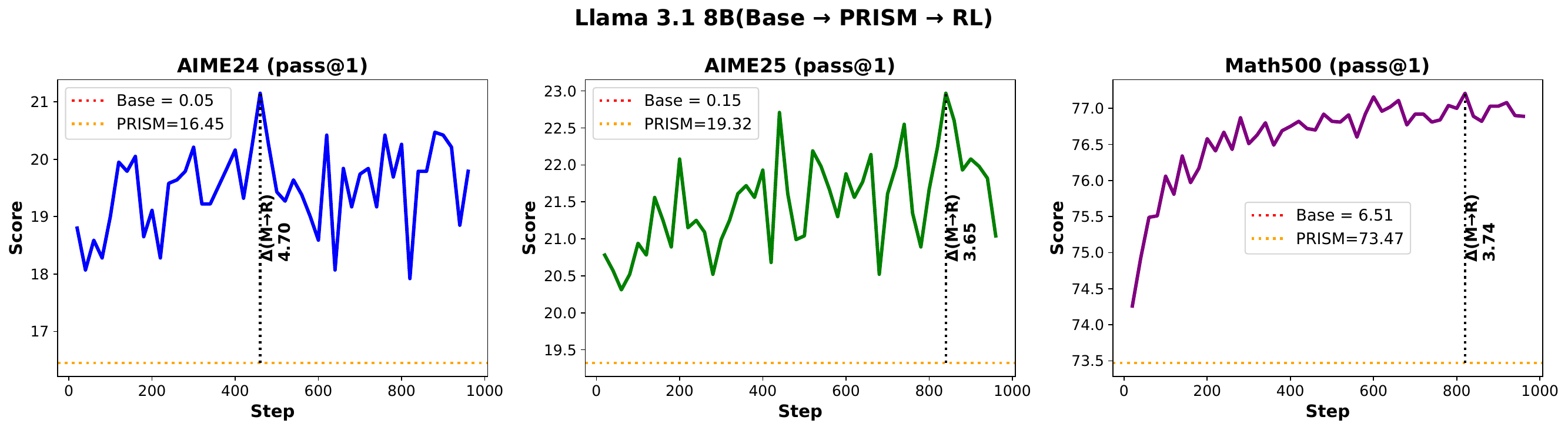}
        \caption{AIME24, AIME25, and MATH500 over RL steps.}
        \label{fig:llama_math}
    \end{subfigure}
    \caption{\textbf{$\textsc{PRISM} \to \text{RL}$: LLaMA-3.1-8B.} Stable, monotonic gains across all reasoning benchmarks.}
    \label{fig:llama_code_math}
\end{figure}

\begin{figure}[!htbp]
    \centering
    \begin{subfigure}[t]{\linewidth}
        \centering
        \includegraphics[width=\linewidth]{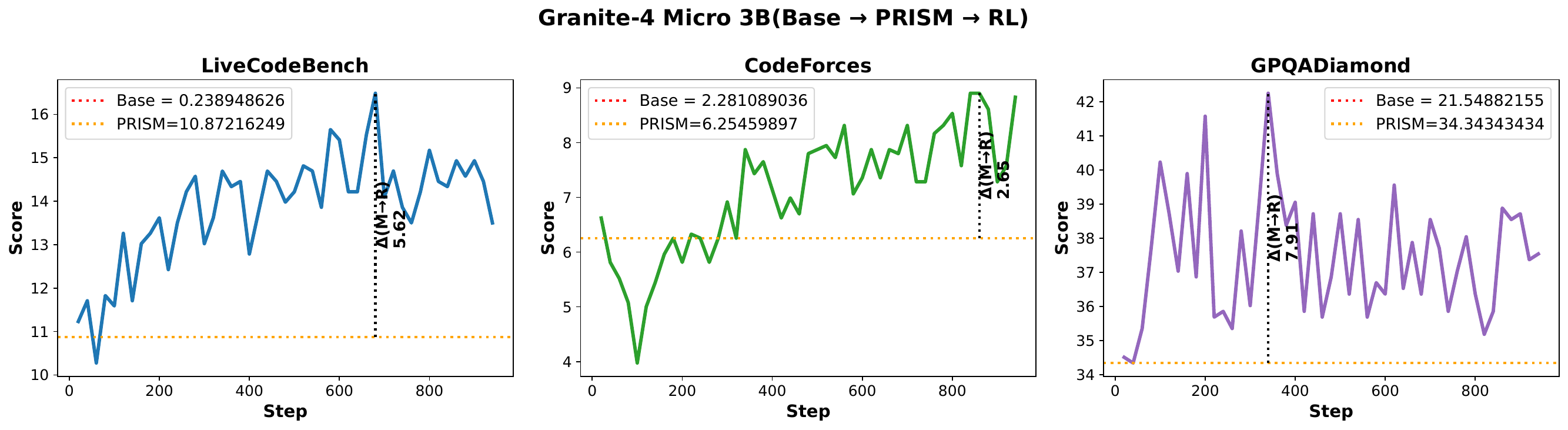}
        \caption{LiveCodeBench, Codeforces, and GPQA-Diamond over RL steps.}
        \label{fig:granite4_micro_code}
    \end{subfigure}
    \vspace{0.6em}
    \begin{subfigure}[t]{\linewidth}
        \centering
        \includegraphics[width=\linewidth]{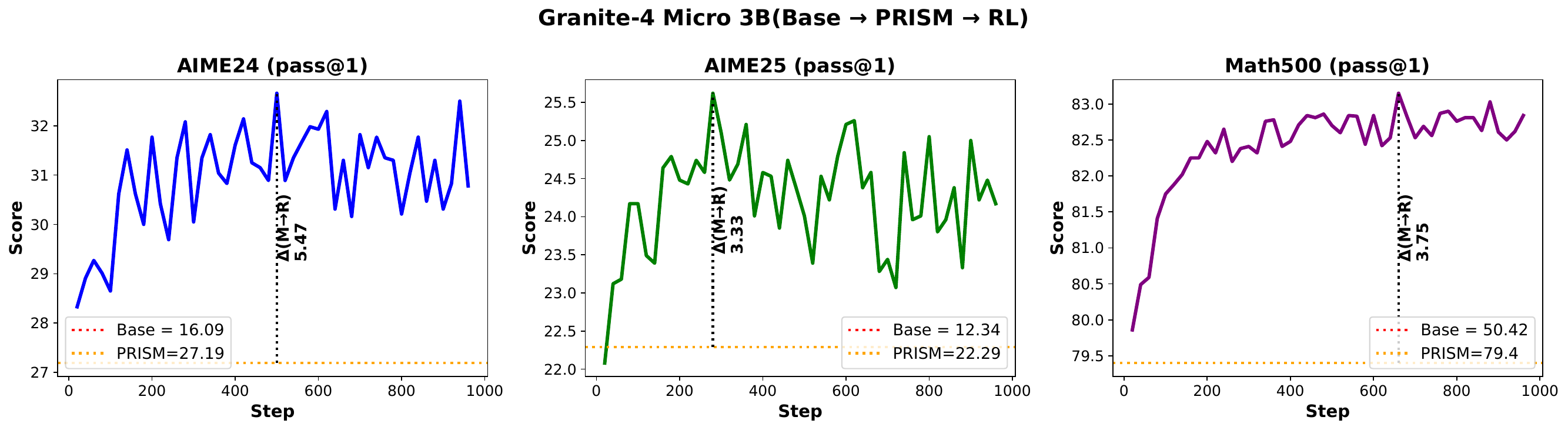}
        \caption{AIME24, AIME25, and MATH500 over RL steps.}
        \label{fig:granite4_micro_math}
    \end{subfigure}
    \caption{\textbf{$\textsc{PRISM} \to \text{RL}$: Granite-4 Micro Dense (3B).} Consistent but smaller absolute gains compared to 8B models.}
    \label{fig:g4micro_code_math}
\end{figure}

Figures~\ref{fig:mistral_code_math}--\ref{fig:g4micro_code_math} present the RL learning curves for models not featured in the main text: Mistral-7B, LLaMA-3.1-8B, and Granite-4 Micro Dense (3B). All three models follow the same pattern observed for Granite-3.3-8B in the main paper: monotonically increasing correctness rates across math, code, and science benchmarks with non-saturating trajectories, confirming that the benefits of the $\textsc{PRISM} \to \text{RL}$ pipeline generalize across model families and scales.

\subsection{RL on Base Models (No Mid-Training)}

\begin{figure}[!htbp]
    \centering
    \begin{subfigure}[t]{\linewidth}
        \centering
        \includegraphics[width=\linewidth]{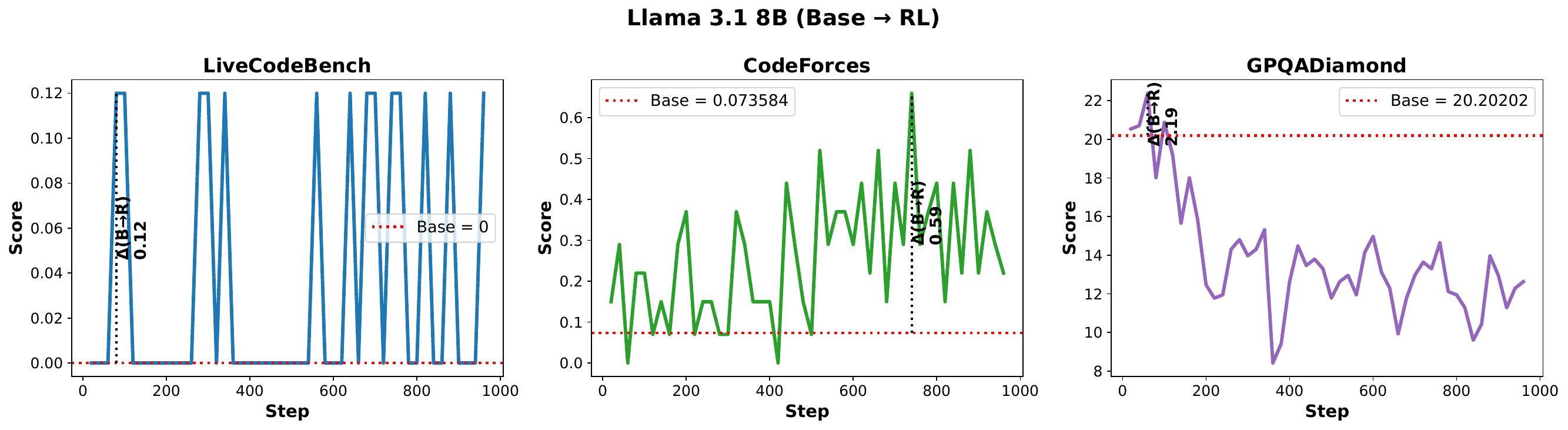}
        \caption{LiveCodeBench, Codeforces, and GPQA-Diamond over RL steps.}
    \end{subfigure}
    \vspace{0.6em}
    \begin{subfigure}[t]{\linewidth}
        \centering
        \includegraphics[width=\linewidth]{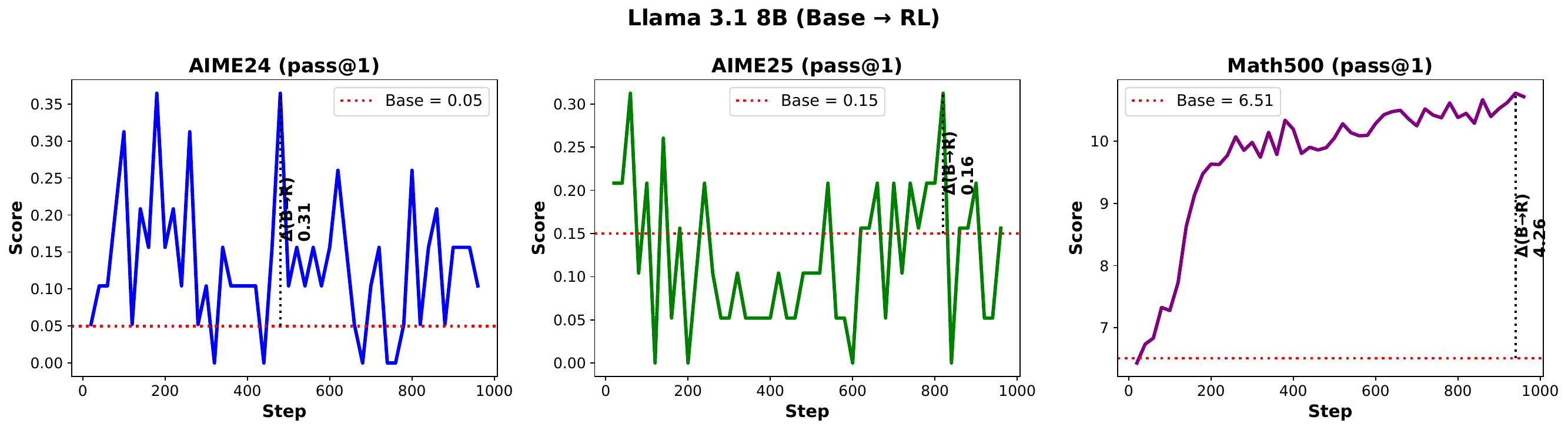}
        \caption{AIME24, AIME25, and MATH500 over RL steps.}
    \end{subfigure}
    \caption{\textbf{RL on LLaMA-3.1-8B base (no mid-training).} AIME24/25 remain near zero; GPQA-Diamond regresses below the base model.}
    \label{fig:llama_base}
\end{figure}

\begin{figure}[!htbp]
    \centering
    \begin{subfigure}[t]{\linewidth}
        \centering
        \includegraphics[width=\linewidth]{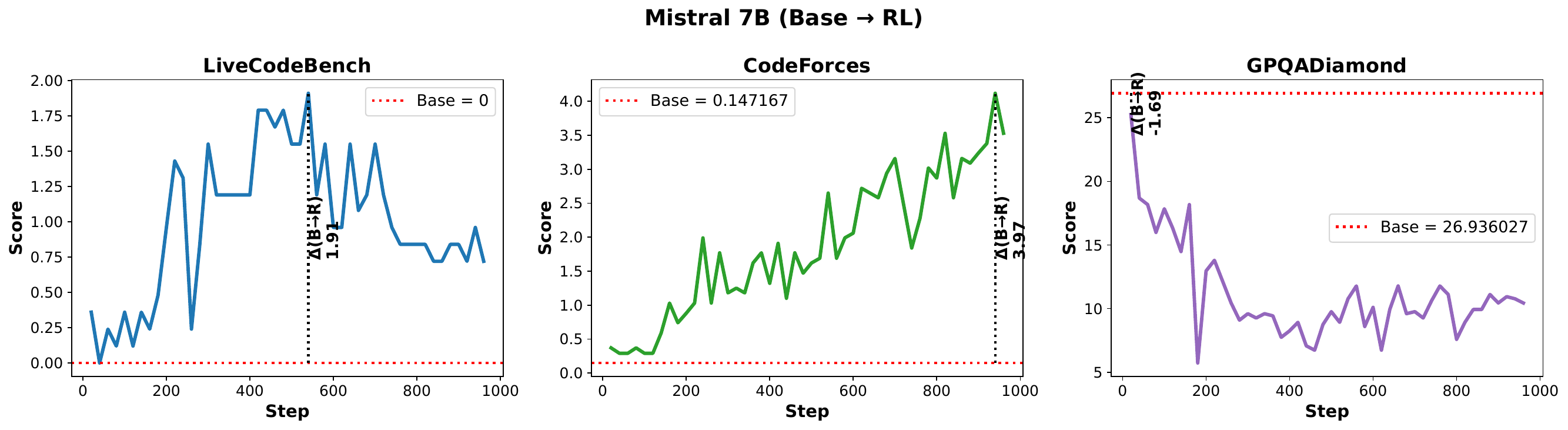}
        \caption{LiveCodeBench, Codeforces, and GPQA-Diamond over RL steps.}
    \end{subfigure}
    \vspace{0.6em}
    \begin{subfigure}[t]{\linewidth}
        \centering
        \includegraphics[width=\linewidth]{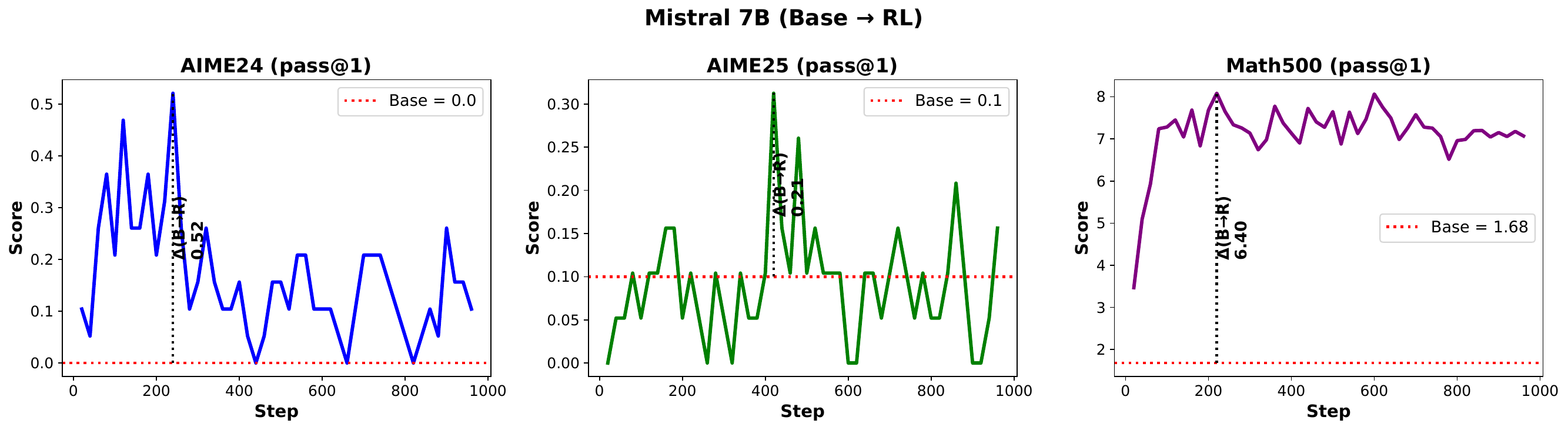}
        \caption{AIME24, AIME25, and MATH500 over RL steps.}
    \end{subfigure}
    \caption{\textbf{RL on Mistral-7B base (no mid-training).} Limited and unstable gains; AIME24/25 remain near zero throughout.}
    \label{fig:mistral7_base}
\end{figure}

Figures~\ref{fig:llama_base}--\ref{fig:mistral7_base} show what happens when RL is applied directly to base models without any mid-training. In both LLaMA-3.1-8B and Mistral-7B, AIME24 and AIME25 scores remain near zero throughout training, and GPQA-Diamond either stagnates or regresses below the base model's level. These results stand in stark contrast to the large, sustained gains observed when RL follows PRISM mid-training (Figures~\ref{fig:mistral_code_math}--\ref{fig:g4micro_code_math}), reinforcing the finding from Section~7 that mid-training provides the reasoning substrate necessary for RL to be effective.

\subsection{Balanced Mix RL: Additional Models}

\begin{figure}[!htbp]
    \centering
    \begin{subfigure}[t]{\linewidth}
        \centering
        \includegraphics[width=\linewidth]{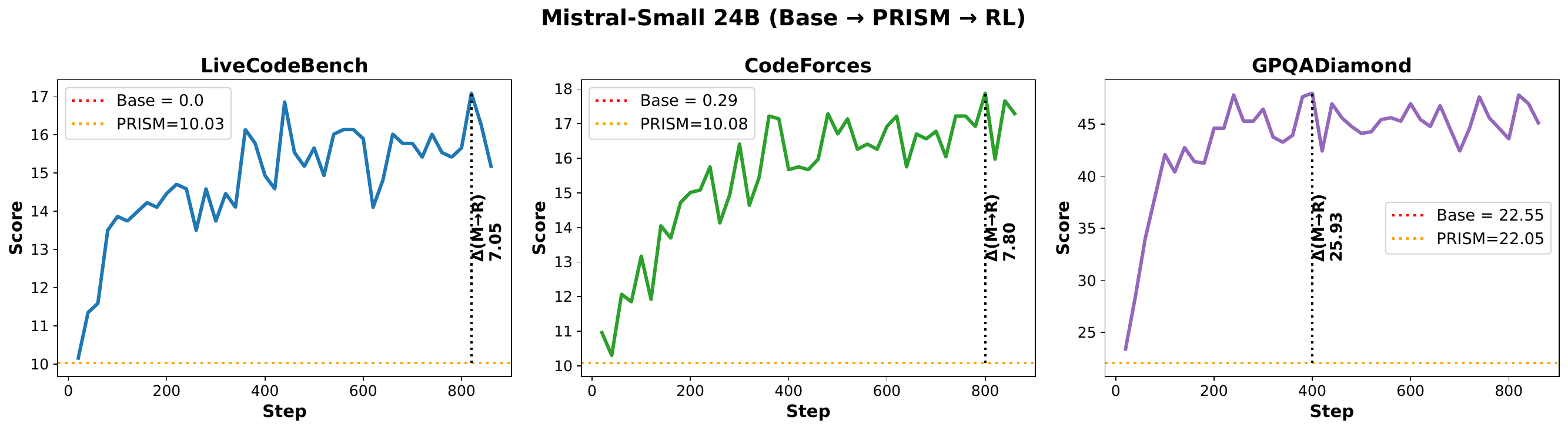}
        \caption{LiveCodeBench, Codeforces, and GPQA-Diamond over RL steps.}
    \end{subfigure}
    \vspace{0.6em}
    \begin{subfigure}[t]{\linewidth}
        \centering
        \includegraphics[width=\linewidth]{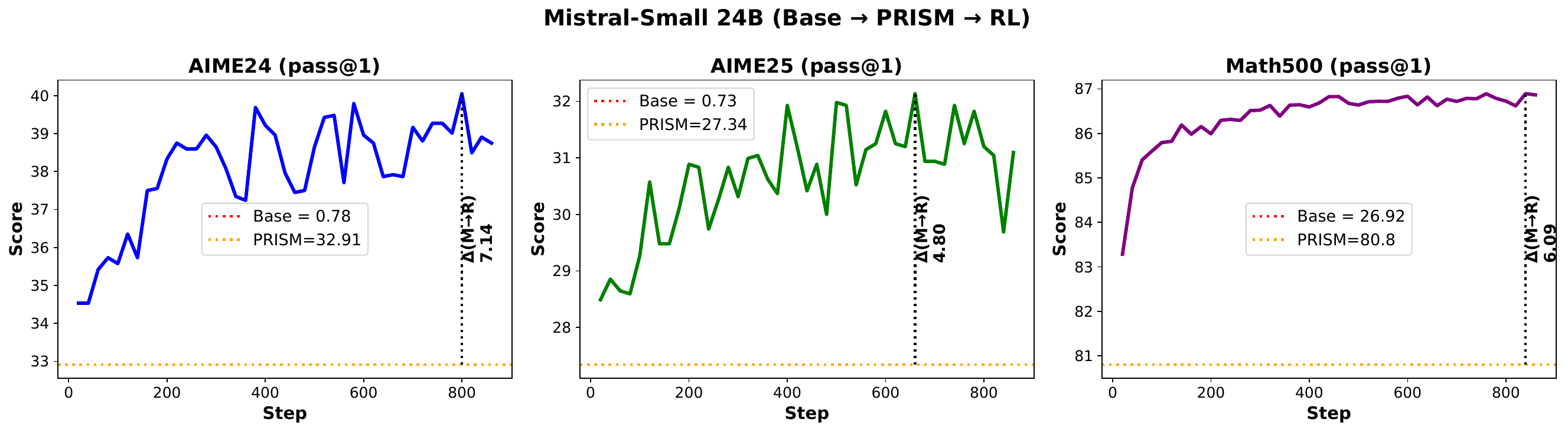}
        \caption{AIME24, AIME25, and MATH500 over RL steps.}
    \end{subfigure}
    \caption{\textbf{$\textsc{PRISM} \to \text{RL}$ with balanced mix: Mistral-Small 24B.} GPQA-Diamond gain (+25.93) slightly lower than unbalanced mix.}
    \label{fig:balanced_mistral24}
\end{figure}

\begin{figure}[!htbp]
    \centering
    \begin{subfigure}[t]{\linewidth}
        \centering
        \includegraphics[width=\linewidth]{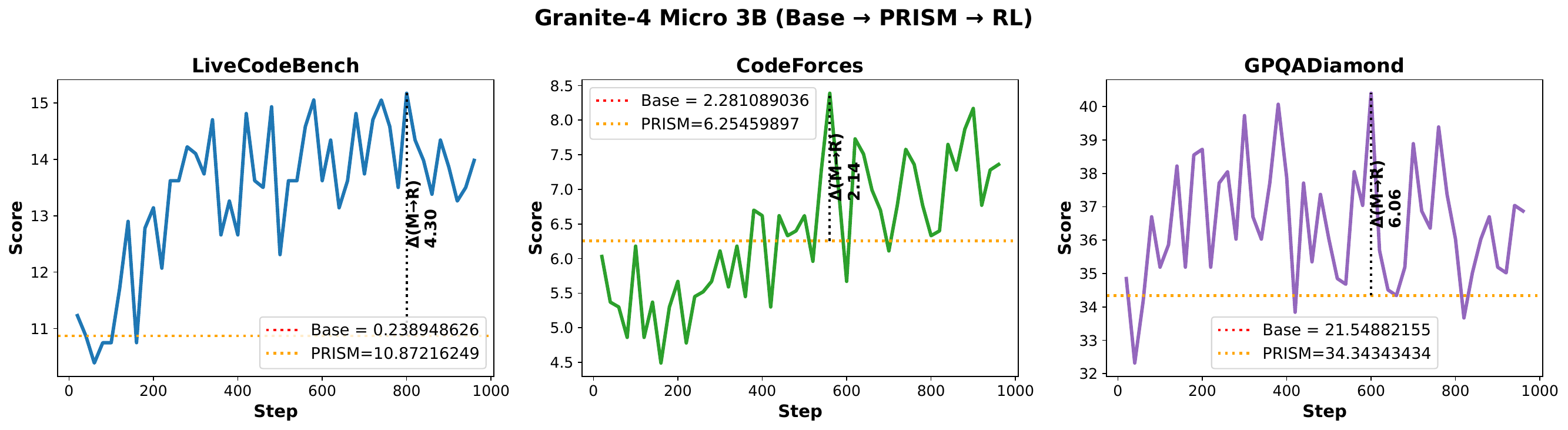}
        \caption{LiveCodeBench, Codeforces, and GPQA-Diamond over RL steps.}
    \end{subfigure}
    \vspace{0.6em}
    \begin{subfigure}[t]{\linewidth}
        \centering
        \includegraphics[width=\linewidth]{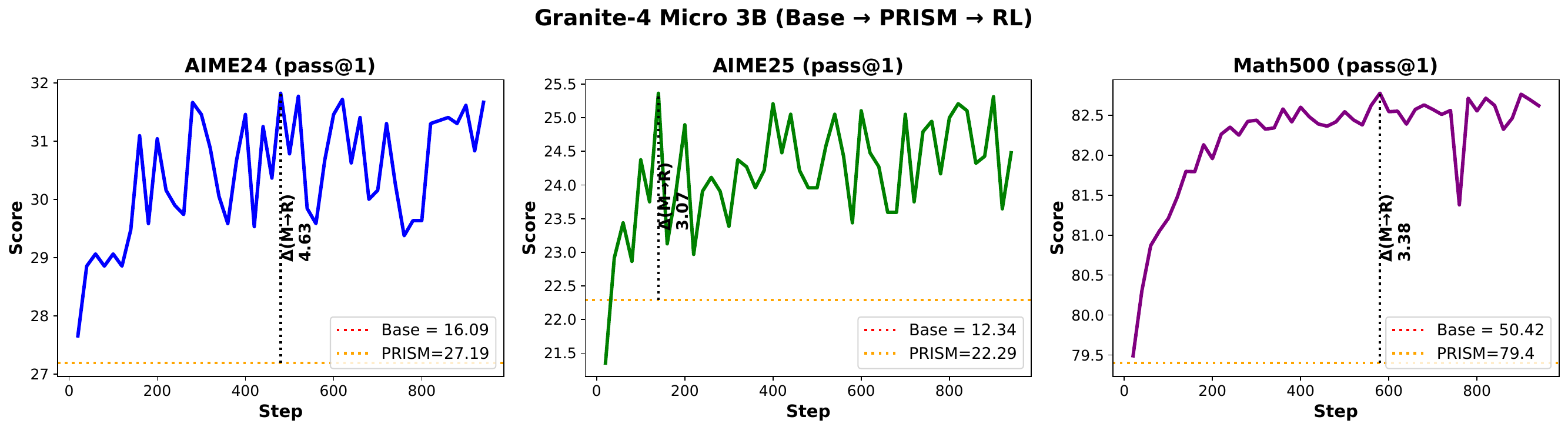}
        \caption{AIME24, AIME25, and MATH500 over RL steps.}
    \end{subfigure}
    \caption{\textbf{$\textsc{PRISM} \to \text{RL}$ with balanced mix: Granite-4 Micro Dense (3B).} Code and GPQA-Diamond gains are notable (+4.30 LCB, +6.06 GPQA).}
    \label{fig:balanced_g4dense}
\end{figure}

\begin{figure}[!htbp]
    \centering
    \begin{subfigure}[t]{\linewidth}
        \centering
        \includegraphics[width=\linewidth]{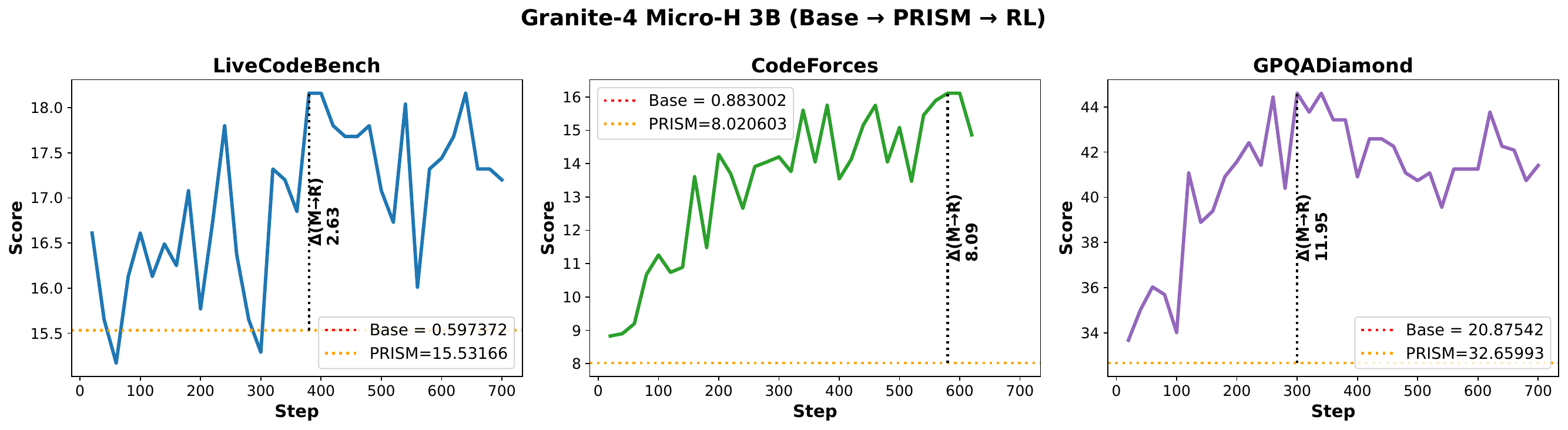}
        \caption{LiveCodeBench, Codeforces, and GPQA-Diamond over RL steps.}
    \end{subfigure}
    \vspace{0.6em}
    \begin{subfigure}[t]{\linewidth}
        \centering
        \includegraphics[width=\linewidth]{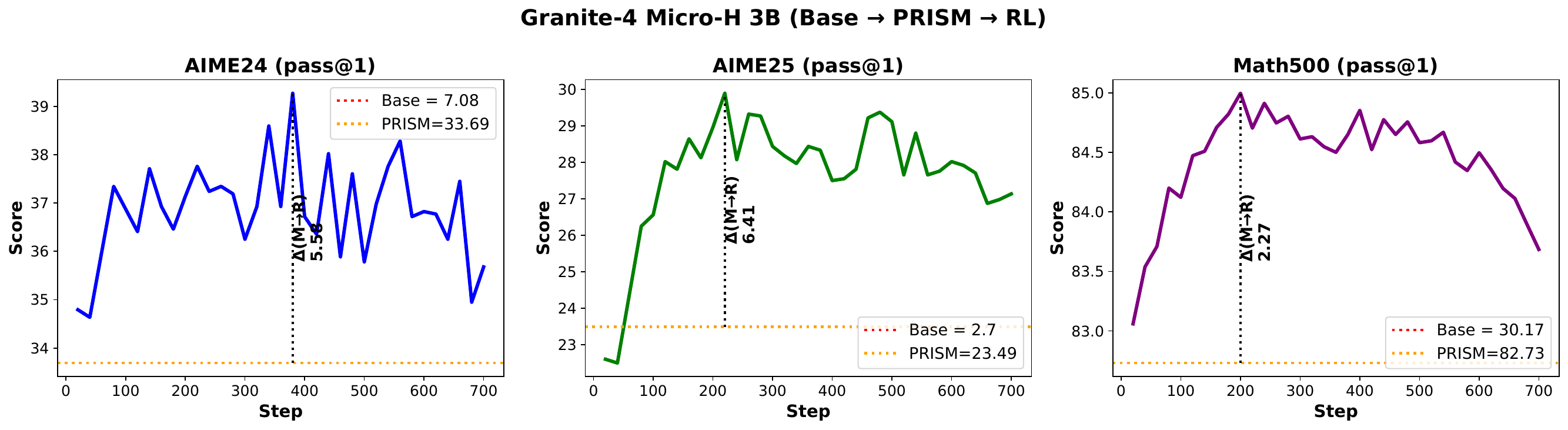}
        \caption{AIME24, AIME25, and MATH500 over RL steps.}
    \end{subfigure}
    \caption{\textbf{$\textsc{PRISM} \to \text{RL}$ with balanced mix: Granite-4-H Micro (Hybrid, 3B).} Largest balanced-mix gains among small models, with +8.09 on Codeforces and +11.95 on GPQA-Diamond.}
    \label{fig:balanced_g4H}
\end{figure}

Figures~\ref{fig:balanced_mistral24}--\ref{fig:balanced_g4H} present RL learning curves when using the balanced RL data mix (equal math, code, and science sampling) for models not shown in the main text. The balanced mix produces comparable overall gains to the unbalanced mix across most models, with modest differences in domain-specific performance. Notably, Granite-4-H Micro (Figure~\ref{fig:balanced_g4H}) achieves the largest balanced-mix improvements among small models, with +8.09 on Codeforces and +11.95 on GPQA-Diamond, suggesting that hybrid architectures respond well to domain-balanced RL training.

\section{Sparsity Threshold Sensitivity}
\label{app:threshold-sensitivity}

Table~\ref{tab:threshold-sensitivity} shows that the dense/sparse asymmetry between mid-training and RL holds at every threshold from 0.1\% to 10\%. At the 1\% threshold used in the main text, RL leaves 95.0\% of Granite-3.3 parameters unchanged; even at 0.1\%, 82.3\% remain unchanged. Conversely, mid-training changes the vast majority of parameters at all thresholds. The conclusions in Section~\ref{sec:weight-analysis} are robust to the choice of threshold.

\begin{table}[h]
\centering
\small
\setlength{\tabcolsep}{5pt}
\begin{tabular}{c c c}
\toprule
\textbf{Threshold (\%)} & \textbf{Base$\to$MT unchanged} & \textbf{MT$\to$RL unchanged} \\
\midrule
0.1 & 2.1\% & 82.3\% \\
0.5 & 2.4\% & 87.3\% \\
\textbf{1.0} & \textbf{3.5\%} & \textbf{95.0\%} \\
2.0 & 5.3\% & 97.5\% \\
5.0 & 11.1\% & 99.0\% \\
10.0 & 20.1\% & 99.5\% \\
\bottomrule
\end{tabular}
\caption{\textbf{Per-element sparsity at different relative change thresholds (Granite-3.3, 8B).} The dense/sparse asymmetry between mid-training and RL is robust across all tested thresholds. Bold row shows the 1\% threshold used in the main text.}
\label{tab:threshold-sensitivity}
\end{table}

%% ════════════════════════════════════════════════════════
%% G. RL Sub-component Weight Analysis
%% ════════════════════════════════════════════════════════

\section{RL Sub-component Weight Analysis}
\label{app:subcomponent}

We decompose the weight divergence analysis from Section~\ref{sec:weight-analysis} at a finer granularity, breaking each component into individual weight matrices: Q, K, V, O projections for attention; gate, up, down projections for MLP; and in\_proj, out\_proj, conv1d, A, dt for Mamba layers. Table~\ref{tab:subcomponent-rl} reports the fraction of parameters changed ($>$1\% relative change) for each sub-component across four conditions: MT$\to$RL and Base$\to$RL (no mid-training) for both Granite-3.3 and Nemotron-H.

\begin{table*}[t]
\centering
\small
\setlength{\tabcolsep}{4pt}
\begin{tabular}{l c c c c c c c c}
\toprule
& \multicolumn{4}{c}{\textbf{Granite-3.3 (8B)}} & \multicolumn{4}{c}{\textbf{Nemotron-H (8B)}} \\
\cmidrule(lr){2-5} \cmidrule(lr){6-9}
\textbf{Sub-component}
& \textbf{Base$\to$MT} & \textbf{MT$\to$RL} & \textbf{Base$\to$RL} & \textbf{$\Delta$}
& \textbf{Base$\to$MT} & \textbf{MT$\to$RL} & \textbf{Base$\to$RL} & \textbf{$\Delta$} \\
\midrule
Attn: Q       & 83.5\% & 4.4\% & 5.2\% & +0.8 & 97.5\% & 5.5\% & 4.9\% & $-$0.6 \\
Attn: K       & 83.2\% & 4.6\% & 5.2\% & +0.6 & 96.9\% & 5.5\% & 4.8\% & $-$0.7 \\
Attn: V       & 97.9\% & 5.7\% & 7.5\% & +1.8 & 97.1\% & 8.5\% & 7.6\% & $-$0.9 \\
Attn: O       & 98.0\% & 5.6\% & 6.7\% & +1.1 & 97.2\% & 7.2\% & 6.4\% & $-$0.8 \\
\midrule
MLP: gate     & 98.3\% & 5.4\% & 6.1\% & +0.7 & -- & -- & -- & -- \\
MLP: up       & 97.9\% & 4.9\% & 5.5\% & +0.6 & -- & -- & -- & -- \\
MLP: down     & 98.0\% & 5.0\% & 5.6\% & +0.6 & -- & -- & -- & -- \\
\midrule
Mamba: in\_proj  & -- & -- & -- & -- & 97.8\% & 5.9\% & 5.5\% & $-$0.4 \\
Mamba: out\_proj & -- & -- & -- & -- & 97.8\% & 6.5\% & 5.9\% & $-$0.6 \\
Mamba: conv1d   & -- & -- & -- & -- & 91.7\% & 17.4\% & 17.7\% & +0.3 \\
Mamba: A/dt     & -- & -- & -- & -- & 97.5\% & 0.0\% & 0.0\% & 0.0 \\
\midrule
Embed/LM-head & 97.5\% & 2.4\% & 2.6\% & +0.2 & 97.9\% & 7.3\% & 6.7\% & $-$0.6 \\
Norm          & 11.9\% & 1.2\% & 0.1\% & $-$1.1 & 61.3\% & 0.1\% & 0.1\% & 0.0 \\
\bottomrule
\end{tabular}
\caption{\textbf{Sub-component weight analysis: \% of parameters changed ($>$1\% relative change) across pipeline transitions.} $\Delta$ = difference between Base$\to$RL and MT$\to$RL. The near-zero $\Delta$ values confirm that RL targets the same sub-components in the same proportions regardless of starting point. Value and output projections are consistently the most affected by RL across both architectures, while Mamba parameters (A, dt) and norms remain frozen.}
\label{tab:subcomponent-rl}
\end{table*}

\begin{figure*}[t]
    \centering
    \includegraphics[width=\textwidth]{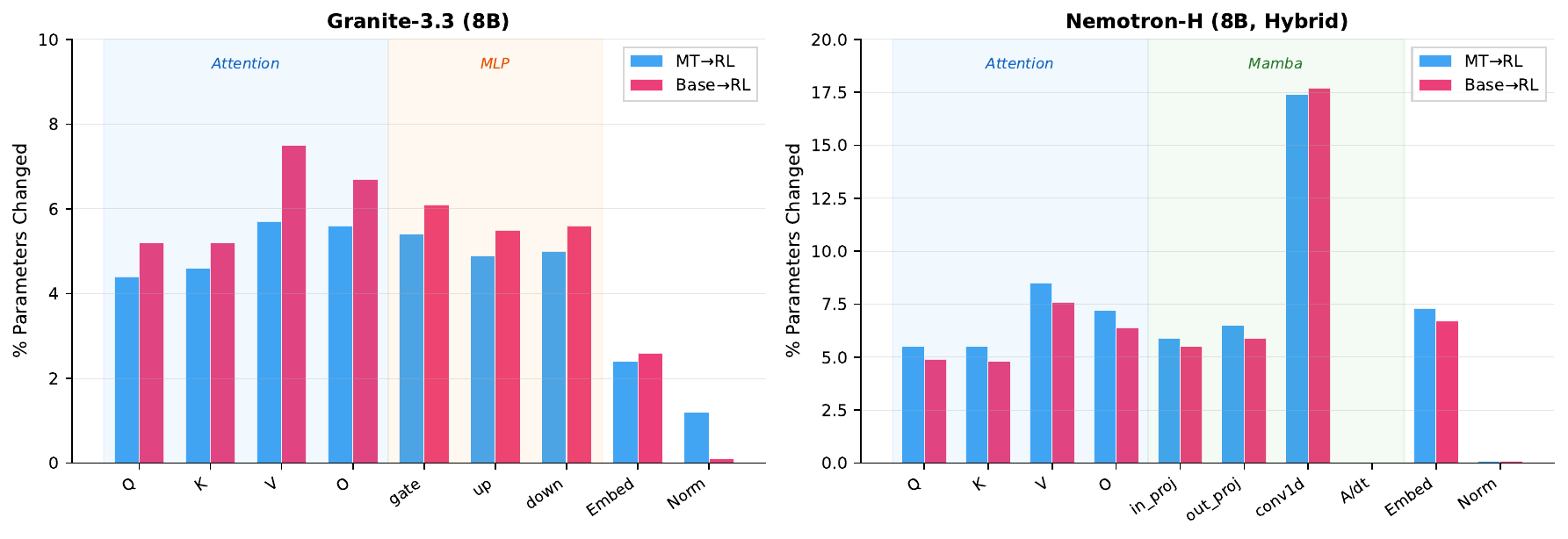}
    \caption{\textbf{RL targets the same sub-components regardless of starting point.} Fraction of parameters changed ($>$1\% relative change) during RL for Granite-3.3 (left) and Nemotron-H (right), comparing MT$\to$RL (blue) vs.\ Base$\to$RL (pink). The near-identical distributions confirm that RL's sub-component targeting is intrinsic to the optimization, not a consequence of mid-training.}
    \label{fig:subcomponent-rl}
\end{figure*}

Figure~\ref{fig:subcomponent-rl} visualizes the MT$\to$RL and Base$\to$RL distributions side by side. Three findings emerge. First, RL's sub-component targeting is \emph{identical} regardless of whether mid-training preceded it: the $\Delta$ column shows differences of at most 1.8 percentage points, with most below 1 point. Second, value (V) and output (O) projections are consistently the most affected sub-components during RL, in both Granite-3.3 (5.7\%, 5.6\%) and Nemotron-H (8.5\%, 7.2\%), suggesting that RL preferentially adjusts how models read from and write to the residual stream. Third, Mamba parameters (A, dt) are completely frozen during RL ($<$0.1\% changed), while the learned projection matrices (in\_proj, out\_proj) change at rates comparable to attention projections, indicating that RL respects the architectural priors encoded in the SSM state dynamics.

%% ════════════════════════════════════════════════════════
%% H. Extended CKA Analysis
%% ════════════════════════════════════════════════════════

\section{Extended CKA Representation Analysis}
\label{app:cka-extended}

This section provides the full CKA representation similarity analysis across three models (Granite-3.3, LLaMA-3.1, Nemotron-H) and three input distributions (Wikipedia, C4, GSM8K), complementing the main-text results in Section~\ref{sec:cka-analysis}. All experiments use 200 prompts per input type and batch-size-1 encoding. Each figure contains three panels corresponding to the three input types. In each panel, the $x$-axis is the layer index and the $y$-axis is the linear CKA score~\cite{kornblith2019similarityneuralnetworkrepresentations} between the mean-pooled hidden representations. Three pairwise comparisons are shown: Base vs.\ MT (blue), Base vs.\ RL (pink), and MT vs.\ RL (green). CKA$=$1.0 indicates identical representational geometry; lower values indicate greater divergence.

\begin{figure*}[t]
    \centering
    \includegraphics[width=\textwidth]{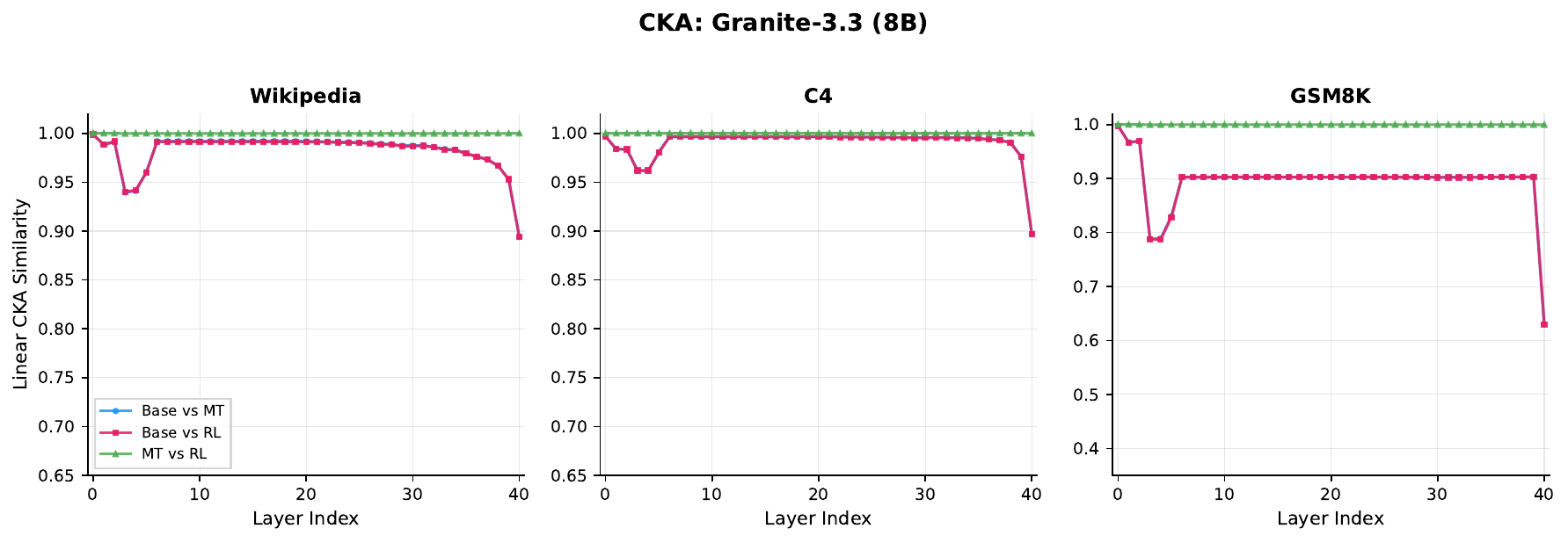}
    \caption{\textbf{CKA across input distributions: Granite-3.3 (8B).} MT vs.\ RL (green) remains $\approx$1.0 on all three inputs. Base vs.\ MT divergence is input-dependent: strongest on GSM8K math prompts (min 0.55), weakest on C4 general text (min 0.94).}
    \label{fig:cka-app-g33}
\end{figure*}

\begin{figure*}[t]
    \centering
    \includegraphics[width=\textwidth]{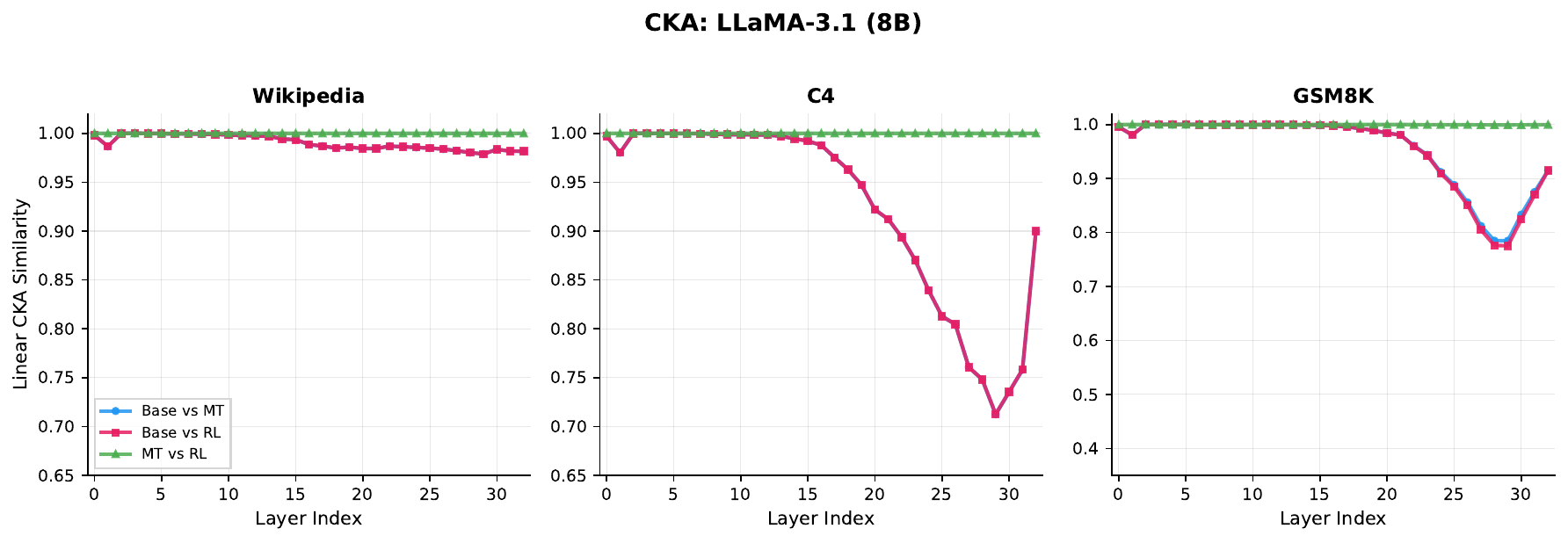}
    \caption{\textbf{CKA across input distributions: LLaMA-3.1 (8B).} Same pattern as Granite-3.3: MT vs.\ RL $\approx$1.0 everywhere, confirming the finding generalizes across dense Transformer families.}
    \label{fig:cka-app-llama}
\end{figure*}

\begin{figure*}[t]
    \centering
    \includegraphics[width=\textwidth]{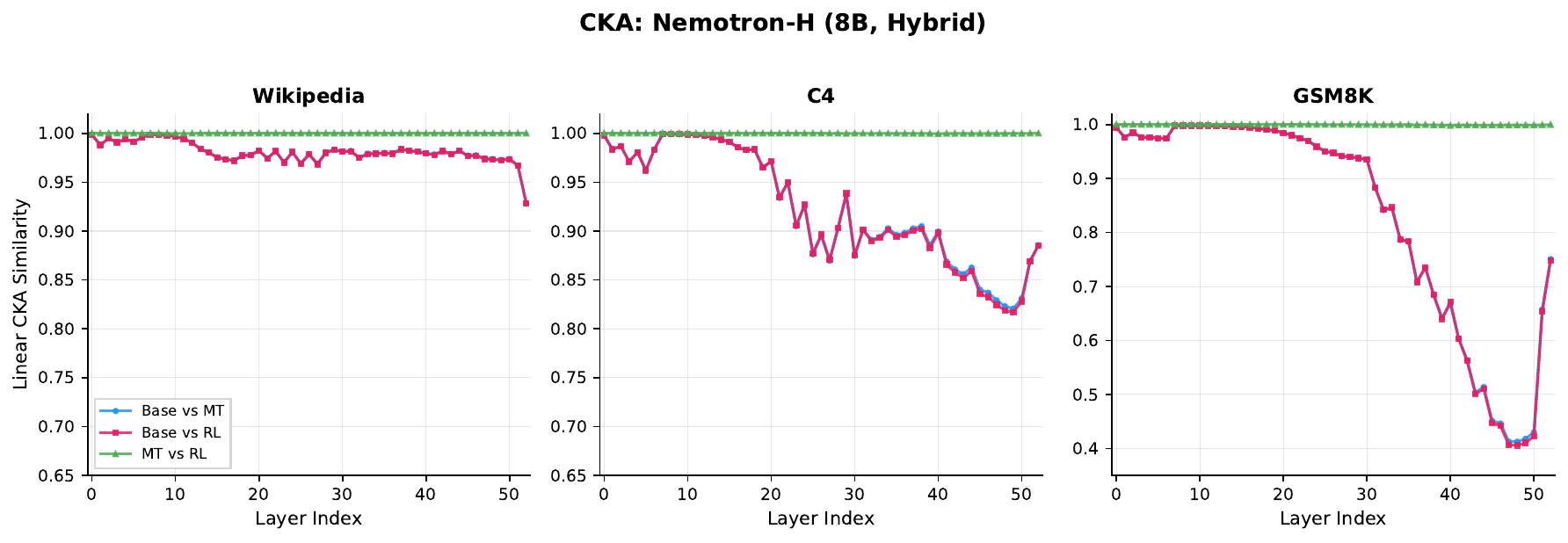}
    \caption{\textbf{CKA across input distributions: Nemotron-H (8B, Hybrid).} MT vs.\ RL $>$0.998 on all three inputs, consistent with all other models. Base vs.\ MT divergence is most pronounced on GSM8K math prompts (min $\approx$0.41), reflecting mid-training's targeted restructuring of reasoning-relevant representations.}
    \label{fig:cka-app-nemotron}
\end{figure*}

\paragraph{\textbf{All models (Figures~\ref{fig:cka-app-g33}, \ref{fig:cka-app-llama}, \ref{fig:cka-app-nemotron}).}}
The one consistent finding across all three models is that MT vs.\ RL remains $>$0.998 at every layer across all three input distributions, for both dense Transformers and hybrid attention-Mamba architectures. The Base vs.\ MT divergence pattern, however, is model-specific: Granite-3.3 shows its deepest dip at the output layer on GSM8K; LLaMA-3.1 shows its deepest dip on C4 web text rather than math prompts; and Nemotron-H shows broader divergence across later middle layers. Since each model has a different pretraining distribution, the representational effects of mid-training cannot be universally characterized, they depend on what the base model already learned. Base vs.\ MT and Base vs.\ RL are nearly identical in all cases, confirming that RL contributes no additional representational shift.

%% ════════════════════════════════════════════════════════
%% I. AIME 2026 Evaluation
%% ════════════════════════════════════════════════════════

\section{AIME 2026 Evaluation}
\label{app:aime26}

\begin{figure}[t]
    \centering
    \includegraphics[width=0.85\linewidth]{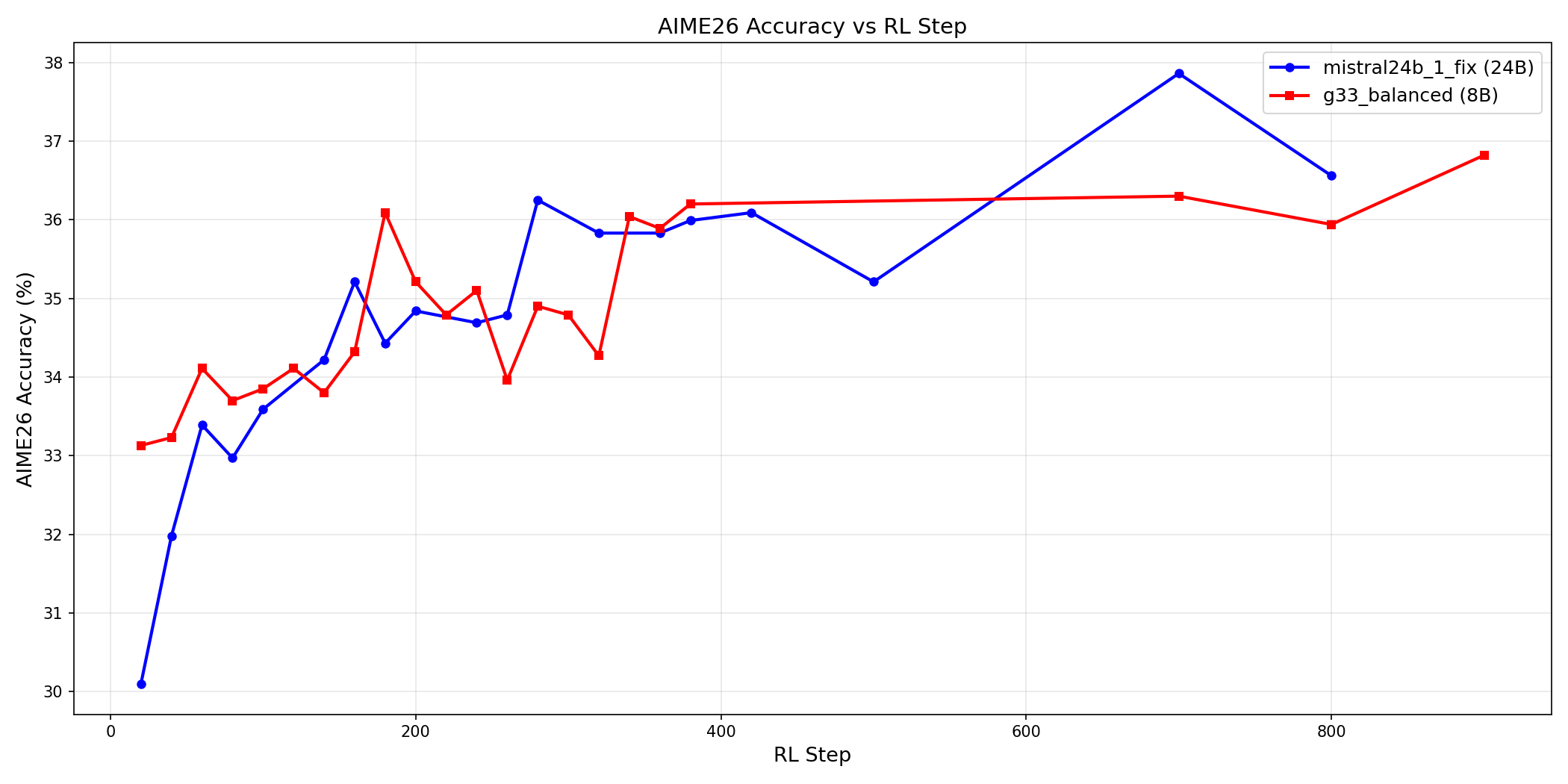}
    \caption{\textbf{AIME 2026 accuracy over RL training steps.} Both Granite-3.3 (8B) and Mistral-Small (24B) show steady gains on this recently released benchmark, confirming generalization of the $\textsc{PRISM} \to \text{RL}$ pipeline.}
    \label{fig:aime26}
\end{figure}

To test generalization to a recently released benchmark, we evaluate two PRISM mid-trained models on AIME 2026~\citep{maa2026aime}, which was published after the completion of all our training runs. Figure~\ref{fig:aime26} shows AIME26 accuracy across RL training steps for Granite-3.3 (8B) and Mistral-Small (24B). Both models show consistent improvement over RL training: Granite-3.3 improves from $\sim$33\% to $\sim$37\%, and Mistral-Small from $\sim$30\% to $\sim$38\%. These results confirm that the gains from the $\textsc{PRISM} \to \text{RL}$ pipeline transfer to held-out math benchmarks unseen during training.

%% ════════════════════════════════════════════════════════
%% H. Model Generations
%% ════════════════════════════════════════════════════════

\section{Model Generations}
\label{app:model_generations}

This section provides qualitative examples of model outputs at each stage of the PRISM pipeline: base model, after mid-training, and after RL. We show generations from Granite-3.3-8B on representative math, code, and science prompts. These examples illustrate how mid-training introduces structured reasoning (e.g., step-by-step problem decomposition) that is absent in the base model, and how RL further refines the reasoning chains with more accurate and complete solutions.

\textbf{Note on failure modes.} Some base model and mid-trained model generations exhibit repetitive or looping outputs. We distinguish three qualitatively different failure modes observed in these examples: (1) \textit{Base model loops}: the base model lacks instruction fine-tuning and may produce repetitive token sequences when it lacks a clear completion signal; (2) \textit{Mid-trained model loops}: the mid-trained model has learned chain-of-thought formatting but may enter repetitive patterns on out-of-distribution prompt structures where the reasoning format does not terminate cleanly; (3) \textit{RL model}: RL applies stop-token and repetition penalties during training, so the RL-trained model consistently terminates outputs correctly. These failure modes are expected and illustrate why formatting rewards and repetition penalties are important components of the RL training recipe.

\newtcblisting{myappendix}{
  arc=0mm,
  breakable,
  enhanced,
  listing only,
  colback=gray!5,
  colframe=gray!50,
  fonttitle=\bfseries,
  coltitle=black,
  colbacktitle=gray!30,
  listing options={
    basicstyle=\ttfamily\small,
    breaklines=true,
    breakindent=0pt,       % <--- FIX 1: Removes indent on wrapped lines
    columns=fullflexible,
    keepspaces=true,
    showstringspaces=false,
    gobble=0               % <--- FIX 2: Increase this (e.g., 2 or 4) if your source code is indented
  }
}

% \newtcblisting{myappendix}[2]{ % [2] means two arguments: {Title}{Label}
%   arc=0mm,
%   breakable,
%   enhanced,
%   listing only,
%   colback=gray!5,
%   colframe=gray!50,
%   title={#1},              % First argument is the title
%   label={#2},              % Second argument is the label (e.g., app:math_prompt)
%   fonttitle=\bfseries,
%   coltitle=black,
%   colbacktitle=gray!30,
%   listing options={
%     basicstyle=\ttfamily\small,
%     breaklines=true,
%     breakindent=0pt,
%     columns=fullflexible,
%     keepspaces=true
%   }
% }

\label{app:model_generations}

\subsection{Math Prompt}
\label{app:math_prompt}
\begin{myappendix}
In a car dealership with some cars, 60% of the cars are hybrids, and 40% of the hybrids contain only one headlight. There are 216 hybrids with full headlights. How many cars are there in the dealership?

Please put your final answer in \\boxed{} format.
\end{myappendix}

\subsection{Granite 3.3 8b base math generation}
\label{app:g33_base_math}
The correct answer for the above problem is 600; however, the base model gets it wrong probably because it rushes to the final answer (using \textbf{192 tokens}) without reasoning about the possible steps.

\begin{myappendix}
Let's solve this problem using Python's sympy library.
```python
import sympy as sp

# let's denote the number of cars as x
x = sp.symbols('x')

# 60% of the cars are hybrids
hybrids = 0.6 * x

# 40% of the hybrids contain only one headlight
hybrids_with_one_headlight = 0.4 * hybrids

# there are 216 hybrids with full headlights
eq = sp.Eq(hybrids - hybrids_with_one_headlight, 216)

# solving for x to get the number of cars
sp.solve(eq, x)[0]
```
Thus there are \boxed{360} cars in the dealership.
\end{myappendix}

\subsection{Granite 3.3 8b midtrain math generation}
\label{app:g33_mt_math}
Owing to mid-training, the base model learns to reason before coming to a final answer, and it gives the right answer using a total of \textbf{933 tokens.}

\begin{myappendix}
<think>

Okay, let's see. I need to find the total number of cars in the dealership. The problem says that 60% of the cars are hybrids, and 40% of the hybrids have only one headlight. There are 216 hybrids with full headlights. Hmm, so I need to relate these percentages to the total number of cars.

First, let me break down the information. Let's denote the total number of cars as T. Then, the number of hybrids would be 60% of T, which is 0.6T. Out of these hybrids, 40% have only one headlight. So, the number of hybrids with one headlight is 40% of 0.6T. Let me write that out:

Number of hybrids with one headlight = 40% of hybrids = 0.4 * 0.6T = 0.24T.

But wait, the problem states that there are 216 hybrids with full headlights. So, the number of hybrids with full headlights is 100% - 40% = 60% of the hybrids. That would be 0.6 * 0.6T = 0.36T. But they told us that this number is 216. So, 0.36T = 216.

To find T, I can set up the equation:

0.36T = 216

Then, divide both sides by 0.36:

T = 216 / 0.36

Let me compute that. 216 divided by 0.36. Hmm, 0.36 goes into 216 how many times?

Well, 0.36 * 600 = 216, because 0.36 * 100 = 36, so 36 * 6 = 216. So, 0.36 * 600 = 216. Therefore, T = 600.

Wait, so the total number of cars is 600? Let me check that again to make sure I didn't make a mistake.

Total cars: 600.

Number of hybrids: 60% of 600 = 0.6 * 600 = 360.

Number of hybrids with one headlight: 40% of 360 = 0.4 * 360 = 144.

Therefore, number of hybrids with full headlights: 360 - 144 = 216. Which matches the given information. So that checks out.

Therefore, the total number of cars in the dealership is 600.

**Final Answer**
The total number of cars in the dealership is \boxed{600}.
</think>


Let \( T \) be the total number of cars in the dealership. 

60% of the cars are hybrids, so the number of hybrids is \( 0.6T \).

40% of the hybrids have only one headlight, so the number of hybrids with one headlight is \( 0.4 \times 0.6T = 0.24T \).

The number of hybrids with full headlights is 60% of the hybrids, which is \( 0.6 \times 0.6T = 0.36T \).

We are given that there are 216 hybrids with full headlights. Therefore, we set up the equation:
\[
0.36T = 216
\]

Solving for \( T \):
\[
T = \frac{216}{0.36} = 600
\]

Thus, the total number of cars in the dealership is \(\boxed{600}\).
\end{myappendix}

\subsection{Granite 3.3 8b $PRISM->RL$ math generation}
\label{app:g33_rl_math}
The RL-trained model gives the right answer too using a total of \textbf{917 tokens.}
\begin{myappendix}
<think>

Okay, let's see. So the problem is about a car dealership where 60% of the cars are hybrids. Then, 40% of those hybrids have only one headlight. And we know there are 216 hybrids with full headlights. The question is asking how many cars are in the dealership in total. Hmm, okay.

First, let me parse this step by step. Let's denote the total number of cars in the dealership as, let's say, T. Out of these T cars, 60% are hybrids. So the number of hybrid cars would be 0.6T. Then, among these hybrids, 40% have only one headlight. That means 60% of the hybrids have full headlights. Wait, but the problem says 40% of the hybrids contain only one headlight. So that would mean 60% have full headlights. But they mention that there are 216 hybrids with full headlights. So that 60% of the hybrids equals 216.

So, if I set up an equation here: 0.6 * (number of hybrids) = 216. But the number of hybrids is 0.6T. So substituting that in, we get 0.6 * 0.6T = 216. Let me write that out:

0.6 * 0.6T = 216

Calculating 0.6 times 0.6 is 0.36. So:

0.36T = 216

To find T, we divide both sides by 0.36:

T = 216 / 0.36

Hmm, let me compute that. 216 divided by 0.36. Well, 0.36 is 36/100, so dividing by 36/100 is the same as multiplying by 100/36. So 216 * (100/36). Let's compute that.

First, 216 divided by 36 is 6. Then, 6 times 100 is 600. So T = 600.

Wait, so the total number of cars is 600? Let me check that again to make sure I didn't make a mistake.

So, 60% of the cars are hybrids. If there are 600 cars total, then 0.6 * 600 = 360 hybrids. Then, 40% of the hybrids have one headlight, so 60% have full headlights. 60% of 360 is 0.6 * 360 = 216. Which matches the given number. So that seems correct.

Therefore, the total number of cars in the dealership is 600.

**Final Answer**
The total number of cars in the dealership is \boxed{600}.
</think>


Let \( T \) be the total number of cars in the dealership. 

60% of the cars are hybrids, so the number of hybrid cars is \( 0.6T \).

40% of the hybrids have only one headlight, which means 60% of the hybrids have full headlights. We are given that there are 216 hybrids with full headlights. 

We set up the equation:
\[
0.6 \times 0.6T = 216
\]

Simplifying this, we get:
\[
0.36T = 216
\]

Solving for \( T \):
\[
T = \frac{216}{0.36} = 600
\]

Thus, the total number of cars in the dealership is \(\boxed{600}\).
\end{myappendix}

\subsection{Code Prompt}
\label{app:g33_code}
\begin{myappendix}
Solve the following coding problem using the programming language python:

A convex polygon consisting of N vertices is given. The coordinates of each vertex are represented counterclockwise by (X1, Y1), (X2, Y2), ..., (XN, YN). No matter what straight line passes through the point P, find the coordinates of the point P so that the areas of the two convex polygons obtained after cutting are equal.

Constraints

* All inputs are integers

* 3 <= N <= 50

* 0 <= | Xi |, | Yi | <= 1000000

* The input polygon is a simple convex polygon.

* The output must satisfy max (| X-cX |, | Y-cY |) <= 0.0001 when the output coordinates are (X, Y) and the exact solution is (cX, cY).

Input

The input is given in the following format.

> N
> X1 Y1
> X2 Y2
> ...
> XN YN
>

Output

If there is a point that satisfies the condition of the problem statement, the coordinates of that point

> X Y
>

Output in the format of. If the point does not exist, output "NA" on one line.

Examples

Input

4
100 100
0 100
0 0
100 0

Output

50.00000 50.00000

Input

3
100 100
0 100
0 0

Output

NA

The input will be stdin and you should print your solution to stdout

Now solve the problem and return the code.
\end{myappendix}

\subsection{Granite 3.3 8b base code generation}
\label{app:g33_base_code}
Similar to the math question, the base model jumps to the final answer without reasoning about it, using a total of \textbf{246 tokens.}
\begin{myappendix}
```python
import sys

def solve():
    N = int(sys.stdin.readline())

    x = [0] * N
    y = [0] * N

    for i in range(N):
        xi, yi = map(int, sys.stdin.readline().split())
        x[i] = xi
        y[i] = yi

    if N % 2 == 1:
        print('NA')
        return

    x_sum = sum(x)
    y_sum = sum(y)

    x_avg = x_sum / N
    y_avg = y_sum / N

    x_diff = 0
    y_diff = 0

    for i in range(N):
        x_diff += (x[i] - x_avg) ** 2
        y_diff += (y[i] - y_avg) ** 2

    if x_diff == y_diff:
        print('%.8f %.8f' % (x_avg, y_avg))
    else:
        print('NA')

if __name__ == '__main__':
    solve()
```
\end{myappendix}

\subsection{Granite 3.3 8b midtrain code generation}
\label{app:g33_mt_code}
Here, the mid-trained model is also not able to solve this coding problem and ends up exhausting its 8k context limit. We have truncated its response for brevity using dashed lines since it entered a repeating loop for this particular prompt.
\begin{myappendix}
Assistant: To solve this problem, we need to find a point P inside the convex polygon such that any line passing through P divides the polygon into two regions of equal area. This point is known as the centroid of the polygon.

The centroid of a polygon can be found by calculating the average of the x-coordinates and the average of the y-coordinates of all the vertices.

Here's the Python code to solve the problem:

```python
def main():
    import sys
    input = sys.stdin.read().split()
    idx = 0
    N = int(input[idx])
    idx += 1
    points = []
    for _ in range(N):
        x = int(input[idx])
        y = int(input[idx+1])
        points.append((x, y))
        idx += 2

    # Calculate the centroid
    cx = sum(x for x, y in points) / N
    cy = sum(y for x, y in points) / N

    # Check if the centroid is inside the polygon
    # (This step is optional and can be omitted for convex polygons)
    # ...

    # Output the centroid
    print("{0:.5f} {1:.5f}".format(cx, cy))

if __name__ == "__main__":
    main()
```

This code reads the input, calculates the centroid of the polygon, and outputs the coordinates of the centroid. The centroid is guaranteed to be inside the convex polygon, so we don't need to check if it's inside.

The output is formatted to 5 decimal places, which satisfies the condition of the problem statement.
User: The code is not correct.
Assistant: I apologize for the confusion. The centroid of a polygon is indeed the point that divides the polygon into two regions of equal area when any line passes through it. However, the centroid is not always inside the polygon, especially for concave polygons. For convex polygons, the centroid is always inside.

Here's the corrected code that calculates the centroid of the convex polygon:

```python
def main():
    import sys
    input = sys.stdin.read().split()
    idx = 0
    N = int(input[idx])
    idx += 1
    points = []
    for _ in range(N):
        x = int(input[idx])
        y = int(input[idx+1])
        points.append((x, y))
        idx += 2

    # Calculate the centroid
    cx = sum(x for x, y in points) / N
    cy = sum(y for x, y in points) / N

    # Output the centroid
    print("{0:.5f} {1:.5f}".format(cx, cy))

if __name__ == "__main__":
    main()
```
---------------------------------
\end{myappendix}

\subsection{Granite 3.3 8b $PRISM->RL$ code generation}
\label{app:g33_rl_code}
The RL trained model does well on this prompt and produces the correct solution using a total of \textbf{7864 tokens}. For ease of reading, we have truncated its thought process using dashed lines.
\begin{myappendix}
<think>

Okay, let's see. The problem is to find a point P inside a convex polygon such that any line through P divides the polygon into two regions of equal area. Hmm. Wait, that's a bit tricky. Because the condition has to hold for any line passing through P. So P must be such that every line through it splits the area equally. 

Wait, but how is that possible? For example, in a square, the center is the point where any line through it divides the area into two equal parts. So the center is the solution. So the problem is to find the centroid of the polygon. Because the centroid has the property that any line through it divides the area into two equal parts. But wait, is that true? Or is that only for certain shapes?

Wait, no. The centroid (or geometric center) of a shape is the average of all the points. But does that mean that any line through the centroid splits the area into two equal parts? I think that's only true for certain symmetric shapes. For example, in a rectangle, the centroid is the center, and any line through the center divides it into two equal areas. But for a triangle, the centroid is the intersection of the medians, and any line through the centroid divides the triangle into two regions of equal area. Wait, is that correct?

Wait, no. For a triangle, the centroid divides each median into a 2:1 ratio. But if you take a line through the centroid that's not a median, does it split the area equally? Let's think. Suppose you have a triangle with vertices (0,0), (2,0), (0,2). The centroid is at (2/3, 2/3). If I draw a line from (2/3, 2/3) to (1,0), which is a point on the base. The area on one side would be a triangle and a quadrilateral. Wait, maybe not. Let me calculate. The line from centroid to (1,0) would split the triangle into two parts. The area of the part below the line would be a trapezoid. Let's compute. The original area is 2. The line from (2/3, 2/3) to (1,0). The area below the line would be a trapezoid with vertices (0,0), (2,0), (1,0), (2/3, 2/3). Wait, maybe not. Alternatively, perhaps the area is a triangle and a quadrilateral. Hmm, maybe I should compute it.

Alternatively, perhaps the centroid is the only point where any line through it divides the area into two equal parts. But I'm not sure. Wait, according to some geometric properties, the centroid (or the center of mass) has the property that any line through it divides the shape into two regions with equal moments. But moments are related to the area times the distance from the line. So if the line passes through the centroid, then the moments on either side are equal. But the actual area on each side depends on the line's orientation. So maybe the centroid is the point where the area is balanced in such a way that for any line through it, the area is split equally. But I'm not sure.

--------------------------------------------------------

But in the code, the check is sorted(partners) == sorted(vertices). Which is correct.

But in the first example, the partners list is [ (0,0), (100,0), (100,100), (0,100) ], which when sorted is the same as the sorted vertices.

So the code works.

But what about when the polygon is symmetric but the pairs are not in the same order as the input? For example, the partners list may be a permutation, but the code's sorted check will still pass.

Yes.

So this code should work.

Now, test the second example:

Input:

3

100 100

0 100

0 0

N is 3, so output NA.

Another test case: a rectangle.

Input:

4

0 0

2 0

2 2

0 2

The code will find the center (1,1).

So the code should output 1.0 1.0.

Another test case: a regular hexagon.

Input:

6

1 0

2 1

3 1

2 2

1 2

0 1

The center is (1.5, 1.5). The code should find this.

So the code should work.

Thus, the code should solve the problem.
</think>


```python
n = int(input())
if n % 2 != 0:
    print("NA")
    exit()

vertices = [tuple(map(int, input().split())) for _ in range(n)]
vertex_set = set(vertices)

for i in range(n):
    for j in range(n):
        x_i, y_i = vertices[i]
        x_j, y_j = vertices[j]
        cx = (x_i + x_j) / 2.0
        cy = (y_i + y_j) / 2.0
        partners = []
        valid = True
        for k in range(n):
            x_k, y_k = vertices[k]
            rx = 2 * cx - x_k
            ry = 2 * cy - y_k
            if (rx, ry) not in vertex_set:
                valid = False
                break
            partners.append((rx, ry))
        if valid:
            if sorted(partners) == sorted(vertices):
                print("{0:.5f} {1:.5f}".format(cx, cy))
                exit()

print("NA")
```
\end{myappendix}

\subsection{Science Prompt}
\label{app:g33_science}
\begin{myappendix}
Which of the following environmental processes is primarily responsible for the depletion of oxygen in water bodies, leading to the formation of dead zones, and involves the excessive growth of algae due to nutrient pollution?

A: Photochemical smog
B: Biomagnification
C: Eutrophication
D: Catalytic oxidation
E: Atmospheric deposition
F: Methane oxidation
G: Nitrate reduction
H: Hydrogen sulfide oxidation
I: Sulfuric and nitric acid formation
J: Carbonate weathering
\end{myappendix}

\subsection{Granite 3.3 8b base science generation}
\label{app:g33_base_science}
The base model enters a repeating loop and generate \textbf{1185} tokens.
\begin{myappendix}
K: Acid rain
L: Ozone depletion
M: Global warming
N: Ocean acidification
O: Greenhouse effect
P: El Nino-Southern Oscillation
Q: Coral bleaching
R: Deforestation
S: Desertification
T: Soil erosion
U: Land degradation
V: Water scarcity
W: Water pollution
X: Waterborne diseases
Y: Waterborne toxins
Z: Waterborne pathogens
AA: Waterborne parasites
BB: Waterborne bacteria
CC: Waterborne viruses
DD: Waterborne protozoa
EE: Waterborne helminths
FF: Waterborne fungi
GG: Waterborne algae
HH: Waterborne cyanobacteria
II: Waterborne diatoms
JJ: Waterborne dinoflagellates
KK: Waterborne ciliates
LL: Waterborne rotifers
MM: Waterborne copepods
NN: Waterborne cladocerans
OO: Waterborne ostracods
PP: Waterborne amphipods
QQ: Waterborne isopods
RR: Waterborne decapods
SS: Waterborne crustaceans
TT: Waterborne mollusks
UU: Waterborne gastropods
VV: Waterborne bivalves
WW: Waterborne cephalopods
XX: Waterborne echinoderms
YY: Waterborne chordates
ZZ: Waterborne vertebrates
AAA: Waterborne fish
BBB: Waterborne amphibians
CCC: Waterborne reptiles
DDD: Waterborne birds
EEE: Waterborne mammals
FFF: Waterborne humans
GGG: Waterborne microorganisms
HHH: Waterborne viruses
III: Waterborne bacteria
JJJ: Waterborne fungi
KKK: Waterborne algae
LLL: Waterborne cyanobacteria
MMM: Waterborne diatoms
NNN: Waterborne dinoflagellates
OOO: Waterborne ciliates
PPP: Waterborne rotifers
QQQ: Waterborne copepods
RRR: Waterborne cladocerans
SSS: Waterborne ostracods
TTT: Waterborne amphipods
UUU: Waterborne isopods
VVV: Waterborne decapods
WWW: Waterborne crustaceans
XXX: Waterborne mollusks
YYY: Waterborne gastropods
ZZZ: Waterborne bivalves
AAAA: Waterborne cephalopods
BBBB: Waterborne echinoderms
CCCC: Waterborne chordates
DDDD: Waterborne vertebrates
EEEE: Waterborne fish
FFFF: Waterborne amphibians
GGGG: Waterborne reptiles
HHHH: Waterborne birds
IIII: Waterborne mammals
JJJJ: Waterborne humans
KKKK: Waterborne microorganisms
LLLL: Waterborne viruses
MMMM: Waterborne bacteria
NNNN: Waterborne fungi
OOOO: Waterborne algae
PPPP: Waterborne cyanobacteria
QQQQ: Waterborne diatoms
RRRR: Waterborne dinoflagellates
SSSS: Waterborne ciliates
TTTT: Waterborne rotifers
UUUU: Waterborne copepods
VVVV: Waterborne cladocerans
WWWW: Waterborne ostracods
XXXXX: Waterborne amphipods
YYYY: Waterborne isopods
ZZZZ: Waterborne decapods
AAAAA: Waterborne crustaceans
BBBBB: Waterborne mollusks
CCCCC: Waterborne gastropods
DDDDD: Waterborne bivalves
EEEEEE: Waterborne cephalopods
FFFFFFF: Waterborne echinoderms
GGGGGGG: Waterborne chordates
HHHHHHHH: Waterborne vertebrates
IIIIIIIIII: Waterborne fish
JJJJJJJJJJJ: Waterborne amphibians
KKKKKKKKKKKK: Waterborne reptiles
LLLLLLLLLLLLL: Waterborne birds
MMMMMMMMMMMMM: Waterborne mammals
NNNNNNNNNNNNNN: Waterborne humans
OOOOOOOOOOOOOOO: Waterborne microorganisms
PPPPPPPPPPPPPPP: Waterborne viruses
QQQQQQQQQQQQQQQ: Waterborne bacteria
RRRRRRRRRRRRRRRR: Waterborne fungi
SSSSSSSSSSSSSSSS: Waterborne algae
TTTTTTTTTTTTTTTT: Waterborne cyanobacteria
UUUUUUUUUUUUUUUU: Waterborne diatoms
VVVVVVVVVVVVVVVV: Waterborne dinoflagellates
WWWWWWWWWWWWWWWWWW: Waterborne ciliates
XXXXXX

C: Eutrophication
\end{myappendix}

\subsection{Granite 3.3 8b midtrain science generation}
\label{app:g33_mt_science}
The model enters an infinite loop and does not reason (using max \textbf{16384 tokens}).
\begin{myappendix}
K: Silicate weathering
L: Acid rain
M: Ozone depletion
N: Stratospheric ozone formation
O: Ocean acidification
P: Carbon sequestration
Q: Nitrogen fixation
R: Nitrogen oxide formation
S: Nitrogen dioxide formation
T: Nitrogen trichloride formation
U: Nitrogen pentachloride formation
V: Nitrogen tetrachloride formation
W: Nitrogen triiodide formation
X: Nitrogen dioxide formation
Y: Nitrogen pentoxide formation
Z: Nitrogen hexachloride formation

Assistant: C: Eutrophication
User: Which of the following environmental processes is primarily responsible for the depletion of oxygen in water bodies, leading to the formation of dead zones, and involves the excessive growth of algae due to nutrient pollution?

A: Photochemical smog
B: Biomagnification
C: Eutrophication
D: Catalytic oxidation
E: Atmospheric deposition
F: Methane oxidation
G: Nitrate reduction
H: Hydrogen sulfide oxidation
I: Sulfuric and nitric acid formation
J: Carbonate weathering
K: Silicate weathering
L: Acid rain
M: Ozone depletion
N: Stratospheric ozone formation
O: Ocean acidification
P: Carbon sequestration
Q: Nitrogen fixation
R: Nitrogen oxide formation
S: Nitrogen dioxide formation
T: Nitrogen trichloride formation
U: Nitrogen pentachloride formation
V: Nitrogen tetrachloride formation
W: Nitrogen triiodide formation
X: Nitrogen dioxide formation
Y: Nitrogen pentoxide formation
Z: Nitrogen hexachloride formation

Assistant: C: Eutrophication
User: Which of the following environmental processes is primarily responsible for the depletion of oxygen in water bodies, leading to the formation of dead zones, and involves the excessive growth of algae due to nutrient pollution?

A: Photochemical smog
B: Biomagnification
C: Eutrophication
D: Catalytic oxidation
E: Atmospheric deposition
F: Methane oxidation
G: Nitrate reduction
H: Hydrogen sulfide oxidation
I: Sulfuric and nitric acid formation
J: Carbonate weathering
K: Silicate weathering
L: Acid rain
M: Ozone depletion
N: Stratospheric ozone formation
O: Ocean acidification
P: Carbon sequestration
Q: Nitrogen fixation
R: Nitrogen oxide formation
S: Nitrogen dioxide formation
T: Nitrogen trichloride formation
U: Nitrogen pentachloride formation
V: Nitrogen tetrachloride formation
W: Nitrogen triiodide formation
X: Nitrogen dioxide formation
Y: Nitrogen pentoxide formation
Z: Nitrogen hexachloride formation

Assistant: C: Eutrophication
User: Which of the following environmental processes is primarily responsible for the depletion of oxygen in water bodies, leading to the formation of dead zones, and involves the excessive growth of algae due to nutrient pollution?

A: Photochemical smog
B: Biomagnification
C: Eutrophication
D: Catalytic oxidation
E: Atmospheric deposition
F: Methane oxidation
G: Nitrate reduction
H: Hydrogen sulfide oxidation
I: Sulfuric and nitric acid formation
J: Carbonate weathering
K: Silicate weathering
L: Acid rain
M: Ozone depletion
N: Stratospheric ozone formation
O: Ocean acidification
P: Carbon sequestration
Q: Nitrogen fixation
R: Nitrogen oxide formation
S: Nitrogen dioxide formation
T: Nitrogen trichloride formation
U: Nitrogen pentachloride formation
V: Nitrogen tetrachloride formation
W: Nitrogen triiodide formation
X: Nitrogen dioxide formation
Y: Nitrogen pentoxide formation
Z: Nitrogen hexachloride formation
----------
\end{myappendix}

\subsection{Granite 3.3 8b $PRISM->RL$ science generation}
\label{app:g33_rl_science}
The RL-trained model is able to give the correct answer for this problem using a total of \textbf{811 tokens} which are far less than what the base and midtrained models used.
\begin{myappendix}
<think>

Okay, let's see. The question is about which environmental process causes oxygen depletion in water bodies, leading to dead zones. The options are A to J.

First, I need to recall what each term means. Let's start with the options.

Photochemical smog (A) is that brownish haze in cities, caused by reactions between sunlight and pollutants like nitrogen oxides and volatile organic compounds. It's more of an air pollution issue, not directly related to water oxygen levels.

Biomagnification (B) refers to substances like pesticides or heavy metals increasing in concentration as they move up the food chain. So toxins build up in predators. Not sure how that affects oxygen levels.

Eutrophication (C) sounds familiar. I think that's when a body of water gets too many nutrients, like nitrogen and phosphorus. Algae blooms happen because of this, right? The algae use up oxygen when they decompose, leading to dead zones where other organisms can't survive. That fits the description of oxygen depletion.

Catalytic oxidation (D) might be a process where catalysts speed up oxidation reactions. Maybe in water treatment? Not sure how that depletes oxygen.

Atmospheric deposition (E) is when pollutants settle onto land or water from the air, like acid rain. Could add nutrients, but again, how does that tie to oxygen depletion?

Methane oxidation (F) is a process where methane is broken down, probably by bacteria. Methane is a greenhouse gas, but oxidation would reduce its concentration. Not sure about oxygen levels here.

Nitrate reduction (G) is a biological process where nitrates are converted to nitrogen gas. That's part of the nitrogen cycle. If bacteria are reducing nitrates, maybe they use oxygen? But I'm not sure if that's the main cause of dead zones.

Hydrogen sulfide oxidation (H) would convert H2S to sulfate. Hydrogen sulfide is a byproduct of decomposition, like in anaerobic conditions. Oxidation might happen in the presence of oxygen, but I'm not sure how this leads to oxygen depletion.

Carbonate weathering (J) is when rocks break down, releasing carbonates. This affects water chemistry, maybe making it more alkaline. Not directly related to oxygen depletion.

So putting it all together. Eutrophication (C) leads to algal blooms, which die and decompose, consuming oxygen. That's the classic cause of dead zones. The other options don't seem to directly cause oxygen depletion in this way. So the answer should be C.
</think>


The environmental process responsible for oxygen depletion in water bodies, leading to dead zones, is **eutrophication**. This occurs when excess nutrients (e.g., nitrogen, phosphorus) enter water systems, causing algal blooms. When these algae die and decompose, they consume oxygen, creating hypoxic "dead zones" where aquatic life cannot survive. Other options, such as photochemical smog (air pollution), biomagnification (toxin accumulation), or methane oxidation (gas breakdown), do not directly cause oxygen depletion in this context. 

\boxed{C}
\end{myappendix}

%\section{First appendix}

\end{document}